\documentclass[a4paper]{article}
\usepackage[a4paper, total={5.5in, 8in}]{geometry}

\usepackage[T1]{fontenc}
\usepackage[latin9]{inputenc}
\usepackage{comment}
\usepackage{geometry}
\usepackage{float}
\usepackage{url}
\usepackage{enumitem}
\usepackage{dcolumn}
\usepackage{calc}
\usepackage{amsmath}
\usepackage{amssymb}
\usepackage{epsfig}
\usepackage{booktabs}
\usepackage{multirow}
\usepackage{appendix}
\usepackage{array}
\usepackage[export]{adjustbox}
\usepackage[table,xcdraw]{xcolor}

\usepackage[table,xcdraw]{xcolor}
\usepackage{amsmath}
\usepackage{amssymb}
\usepackage{xspace} %For nice spacing in commands
\usepackage{multirow}
\usepackage{url}
\usepackage{enumitem}

\usepackage{soul,color}

\usepackage{array}
\newcolumntype{?}{!{\vrule width 1.2pt}} %to create think vertical lines in tables

\usepackage{graphicx}
\graphicspath{{./figs/}}
\usepackage[caption=false]{subfig}
\makeatletter
\newenvironment{subfigures}
 {\begin{minipage}{\columnwidth}\def\@captype{figure}\centering}
 {\end{minipage}}
\makeatother

\usepackage{algorithm}
\usepackage{algpseudocode} %[noend]
\algnewcommand\algorithmicforeach{\textbf{for each}}
\algnewcommand{\LeftComment}[1]{\State \hspace*{-0.3em} \(\triangleright\) #1}
\algnewcommand{\LeftCommentx}[1]{\State \(\triangleright\) #1}
\algdef{S}[FOR]{ForEach}[1]{\algorithmicforeach\ #1\ \algorithmicdo}

\makeatletter
\renewcommand{\ALG@beginalgorithmic}{\scriptsize}
\algrenewcommand\alglinenumber[1]{\scriptsize #1:}
\makeatother

\algrenewcommand\textproc{}

\let\oldReturn\Return
\renewcommand{\Return}{\State\oldReturn}

\usepackage{esvect}

\newcolumntype{P}[1]{>{\centering\arraybackslash}p{#1}}
\floatstyle{ruled}
\newfloat{algorithm}{tbp}{loa}

\providecommand{\algorithmname}{Algorithm}
\floatname{algorithm}{\protect\algorithmname}

\usepackage{babel}

\usepackage{lineno}
\modulolinenumbers[5]

\addto\captionsenglish{} %to fix a bug in elsearticle

\newcommand{\ALGNAME}{Distributed Hill Climbing}
\newcommand{\ALGPROC}{DistributedHillClimbing}
\newcommand{\ALGABRV}{DHC}

\begin{document}
\begin{comment}

\title{Online Distributed Evolutionary Optimization of Time Division Multiple Access Protocols}

%\corref{}

\author{Anil Yaman}
\ead{a.yaman@vu.nl}
\address{Department of Computer Science\\
Vrije Universiteit Amsterdam\\
Amsterdam, 1081 HV, The Netherlands}

\author{Tim van der Lee}
\ead{t.lee@tue.nl}
\address{Department of Electrical Engineering\\
Eindhoven University of Technology\\
Eindhoven, 5600MB , The Netherlands}

\author{Giovanni Iacca}
\ead{giovanni.iacca@unitn.it}
\address{Department of Information Engineering and Computer Science\\
University of Trento\\
Trento, 38123, Italy}

%\orcidID{0000-0003-1379-3778} %Anil
%\orcidID{0000-0001-9723-1830} %Giovanni
%\authorrunning{A. Yaman and G. Iacca}

%%%%%%%%%%%%%%%%%%%%%%%
\end{comment}

\begin{flushleft}
{\huge Online Distributed Evolutionary Optimization of Time Division Multiple Access Protocols} %on 
\bigskip

{\normalsize Anil Yaman\textsuperscript{1}*, Tim van der Lee\textsuperscript{2}, Giovanni Iacca\textsuperscript{3}}
\bigskip

{\small \textbf{1} Vrije Universiteit Amsterdam, Amsterdam, the Netherlands\\
\textbf{2} Eindhoven University of Technology, Eindhoven, the Netherlands\\
\textbf{3} University of Trento, Trento, Italy\\
}
\bigskip

{\small * AY: a.yaman@vu.nl}

\end{flushleft}

\section*{Abstract}
With the advent of cheap, miniaturized electronics, ubiquitous networking has reached an unprecedented level of complexity, scale and heterogeneity, becoming the core of several modern applications such as smart industry, smart buildings and smart cities. A crucial element for network performance is the protocol stack, namely the sets of rules and data formats that determine how the nodes in the network exchange information. A great effort has been put to devise formal techniques to synthesize (offline) network protocols, starting from system specifications and strict assumptions on the network environment. However, offline design can be hard to apply in the most modern network applications, either due to numerical complexity, or to the fact that the environment might be unknown and the specifications might not available. In these cases, online protocol design and adaptation has the potential to offer a much more scalable and robust solution. Nevertheless, so far only a few attempts have been done towards online automatic protocol design. Here, we envision a protocol as an emergent property of a network, obtained by an environment-driven \ALGNAME~algorithm that uses node-local reinforcement signals to evolve, at runtime and without any central coordination, a network protocol from scratch. We test this approach with a 3-state Time Division Multiple Access (TDMA) Medium Access Control (MAC) protocol and we observe its emergence in networks of various scales and with various settings. We also show how \ALGNAME~can reach different trade-offs in terms of energy consumption and protocol performance.
\\
\\
\textbf{Keywords.} Distributed Evolutionary Algorithm, Network Protocol, Online Adaptation.
%

%%%%%%%%%%%%%%%%%%%%%%%
%%%%%%%%%%%%%% INTRO %%%%%%%%%%%%%%%%%%%%%
\section{Introduction}
\label{sec:intro}

% NOTE & IDEAS
% - assuming that they work on the same task (or may this also not be required?)
% - malfunction of one node, adapt automatically?
% - mobile? why we need?
% - information exchange?
% - learning to communicate
% - when to communicate and avoid collisions
% - evolution of communication
% - dynamic, co-evolution, interesting because one change in the behavior of one node changes fitness of others

A fundamental element in many engineering and industrial applications is the use of networked systems: be it environment monitoring, smart industries, smart cities, or distribution systems, networks of various scales and complexity are employed today practically everywhere. One of the most important aspects in network design is the protocol stack, which determines the way (data format and rules) the nodes in a network communicate with each other~\cite{holzmann1991design}.

Traditionally, network protocols have been modelled as a \emph{reactive system}, i.e., a two-player game where an agent (a node in the network) \emph{reacts} --by performing a certain action-- to predefined conditions in the environment (the rest of the network): for instance, the agent retries a packet transmission if it does not receive an acknowledgment. As such, a protocol can be described with an automaton, for which formal specifications can be logically expressed and verified. For that, one usually needs to have complete knowledge about (and strict assumptions on) the environment. This approach, rooted in the theory of Temporal Logic and infinite (B{\"u}chi) automata~\cite{buchi1990solving}, has been the gold standard in protocol design and verification for decades. Since the late '60, an impressive number of theoretical and practical results have been obtained in this area, gearing towards the automatic synthesis of protocol from service specifications~\cite{saleh1991automatic,probert1991synthesis,carchiolo1992formal,saleh1996synthesis} and the development of automatic model checker tools, such as SPIN~\cite{holzmann1997model}. 

Despite these many successes, this approach to protocol design has also limitations. First of all, it assumes, in general, the environment --and all its states-- to be known: this might not be the case of some modern network applications, where the environment conditions might be unpredictable. Furthermore, this approach models the environment \emph{as a whole}, i.e., without describing the mutual interactions between the other nodes in the network (with few notable exceptions, see e.g.~\cite{al2012synthesizing,finkbeiner2017synthesis}). Another issue is the numerical (time and space) complexity of these methods, which makes them impractical when the number of protocol and environment states grows~\cite{vardi2018siren}. Finally, and this is arguably the main limitation, this approach fundamentally follows a \emph{waterfall design model}, as the design and verification steps are performed \emph{offline}, before deployment, and usually no online adaptation or feedback design cycles are considered. This is the case of projects such as x-kernel~\cite{hutchinson1991x}, Horus and Ensemble~\cite{birman2000horus} and --to some extent-- ANTS~\cite{wetherall1998ants}: all these tools offer a great level of modularity and abstraction, but --as pointed out by Keller et al.~\cite{keller2008system}-- their being based on offline design limits their flexibility and applicability. With the exception of ANTS, which also provides a way to dynamic reprogramming/redeploying the protocol code over the network, in all the other cases if one wants to change the protocol code all network nodes must be manually reprogrammed with the new protocol implementation: clearly, not only this approach disrupts the operation of the network, but also it becomes expensive when the network size increases.

In contrast to this traditional ``rigid'' offline approach, online protocol design and adaptation \cite{lee2020distributed} has the potential to offer a much more scalable and robust solution. Some researchers have even suggested that flexibility --in the form of self-adaptation and \emph{empowerment}, i.e., the principle of agents performing the actions which maximize the number of reachable states-- should be the key design principle in modern network engineering~\cite{kellerer2015flexibility,kalmbach2018empowering,kellerer2019adaptable}. Nevertheless, so far only a few attempts have been done in this direction, oriented towards the grand vision of \emph{autonomic} or \emph{self-adaptive} networking~\cite{bouabene2009autonomic,xiao2016bio}. Most of these attempts are based on Machine Learning (ML) and bio-inspired techniques, although they suffer from some limitations, as we briefly summarize below.

\textbf{Machine Learning}: Various solutions based on collective intelligence \cite{wolpert1999using} and Reinforcement Learning (RL) \cite{tao2001multi,peshkin2002reinforcement,stampa2017deep} have been proposed in the context of routing protocols and particularly in Wireless Sensor Networks (WSNs)~\cite{kulkarni2010computational,forster2010machine,alsheikh2014machine}; other works~\cite{he2019model} used Deep Learning (DL) to model and optimize the physical layer. Albeit quite powerful, the main limitation of most of these approaches is that they often require a large amount of data collected from the network at runtime, in order to build (i.e., train offline) a model of the protocol, to be used later for online adaptation and optimization. In this sense, we can consider these methods as ``semi-online''.

\textbf{Bio-inspired techniques}: a large body of research exists in this field, as surveyed for instance in~\cite{nakano2010biologically,dressler2010survey}. Similarly to the ML-based methods, most of the existing literature focuses however on offline optimization, based on Swarm Intelligence algorithms, such as Particle Swarm Optimization~\cite{guo2020optimizing} or Ant Routing~\cite{zhang2020optimized}, and especially on Evolutionary Algorithms (EAs)\footnote{Interestingly, a loop between network engineering and evolutionary theory exists. Recent evidence has shown that the ``hourglass'' shape of most protocol stacks is the result of an implicit evolutionary process that led to a minimal complexity, maximal robustness architecture~\cite{siyari2017emergence,siyari2017emergence}.}. In the context of EAs, a seminal paper is represented by the work from late '90 by El-Fakih et al.~\cite{el1999method}. Later on, Genetic Programming (GP) has been successfully used to \emph{evolve} --i.e., optimize offline-- protocol adaptors~\cite{van2003using}, wireless protocols based on Carrier Sense Multiple Access (CSMA)~\cite{tekken2017derivation}, aggregation protocols~\cite{weise2007genetic,weise2008evolving,weise2011evolving}, or MAC access protocols~\cite{lewis2006enhancing,roohitavaf2018synthesizing}. The latter have been evolved also by means of evolvable Finite State Machines (FSMs)~\cite{sharples2000protocol,hajiaghajani2015feasibility,hajiaghajani2015mac}.
As for online methods, an interesting bio-inspired (distributed) learning approach was introduced in ~\cite{su2010dynamic}, where each node observes the other nodes' behavior and forms internal conjectures on how they would react to its actions, to then choose the action that maximizes a local utility function: the authors demonstrated, analytically and through numerical simulations, that this method reaches Nash equilibria corresponding to optimal traffic fairness and throughput. Other works have investigated distributed EAs~\cite{iacca2013distributed} and distributed GP~\cite{johnson2005genetic,valencia2010distributed} to evolve the nodes' parameters and functioning logics of WSNs, or distributed optimization in multi-agent network systems. Finally, two notable online methods are STEM-Net~\cite{aloi2014stem} and Fraglets~\cite{yamamoto2007self}. The first one is a wireless network where each node uses an EA ``to reconfigure itself at multiple layers of the protocol stack, depending on environmental conditions, on the required service and on the interaction with other analogous device"~\cite{aloi2014stem}. The latter is based on the concept of ``autocatalytic software''~\cite{tschudin2005self}, or chemical computing~\cite{miorandi2008evolutionary}: essentially, protocols emerge automatically as collections of ``fraglets'', i.e., combinations of code segments and parameters which are evolved, respectively, by distributed GP~\cite{yamamoto2005genetic} and distributed EAs~\cite{alouf2010fitting}, and spread over the network through opportunistic (epidemic) propagation~\cite{alouf2007embedding} regulated by interactions with the environment. On top of this, an additional EA optimizes the combination of protocols~\cite{imai2010practical}.
%tschudin2003fraglets,
%yamamoto2005experiments,
%miorandi2006service,
%baude2010mixing,
%
\begin{figure}[!ht]
\begin{center}
\includegraphics[width = 0.7\columnwidth]{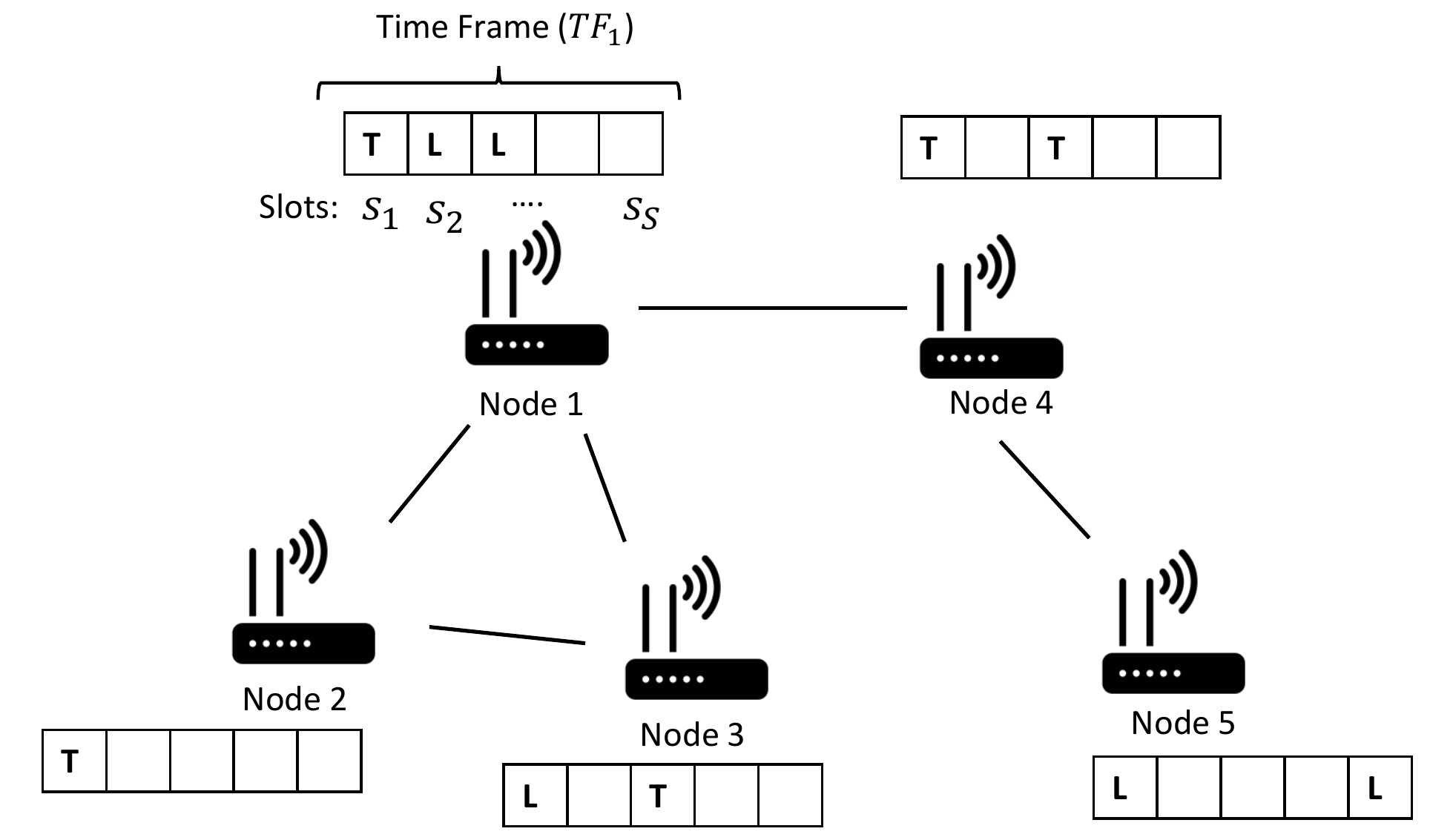}
\caption{Example network model used for evolving TDMA MAC protocols: each node has a time frame consisting of $S$ slots, where each slot $s_i, i \in \{1,\hdots, S\}$ can be either ``T'' (transmit), ``L'' (listen) or empty (idle).}
\label{fig:networkIllustration}
\end{center}
\end{figure}

%https://en.wikipedia.org/wiki/Medium_access_control
In this work, we continue along this research direction on online evolution of protocols. We consider a network of spatially distributed, locally connected nodes, and --to illustrate our method's applicability-- we focus on a Time Division Multiple Access (TDMA)-based Medium Access Control (MAC) protocol, where given a time frame consisting of a fixed number of time slots, each nodes should learn in which slots it should transmit, listen, or stay idle, see Figure~\ref{fig:networkIllustration} for an example. TDMA protocols are of particular interest in modern network research since the existing protocols are usually not suitable to handle the most recent network instances such as WSNs~\cite{lee2017tree}, vehicular ad-hoc networks~\cite{sun2020extensible} or underwater acoustic networks~\cite{yun2013hierarchical}. On the other hand, the proposed approach is applicable also to other protocols and stack layers.%thereof.

Our proposed solution works as follows. Starting from a \emph{tabula rasa}, nodes automatically learn the optimal protocol configuration --the one that minimizes collisions and/or reduces the energy consumption (for which we use, as proxy, the number of slots used for transmit / listen actions)-- online, and in a distributed manner. To do this, we propose an environment-driven approach, based on our previous work~\cite{yaman2020}, in which each node in the network runs \emph{in situ} a minimalist evolutionary search, receives reinforcement signals corresponding to its actions, and occasionally shares its protocol parameters with its neighbors. This approach aims to optimize the collective behavior of a population of agents that interact with a certain environment, and uses such interaction to drive a distributed evolutionary search. This approach has proven particularly successfully in robotics~\cite{eiben2010embodied,bredeche2012environment,haasdijk2014combining,prieto2016real,Bredeche2018}, while to the best of our knowledge it has never been applied to the evolution of network protocols. Compared to traditional protocol design, the advantage of our approach is at least twofold:
\begin{enumerate}[leftmargin=*]
 \item \textbf{Scalability}: since there is no central unit that guides the evolutionary search, this method can potentially scale up to very large networks.
 \item \textbf{Robustness}: since the evolutionary search runs continuously and open-ended, it allows a form of continual learning that can respond adequately to different network conditions, even unknown prior to deployment.
\end{enumerate}
In addition to this, the proposed method is fairly agnostic to the environment: by definition, the other nodes' behavior --in our case, the structure of their time frames-- is not known a priori as it will adapt online via embedded evolution. Furthermore, differently from other works such as~\cite{su2010dynamic}, there is not even the need to build a node-local conjecture on how the other nodes would behave. 
%NOTE: mention in the assumptions also synchronized frames and infinite queue?
% ack is received instantaneously (each T slot is divided into two sub-slots: packet transmission + ack listening)
In essence, our only assumptions are: 1) the node actions are taken from a finite action set that is the same for all nodes in the network (in our case: transmit, listen, idle); 2) each action is associated to a certain reinforcement signal that is the same in all the network; 3) for each transmission an ack packet can be received at the transmitter to acknowledge that the transmitted packet has been received correctly. Finally, it is worth highlighting once more that one distinct characteristic of our proposal w.r.t. the existing evolutionary approaches to MAC protocol optimization~\cite{lewis2006enhancing,roohitavaf2018synthesizing,sharples2000protocol,hajiaghajani2015feasibility,hajiaghajani2015mac} is that our approach is fully distributed (decentralized) and works online rather than offline.

The remaining of this paper is organized as follows. In the next section, we introduce the TDMA problem and the proposed \ALGNAME~algorithm. In Section~\ref{sec:setup}, we present the experimental setup, while the numerical results are discussed in Section~\ref{sec:results}. Finally, in Section~\ref{sec:conclusions} we draw the conclusions and hint at future works.
%%%%%%%%%%%%%%%%%%%%%%%%%%%%%%%%%%%%%%%%%%%%%%%%%%%%%%%%%%%%%%%

%%%%%%%%%%%%%%%%%%%%%%%%% METHODS %%%%%%%%%%%%%%%%%
\section{Methods}
\label{sec:methods}

% DEFINITIONS:
% n_i: node, i from 1 to N
% t_k: time step, k from 1 to K
% TF_i: time frame of node n_i
% s_t: time slot of TF_i, from 1 to S
% Q_i: (unlimited) queue of node n_i

\subsection{Problem settings} \label{sec:problemSettings}

The network, illustrated in Figure~\ref{fig:networkIllustration}, can be represented as a graph $G=(V,E)$ consisting of a set of nodes $V$, and undirected edges $E\subset V\times V$ which represent the possibility of communication between the nodes~\cite{ergen2010tdma}. Two nodes, seed and target, are selected in the network. All the packets originating from the seed node and are targeted to the target node.

Each node $n_i$ has a time frame, $\boldsymbol{TF}_i$, consisting of $S$ time slots that can take one of three values: ``T'', ``L'' or empty, to indicate transmit, listen and idle actions respectively. We assume that the time frames of all nodes have the same length and that their time slots are synchronized. In addition, each node has a queue data structure $Q_i$ of unlimited size to allow storing packets if they cannot be transmitted.

\begin{comment}
%
\begin{figure}[!ht]
\begin{center}
\includegraphics[width = \columnwidth]{figs/exampleGrid9.pdf}
\caption{Example network with nine nodes spatially distributed in a grid topology. Seed and target nodes are shown in red and green respectively.}
\label{fig:exampleGrid9}
\end{center}
\end{figure}

\end{comment}

The evaluation process of the whole network is performed for $K$ time steps. The process starts at time step $t_1 = 1$ by simultaneously executing the action in the first time slot of each node. Then, at each time step $t_k, k\in \{2,\hdots,K\}$, the action to be executed is the one in the next slot in the time frame, until the last slot is reached. Once the last slot is reached, the next action to be executed becomes the one from the time slot, and so on. This process is repeated until the last time step $t_{K}$ is reached.

During the evaluation process, a predefined number of packets are introduced into the queue of the seed node with a certain frequency. Depending on the actions in the time slots of the nodes, these packets are transmitted and received across nodes. For instance, if there is a packet in the queue of a node, and the action in the current time slot is ``T'', then the node attempts to transmit the packet. Every node in the neighborhood that performs the action ``L'' in the same time step can receive the packet. However, a \textit{collision} occurs when a node performing the action ``L'' has more than one neighbor that is attempting to transmit a packet at the same time. This can be observed in Figure~\ref{fig:networkIllustration}. In the first time slot, $s_0$, while ``Node 3'' listens, its two neighbors ``Node 1'' and ``Node 2'' try to transmit at the same time, which causes a collision. In this case the packet cannot be received. If there are no collisions, the packet is received by the neighbor and put into its queue. The neighbor that receives the packet sends an acknowledgement. In this case, the transmitting node removes the packet from its queue. If a packet is received at the target node, it is removed from its queue. The number of packets received at the target node and the number of time steps took for each packet to reach the target are recorded.

The goal of the problem is to assign an action to each time slot of the time frame of each node in such a way to allow all packets generated at the seed node to reach the target node ($100\%$ delivery rate). Ideally, it is also desirable to reduce the number of used slots (i.e., slots assigned to transmit and listen actions), to reduce the network energy consumption.

As a note, to avoid multiple transmissions of the same packet we include an additional data structure to keep track of the received packets in each node. To achieve that, we simply record the packet ID of each packet in each node when they are received. Moreover, if a node receives a packet with the same ID that was received before, it simply ignores it and does not send an acknowledgement.
%\vspace{-0.5cm}
\subsection{\ALGNAME}\label{sec:algorithm}

%TODO: the repo is not up?
Algorithm~\ref{alg:proposedalgo} shows the proposed \ALGNAME~(\ALGABRV) algorithm for optimizing TDMA MAC protocols over a network consisting of $N$ nodes. Each node $n_i, i\in \{1,\hdots,N\}$ runs the \ALGABRV~algorithm independently. The \ALGABRV~algorithm %\footnote{The code will be made publicly available.
%Code available at: \url{https://github.com/anilyaman/Evolution-of-Protocols}.
%}
starts with the initialization of the time frame $\boldsymbol{TF}_i$. Here, we initialize $\boldsymbol{TF}_i$ with empty time slots, to allow ``complexification'' of the time frames and emergence of frames with a minimum number of transmit/listen actions\footnote{On the other hand, random initialization is crucial in centralized approaches to obtain a diverse set of global network solutions that facilitate exploration and crossover.}.

\begin{algorithm}[!ht]
 \begin{algorithmic}[1]
 \Procedure{\ALGPROC}{}
 \State $\boldsymbol{TF}_i \gets$ initialize() \Comment{Empty slots}
 \State $f_i =$ eval($\boldsymbol{TF}_i$) \Comment{Local fitness}
 %\State $g = 1$
 \While {True}
 \State $\boldsymbol{TF}^{\prime} \gets $ mutate($\boldsymbol{TF}_{i}, mr$) \Comment{Mutation}\label{ln:mutation}
 \State $f^{\prime} \gets$ eval($\boldsymbol{TF}^{\prime}$) \Comment{Local fitness}
 \If {$f^{\prime} > f_i$} \Comment{Maximizing reward}\label{ln:comparison}
 \State $f_i \gets f^{\prime}$
 \State $\boldsymbol{TF}_i\gets \boldsymbol{TF}^{\prime}$
 \EndIf
 %\State $g = g + 1$
 \EndWhile \label{ln:endWhile}
 \EndProcedure
 \end{algorithmic}
\caption{\ALGNAME~algorithm used to optimize the time frame $\boldsymbol{TF}_i$ of node $n_i$ in a network.}\label{alg:proposedalgo}
\end{algorithm}

\begin{table}[!ht]
\caption{Node behavior reinforcement associations (NBRAs).}\label{tab:reinforcementFunction}
\begin{center}
\resizebox{0.7\columnwidth}{!}{
\begin{tabular}{|c|l|l|l|c|}
\hline
\multicolumn{1}{|l|}{\textbf{Behavior ID}} & \textbf{Action} & \textbf{Queue} & \textbf{Outcome} & \multicolumn{1}{l|}{${r_k}$} \\ \hline
\textbf{1} & Transmit & empty & - & ${r_1}$ \\ \hline
\textbf{2} & Transmit & non-empty & (ack) received & ${r_2}$ \\ \hline
\textbf{3} & Transmit & non-empty & (ack) not received & ${r_3}$ \\ \hline
\textbf{4} & Idle & empty & & ${r_4}$\\ \hline %+1
\textbf{5} & Idle & non-empty & & ${r_5}$ \\ \hline
\textbf{6} & Listen & empty & (packet) received & ${r_6}$ \\ \hline
\textbf{7} & Listen & non-empty & (packet) received & ${r_7}$\\ \hline %maybe -1 to encourage emptying the queue
\textbf{8} & Listen & empty & (packet) not received & ${r_8}$\\ \hline
\textbf{9} & Listen & non-empty & (packet) not received & ${r_9}$ \\ \hline
\end{tabular}
}
\end{center}
\end{table}

\subsubsection{Local fitness computation}

During the evaluation process, each node executes its time frame as discussed in Section~\ref{sec:problemSettings}. The fitness value of each node is then computed locally, as a cumulative sum of scores. We refer to these scores as \emph{reinforcements}. Each reinforcement is associated to the node behavior (i.e., a combination of action, queue and outcome) displayed in each time step of the evaluation. These reinforcements specify the rewards for the possible node behaviors. For instance, a transmit action would be more preferable if there is a packet in the queue of a node. However, if the queue of a node is empty, performing a transmit action would be unnecessary. Therefore, to encourage/discourage specific behaviors, it is possible to define rewarding (positive) or punishing (negative) reinforcements depending on the preference of the action in a certain situation. 

The complete list of node behavior reinforcement associations (NBRAs) defined in our experiments is given in Table~\ref{tab:reinforcementFunction}. For each node behavior, the corresponding reinforcement value $r_k$ is provided as the reward $r(t)$ at time step $t$. These reinforcements are aggregated to compute the local fitness value $f_i$ of each node $n_i$, as follows:
\begin{equation}
f_{i} = \begin{cases}
 C & \text{if } TF_{i,j} \text{ is empty } \forall j \in \{1,\dots,S\}; \\
 \sum^{K}_{t=1} r(t) &\text{otherwise.} \\ %f_{i} +
 \end{cases}
\label{eq:fitnessComputation}
\end{equation}
i.e., if all the time slots in $\boldsymbol{TF}_i$ are empty, we set $f_i$ to a large negative constant value $C$, which is intended to encourage transmit/listen actions. Else, we compute $f_i$ by aggregating the reinforcement values $r(t)$ obtained in each step $t$.

The NBRAs can be divided into two groups. The first group (ID: 2, 3, 5, 6, 7) aims to encourage the reception and transmission of the packets in the network. The second group (ID: 1, 4, 8, 9) aims to penalize unnecessary actions, thus implicitly minimizing the number of slots used for transmit/listen actions, a proxy for energy consumption.

Since each node tries to maximize the sum of its rewards, these reinforcements obviously play a crucial role in the optimization process. In principle, for the first group of behaviors the nodes should be rewarded/punished more severely since the actions corresponding to those behaviors directly affect the transmission of packets. On the other hand, the reinforcement signals for the second group can be relaxed, since as said they aim mainly at reducing unnecessary actions. However, it is especially important to find a balance on the reinforcement signals of the second group because if these rewards/punishments are too high the nodes will avoid actions which are supposedly unnecessary yet may lead to exploration of new connections and establishing of new pathways. On the other hand, if these rewards/punishments are too low, the nodes will perform too many unnecessary actions. Although this may lead to several established pathways, it would also increase the network energy consumption and yield to an unnecessarily high number of duplicate packets. For instance, punishing a node in the cases where it listens but nothing is received (behavior 8 and 9) would reduce the number of listen actions in the time frame. However, this would also reduce the probability of establishing possible new connections. Similarly, behavior 4 rewards a node if it remains idle when the queue is empty, yet other actions may be tried. To demonstrate that, we performed a sensitivity analysis to show the difference in results when different reinforcement values are used in behavior 4, 8 and 9, see Table~\ref{tab:reinforcementRuleTable} in the next Section.

\subsubsection{Mutation operator}

The \ALGABRV~algorithm uses a mutation operator to perturb the time frame $\boldsymbol{TF}_i$ of each node. The mutation operator (see Line~\ref{ln:mutation} in Algorithm~\ref{alg:proposedalgo}) simply samples, for each time slot, a new value with a probability of $mr$ (mutation rate). Sampling is performed by selecting with uniform probability a random action different from the one present in the time slot. 

%%%%%%%%%%%%%%%%%%%%%%%%%%%%%%%%%%%%%%%%%%%%%

%%%%%%%%%%%%%%%%%%%%%%%%  SETuP %%%%%%%%%%%%%%%%%%%%%%%%%%%%%%%%%%%%
\section{Experimental setup}
\label{sec:setup}

\subsection{Algorithm settings}

The exact reinforcement values used in the NBRAs (see $r_k$ in Table~\ref{tab:reinforcementFunction}) can be assigned differently. This in turn can change the rewards/punishments of certain actions. We tested the proposed algorithm for seven different NBRA assignments, given in Table~\ref{tab:reinforcementRuleTable}. These particular assignments (in the following, referred to as ``rules'') were defined based on domain knowledge. These rules differ only for the reinforcement values $r_4$, $r_8$ and $r_9$, to assess their effect on the minimization of the used slots. The other reinforcement values concern mainly the packet transmission, as discussed earlier in Section~\ref{sec:algorithm}. 

\begin{table}[ht!]
\begin{center}\caption{Rules for the node behavior reinforcement associations.}\label{tab:reinforcementRuleTable}
\begin{tabular}{|c|c|c|c|c|c|c|c|c|c|}
\hline
\textbf{Rule ID} & ${r_1}$ & ${r_2}$ & ${r_3}$ & ${r_4}$ & ${r_5}$ & ${r_6}$ & ${r_7}$ & ${r_8}$ & ${r_9}$ \\ \hline
\textbf{1} & -1 & 1 & -1 & 0 & -1 & 1 & 1 & 0 & 0 \\ \hline
\textbf{2} & -1 & 1 & -1 & 0.5 & -1 & 1 & 1 & 0 & 0 \\ \hline
\textbf{3} & -1 & 1 & -1 & 1 & -1 & 1 & 1 & 0 & 0 \\ \hline
\textbf{4} & -1 & 1 & -1 & 0 & -1 & 1 & 1 & -0.5 & -0.5 \\ \hline
\textbf{5} & -1 & 1 & -1 & 0 & -1 & 1 & 1 & -1 & -1 \\ \hline
\textbf{6} & -1 & 1 & -1 & 0.5 & -1 & 1 & 1 & -0.5 & -0.5 \\ \hline
\textbf{7} & -1 & 1 & -1 & 1 & -1 & 1 & 1 & -1 & -1 \\ \hline
\end{tabular}
\end{center}
\end{table}

Among the tested rules, rule 1 is the most relaxed one in terms of unnecessary slot use since it does not provide any rewards/punishments for $r_4$, $r_8$ and $r_9$. Rules 2 and 3 reward only for $r_4$, to keep idle slots empty if they are not used. Rules 4 and 5 aim to provide punishments for unnecessary listen actions. Rules 6 and 7 provide both rewards and punishments for not using unnecessary idle slots and performing unnecessary listen actions respectively.

We tested for various mutation rate values, $mr = \{0.01, 0.02, 0.04, 0.08, 0.16, 0.32, 0.64\}$. The maximum number of function evaluations was set to $10000$. However, to reduce the running time we stop the evolutionary process as soon as a solution that obtains $100\%$ delivery rate of the packets to the target node is found (even though further refinement might be possible). For each rule and mutation rate, the algorithm was executed for $28$ independent runs. %Similar settings were used also for the benchmark algorithms, described below in Section~\ref{sec:benchmarkAlg}, except that for those algorithms we allowed to reach the maximum number of function evaluations ($10000$) and we used a default mutation rate.

\subsection{Network settings}\label{subsec:networkconf}

%An example grid network structure with 9 nodes is illustrated in Figure~\ref{fig:exampleGrid9}. 
We tested the proposed algorithm on square (Manhattan-like) grids and randomly generated network topologies consisting of 9, 36 and 81 nodes. In the case of grid networks, the seed and target nodes nodes are placed in opposite corners. %In these networks the average number of connections per node is 3.55, and the shortest distance (in terms of number of connections) between the seed and target nodes is 4, 10 and 16 in the case of 9, 36 and 81 nodes respectively.

%The network statistics of the grid structured networks (average connectivity per node, and the shortest distance between seed and target nodes) are given in Table~\ref{tab:gridNetworkStats}. 
% \begin{comment}
% \begin{table}[!ht]
% \caption{Average number of connections per node and shortest distance between the seed and target nodes in networks with grid topologies consisting of 9, 36 and 81 nodes.}
% \label{tab:gridNetworkStats}
% \begin{center}
% \begin{tabular}{|c|c|c|c|}
% \hline
% \textbf{Topology} & \textbf{3$\times$3} & \textbf{6$\times$6} & \textbf{9$\times$9} \\ \hline
% \textbf{Average connections} & 3.55 & 3.55 & 3.55 \\ \hline
% \textbf{Shortest distance} & 4 & 10 & 16 \\ \hline
% \end{tabular}
% \end{center}
% \end{table}
% \end{comment}

In the case of random topologies, we
considered a 2D Cartesian plane $[0,1]^2$ where the seed and target nodes are assigned to the coordinates $(0,0)$ and $(1,1)$ respectively, while the remaining $N-2$ nodes
are assigned to random $(x,y)$ coordinates. The connectivity of the random networks are adjusted based on two parameters, \textit{connection distance} ($cd$) and \textit{connection probability} ($cp$). A connection is established between two nodes when the Euclidean distance between them is smaller than $cd$, with probability $cp$: i.e., for all $i,j\in \{1,\hdots,N\} \text{ where } i\neq j$, node $n_i$ is connected to node $n_j$ if $\text{dist(}n_i,n_j\text{)}<cd \text{ and } rand<cp$, where dist() is the Euclidean distance and $rand$ is a real valued uniform random variable in $[0,1]$. 

\begin{figure}[!b]
\vspace{-0.6cm}
\begin{subfigures}
% \subfloat[{\scriptsize $9$ nodes, $cd = 0.8$, $cp = 1$}]{\includegraphics[width=0.5\columnwidth]{figs/randNet9cd08cp1.pdf}\label{fig:randNet9cd08cp1}}
% \subfloat[{\scriptsize $9$ nodes, $cd = 0.8$, $cp = 0.5$}]{\includegraphics[width=0.5\columnwidth]{figs/randNet9cd08cp05.pdf}\label{fig:randNet9cd08cp05}}
% \subfloat[{\scriptsize $9$ nodes, $cd = 0.8$, $cp = 0.25$}]{\includegraphics[width=0.5\columnwidth]{figs/randNet9cd08cp025.pdf}\label{fig:randNet9cd08cp025}}
% \subfloat[{\scriptsize $9$ nodes, $cd = 0.8$, $cp = 0.125$}]{\includegraphics[width=0.5\columnwidth]{figs/randNet9cd08cp0125.pdf}\label{fig:randNet9cd08cp0125}}
%
%
% \subfloat[{\scriptsize $36$ nodes, $cd = 0.5$, $cp = 1$}]{\includegraphics[width=0.5\columnwidth]{figs/randNet36cd05cp1.pdf}\label{fig:randNet36cd05cp1}}
% \subfloat[{\scriptsize $36$ nodes, $cd = 0.5$, $cp = 0.5$}]{\includegraphics[width=0.5\columnwidth]{figs/randNet36cd05cp05.pdf}\label{fig:randNet36cd05cp05}}
% \subfloat[{\scriptsize $36$ nodes, $cd = 0.5$, $cp = 0.25$}]{\includegraphics[width=0.5\columnwidth]{figs/randNet36cd05cp025.pdf}\label{fig:randNet36cd05cp025}}
% \subfloat[{\scriptsize $36$ nodes, $cd = 0.5$, $cp = 0.125$}]{\includegraphics[width=0.5\columnwidth]{figs/randNet36cd05cp0125.pdf}\label{fig:randNet36cd05cp0125}}
%
\subfloat[{\scriptsize $81$ nodes, $cd = 0.3$, $cp = 1$}]{\includegraphics[width=0.5\columnwidth]{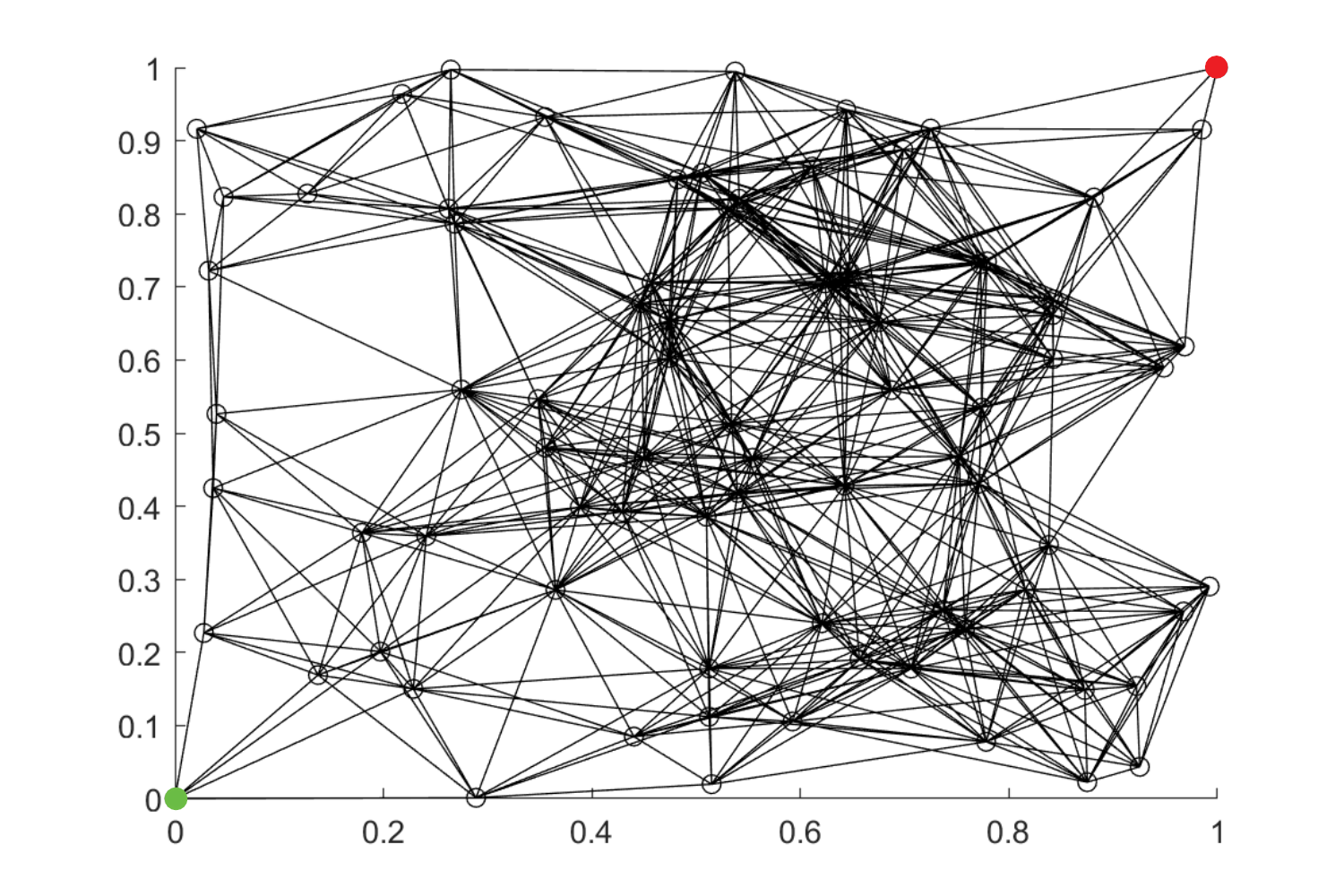}\label{fig:randNet81cd03cp1}}
%\subfloat[{\scriptsize $81$ nodes, $cd = 0.3$, $cp = 0.5$}]{\includegraphics[width=0.5\columnwidth]{figs/randNet81cd03cp05.pdf}\label{fig:randNet81cd05cp05}}
%\subfloat[{\scriptsize $81$ nodes, $cd = 0.3$, $cp = 0.25$}]{\includegraphics[width=0.5\columnwidth]{figs/randNet81cd03cp025.pdf}\label{fig:randNet81cd03cp025}}
\subfloat[{\scriptsize $81$ nodes, $cd = 0.3$, $cp = 0.125$}]{\includegraphics[width=0.5\columnwidth]{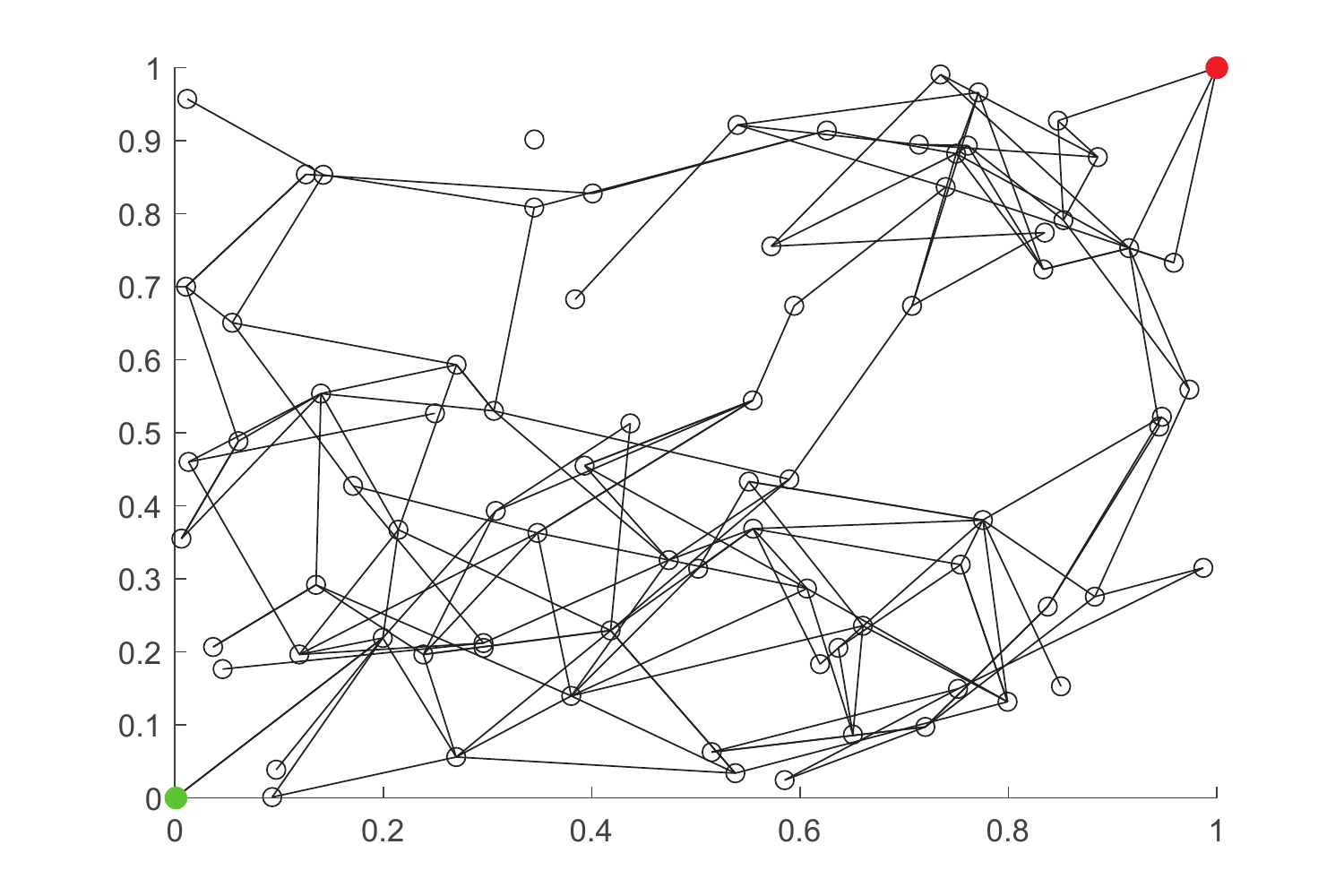}\label{fig:randNet81cd03cp0125}}
\end{subfigures}
%Examples of random networks with 9, 36 and 81 nodes.
\caption{Examples of random networks with 81 nodes. Connection distance ($cd$) and connection probability ($cp$) are used to control the connectivity of the networks. Seed (in location $(0,0)$) and target (in location $(1,1)$) nodes are shown in green and red respectively.}\label{fig:random-nets}
\end{figure}

We set the number of time slots $S$ to be equal to the number of nodes $N$. During the evaluation process of the networks, $M=5$ packets are generated in the seed node to be delivered to the target node. We set the number of steps $K$ to $M \times S$. For $9$, $36$ and $81$ nodes we set $cd$ to $0.8$, $0.5$ and $0.3$ respectively. For each network size, we considered four $cp$ values, namely $cp=\{1, 0.5, 0.25, 0.125\}$. Therefore, in total we considered $12$ random network configurations with various sizes and connectivity, see Figure~\ref{fig:random-nets} for two examples. For each random network configuration we generated $28$ networks and ran the algorithm once for each one independently.

% \begin{table}[!ht]
% \caption{Average number of connections per node and average shortest distance between the seed and target nodes in random networks with 9, 36 and 81 nodes.}
% \label{tab:randNetworkStats}
% \begin{center}
% \resizebox{\columnwidth}{!}{
% \begin{tabular}{|c|c|c|c|c|c|}
% \hline
% \textbf{Nodes} &\textbf{Connection probability} ($\boldsymbol{cp}$) & \textbf{1} & \textbf{0.5} & \textbf{0.25} & \textbf{0.125} \\ \hline
% \multirow{2}{*}{$9$} & \textbf{Average connections} & 5.61 & 2.91 & 1.58 & 0.99 \\ \cline{2-6}
% & \textbf{Average shortest distance} & 2.42 & 3.07 & 2.96 & 3.07 \\ \hline\hline
% \multirow{2}{*}{$36$} &\textbf{Average connections} & 15.79 & 7.92 & 3.85 & 2.07 \\ \cline{2-6}
% &\textbf{Average shortest distance} & 3.89 & 4.10 & 4.75 & 6.35 \\ \hline\hline
% \multirow{2}{*}{$81$} & \textbf{Average connections}&16.40 & 8.34 & 4.07 & 2.09 \\ \cline{2-6}
% & \textbf{Average shortest distance} & 6 & 6.30 & 7.67 & 9.97 \\ \hline
% \end{tabular}
% }
% \end{center}
% \end{table}

\subsection{Benchmark algorithms}\label{sec:benchmarkAlg}

To compare the results of our proposed approach, we considered seven centralized offline algorithms from the literature. For these algorithms, we conducted the optimization in a centralized (network-level) fashion, i.e., optimizing global network solutions obtained by concatenating the time frames of all nodes. The evaluation was performed as described in Section \ref{sec:problemSettings}, in this case after deconstructing the global network solution into individual node time frames.

Contrarily to the proposed distributed approach, where each node locally maximizes its aggregated reward (Eq.~\ref{eq:fitnessComputation}), the centralized algorithms do not make use of any online reinforcement during the fitness evaluation. Instead, they use \emph{a posteriori} objective functions based on the global performance of the networks, namely the hop distance of the packets to the target node, and the node activity (used slots). The former is computed as follows:
\begin{equation}
 f^{(1)} = \max_{i=1,\hdots,M}\left(\min_{c=1,\hdots,L_i} \left(h^{c}_{i}\right)\right)
 \label{eq:fglobal1}
\end{equation}
%
% %
% \begin{comment}
% \begin{equation}
% p_j = min\left( h_{j,c}, \forall \in \{1,\hdots,L_{copies}\}\right)
% \label{eq:pcopies}
% \end{equation}
% \end{comment}
where $h^{c}_{i} = \text{hop}(p^{c}_{i}, n_{target})$. Here,
$p^{c}_{i}$ indicates the $c$-th copy of the $i$-th packet and hop() measures the shortest hop distance from the node reached by $p^{c}_{i}$ to the target node $n_{target}$, whereas $L_{i}$ and $M$ indicate respectively the number of copies of a given packet and the number of unique packets originating from the seed node (in our case $M=5$). Finally, $f^{(1)}$ is scaled in $[0,1]$ by dividing the maximum possible hop distance in a network (from any node to the target node).

For the latter, we simply find the average number of non-empty slots per node as follows:
\begin{equation}
 f^{(2)} = \frac{1}{N S} \sum^N_{i=1} \sum^S_{j=1} A(i,j)
 \label{eq:fglobal2}
\end{equation}
where:
\[A(i,j) = \begin{cases}
 1, & \text{if }TF_{i,j} \text{ is ``T'' (transmit) or ``L'' (listen),} \\
 0, & \text{otherwise.} \\
 \end{cases}
\]
We divide the benchmark algorithms into three groups:

\begin{enumerate}[leftmargin=*]
\item \textbf{Group 1 (Pareto multi-objective optimization)}: NSGA-II~\cite{deb2002fast} and MSEA~\cite{tian2019multistage}. The two algorithms were configured to perform Pareto optimization minimizing $f^{(1)}$ and $f^{(2)}$. For these experiments we used the implementation of NSGA-II and MSEA available in the PlatEMO platform~\cite{PlatEMO}, with the default parameter settings. The representation and evolutionary operators were adjusted to handle our ternary representation. %In case of MSEA, we also adjusted the distance measure to Jaccard distance between the solutions.

\item \textbf{Group 2 (scalarized multi-objective optimization)}: Centralized Hill Climbing with 2 Objectives (CHC2O), Centralized Simulated Annealing with 2 Objectives (CSA2O) and Genetic Algorithm with 2 Objectives (GA2O). In this group, we minimize the sum of two objectives as $f^{(1)} + f^{(2)} \in [0,2]$. The implementation of CHC2O and CSA2O is the same as CHC and CSA (in Group 3), except for the use of the second objective. In the case of GA2O, we configured the algorithm with roulette wheel selection with 10 elites, $1$-point crossover operator with $0.9$ probability, and the mutation operator used in \ALGABRV.

%The objectives are the same used for NSGA-II and MSEA, i.e., maximum distance to the target and number of used slots. In this case both objectives were normalized in the range $[0,1]$ 
 
\item \textbf{Group 3 (single-objective optimization)}: Centralized Hill Climbing (CHC) and Centralized Simulated Annealing (CSA) algorithms. The CHC and CSA iteratively perform mutations on a single centralized global network solution. In CHC, only solutions that are better or equal to the current solution are accepted for the next iteration, whereas CSA uses a temperature parameter $T$ to accept worse solutions based on the probability $e^{{(f_{new} - f_{old})}/{T}}$. $T$ is scheduled to be reduced at each iteration to decrease this probability ($T\gets \alpha T, \alpha < 1$) ~\cite{kirkpatrick1983optimization}. In our experiments, we assign $0.9$ for $\alpha$. Both CHC and CSA minimize $f^{(1)}$.
\end{enumerate}

For the population-based algorithms (NSGA-II, MSEA, and GA2O) we used a population of $50$ solutions and we set the maximum number of function evaluations to $10000$ (without any early stop). For CHC and CSA (both with single and two objectives) we set $10000$, with early stop in case of $100\%$ delivery rate solution found.

%%%%%%%%%%%%%%%%%%%%%%%%%%%%%%%%%%%%%%%%%%%%%%%%%%%%%%%%%%%%

%%%%%%%%%%%%%%%%%%%%%%%%%%% RESULTS %%%%%%%%%%%%%%%%%%%%%%%%%%%%%%%%%

\section{Experimental results}\label{sec:results}

%%%%%%%%%%%%%%%% SCALABILITY %%%%%%%%%%%%%%%%
\subsection{Evaluating scalability}

The complete comparison of the results is given in Table~\ref{tab:resultsComparison}. The results of \ALGABRV~and Group 3 are the best results obtained across all mutation rates tested (see the Supplementary Material). We should note that, except for the cases marked by ``-'', all the compared algorithms reach $f^{(1)}=0$ (distance to target node), which yields $100\%$ delivery rate. For this reason, we compare the algorithms on $f^{(2)}$ (slot use). In the case of Group 1, we select from the final Pareto front the solution with $f^{(1)}=0$ and minimum $f^{(2)}$ for comparison. Once again it is worth stressing that our method does not optimize explicitly nor $f^{(1)}$ or $f^{(2)}$, but it optimizes node-local rewards (Eq.~\ref{eq:fitnessComputation}).%Furthermore, the algorithms in Group 3 optimize only $f^{(1)}$.

The results show the superiority in terms of scalability of \ALGABRV~over the benchmark algorithms\footnote{The standard deviations and statistical analysis of the results are provided in the Supplementary Material. The significance of the results is measured by the Wilcoxon rank-sum test.}. When the size of the problem is low (i.e., in the case of 9 nodes), the algorithms based on a multi-objective optimization approach (in particular NSGA-II) and the ones that combines two objectives (i.e., GA2O, CHC2O and CSA2O) find better solutions in terms of ratio of used slots. On the other hand, while the size of the problem increases, \ALGABRV~is able to find solutions with a smaller ratio of used slots relative to the other algorithms. We observe a further improvement in the performance of \ALGABRV~when sparsity increases (i.e., $cp$ decreases). However, the standard deviations of the results provided by \ALGABRV~are relatively higher than those obtained by the multi-objective optimization algorithms, which might be due to the fact that differently from these algorithms \ALGABRV~does not minimize explicitly the ratio of used slots.

%NOTE: saved tgn file in DEE folder
\begin{table*}[!ht]
\caption{Comparison of the algorithms on grid and random networks with 9 to 81 nodes. The problem size is computed as $N$ (no. of nodes) $\times$ $S$ (no. of slots), which yields to $N^2$ since we set $N=S$. The value in each cell shows the ratio of used slots (median across 28 runs at the end of the optimization process) for the solutions found by the algorithms (``-'' indicates that no solution is found).}\label{tab:resultsComparison}
\vspace{-0.3cm}

\begin{center}
\resizebox{\textwidth}{!}{
\begin{tabular}{lc?c|c?c|c|c?c|c?c|c|c|c|c|c|c|}
%\hline
\cline{3-16} & & \multicolumn{2}{c?}{\textbf{Group 1}} & \multicolumn{3}{c?}{\textbf{Group 2}} & \multicolumn{2}{c?}{\textbf{Group 3}} & \multicolumn{7}{c|}{\textbf{\ALGABRV}}\\ \hline
\multicolumn{1}{|l|}{\textbf{Problem}} & \textbf{Size} & \textbf{NSGA-II} & \textbf{MSEA} & \textbf{GA2O} & \textbf{CHC2O} & \textbf{CSA2O} & \textbf{CHC} & \textbf{CSA} & \textbf{R1} & \textbf{R2} & \textbf{R3} & \textbf{R4} & \textbf{R5} & \textbf{R6} & \textbf{R7} \\ \hline
\rowcolor[HTML]{EFEFEF} 
\multicolumn{1}{|l|}{\textbf{Grid9}} & 81 & \textbf{0.10} & 0.23 & \textbf{0.10} & 0.17 & 0.15 & 0.66 & 0.66 & 0.64 & 0.38 & 0.35 & 0.41 & 0.28 & 0.33 & 0.34 \\ \hline
\multicolumn{1}{|l|}{\textbf{9cp1}} & 81 & \textbf{0.07} & 0.22 & \textbf{0.07} & \textbf{0.07} & \textbf{0.07} & 0.60 & 0.60 & 0.50 & 0.31 & 0.31 & 0.32 & 0.25 & 0.28 & 0.25 \\ \hline
\rowcolor[HTML]{EFEFEF} 
\multicolumn{1}{|l|}{\textbf{9cp05}} & 81 & \textbf{0.07} & 0.23 & \textbf{0.07} & \textbf{0.07} & \textbf{0.07} & 0.64 & 0.64 & 0.57 & 0.34 & 0.31 & 0.31 & 0.26 & 0.30 & 0.26 \\ \hline
\multicolumn{1}{|l|}{\textbf{9cp025}} & 81 & - & - & \textbf{0.07} & 0.10 & 0.10 & 0.71 & 0.69 & 0.55 & 0.28 & 0.28 & 0.28 & 0.25 & 0.28 & 0.23 \\ \hline
\rowcolor[HTML]{EFEFEF} 
\multicolumn{1}{|l|}{\textbf{9cp0125}} & 81 & - & - & \textbf{0.07} & \textbf{0.07} & 0.12 & 0.75 & 0.73 & 0.56 & 0.27 & 0.28 & 0.27 & 0.23 & 0.27 & 0.25 \\ \hline\hline
\multicolumn{1}{|l|}{\textbf{Grid36}} & 1296 & 0.43 & 0.55 & 0.49 & \textbf{0.31} & \textbf{0.31} & 0.63 & 0.64 & 0.60 & 0.44 & 0.43 & 0.37 & 0.46 & 0.41 & 0.52 \\ \hline
\rowcolor[HTML]{EFEFEF} 
\multicolumn{1}{|l|}{\textbf{36cp1}} & 1296 & 0.45 & 0.56 & 0.54 & 0.38 & 0.38 & 0.55 & 0.59 & 0.35 & 0.35 & 0.50 & \textbf{0.34} & 0.39 & 0.39 & 0.50 \\ \hline
\multicolumn{1}{|l|}{\textbf{36cp05}} & 1296 & 0.43 & 0.56 & 0.49 & 0.33 & 0.33 & 0.40 & 0.53 & 0.40 & 0.35 & 0.38 & 0.38 & \textbf{0.25} & 0.26 & 0.41 \\ \hline
\rowcolor[HTML]{EFEFEF} 
\multicolumn{1}{|l|}{\textbf{36cp025}} & 1296 & 0.42 & 0.58 & 0.48 & 0.32 & 0.31 & 0.58 & 0.60 & 0.47 & 0.31 & 0.29 & 0.29 & \textbf{0.21} & 0.25 & 0.37 \\ \hline
\multicolumn{1}{|l|}{\textbf{36cp0125}} & 1296 & 0.41 & 0.60 & - & 0.57 & 0.55 & 0.66 & 0.65 & 0.53 & 0.26 & 0.33 & 0.22 & 0.22 & \textbf{0.21} & 0.26 \\ \hline\hline
\rowcolor[HTML]{EFEFEF} 
\multicolumn{1}{|l|}{\textbf{Grid81}} & 6561 & 0.59 & 0.62 & 0.58 & 0.52 & 0.52 & 0.59 & 0.59 & 0.57 & 0.49 & 0.48 & 0.46 & 0.52 & \textbf{0.44} & 0.52 \\ \hline
\multicolumn{1}{|l|}{\textbf{81cp1}} & 6561 & 0.66 & 0.65 & 0.63 & 0.63 & 0.64 & 0.49 & 0.49 & \textbf{0.42} & 0.67 & 0.67 & 0.67 & 0.83 & - & 0.66 \\ \hline
\rowcolor[HTML]{EFEFEF} 
\multicolumn{1}{|l|}{\textbf{81cp05}} & 6561 & 0.60 & 0.62 & 0.59 & 0.55 & 0.56 & 0.43 & 0.44 & \textbf{0.38} & 0.49 & 0.58 & 0.50 & 0.51 & 0.56 & 0.50 \\ \hline
\multicolumn{1}{|l|}{\textbf{81cp025}} & 6561 & 0.59 & 0.64 & 0.58 & 0.52 & 0.52 & 0.51 & 0.57 & 0.46 & \textbf{0.43} & 0.51 & 0.45 & 0.46 & 0.49 & 0.48 \\ \hline
\rowcolor[HTML]{EFEFEF} 
\multicolumn{1}{|l|}{\textbf{81cp0125}} & 6561 & 0.59 & 0.64 & - & 0.59 & 0.58 & 0.66 & 0.65 & 0.50 & 0.37 & 0.39 & \textbf{0.36} & 0.37 & 0.37 & 0.39 \\ \hline
\end{tabular}
}
\end{center}
\end{table*}

%%%%%%%%%%%%%%%% TRADEOFF %%%%%%%%%%%%%%%%

\begin{figure*}[!ht]
\vspace{-0.6cm}
%\begin{center}
\begin{subfigures}
\subfloat[9 nodes ($cd = 0.8, cp = 1$)]{\includegraphics[width=0.45\columnwidth]{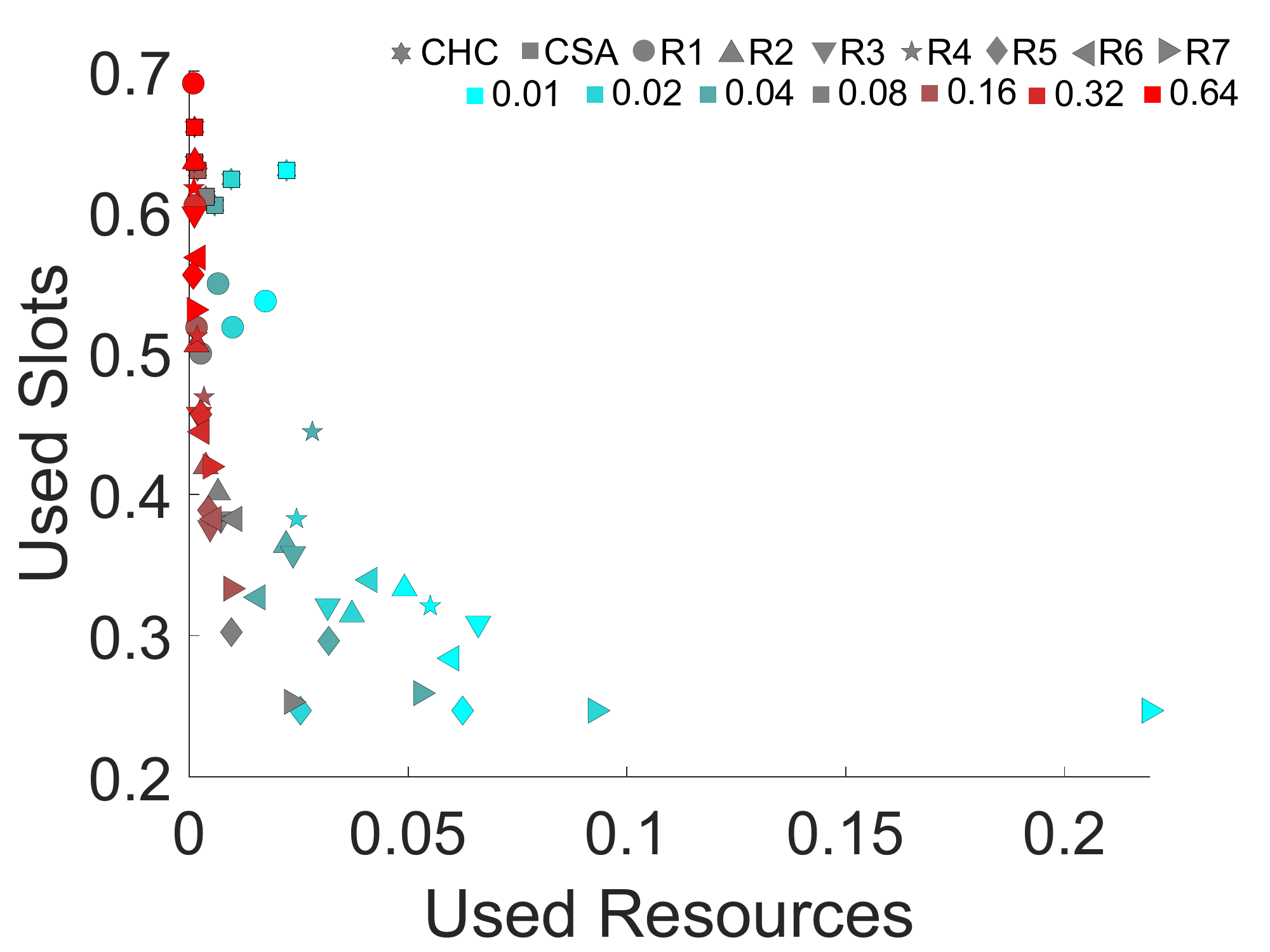}\label{fig:scatter9nodesRcd08cp1}}
\subfloat[9 nodes ($cd = 0.8, cp = 0.125$)]{\includegraphics[width=0.45\columnwidth]{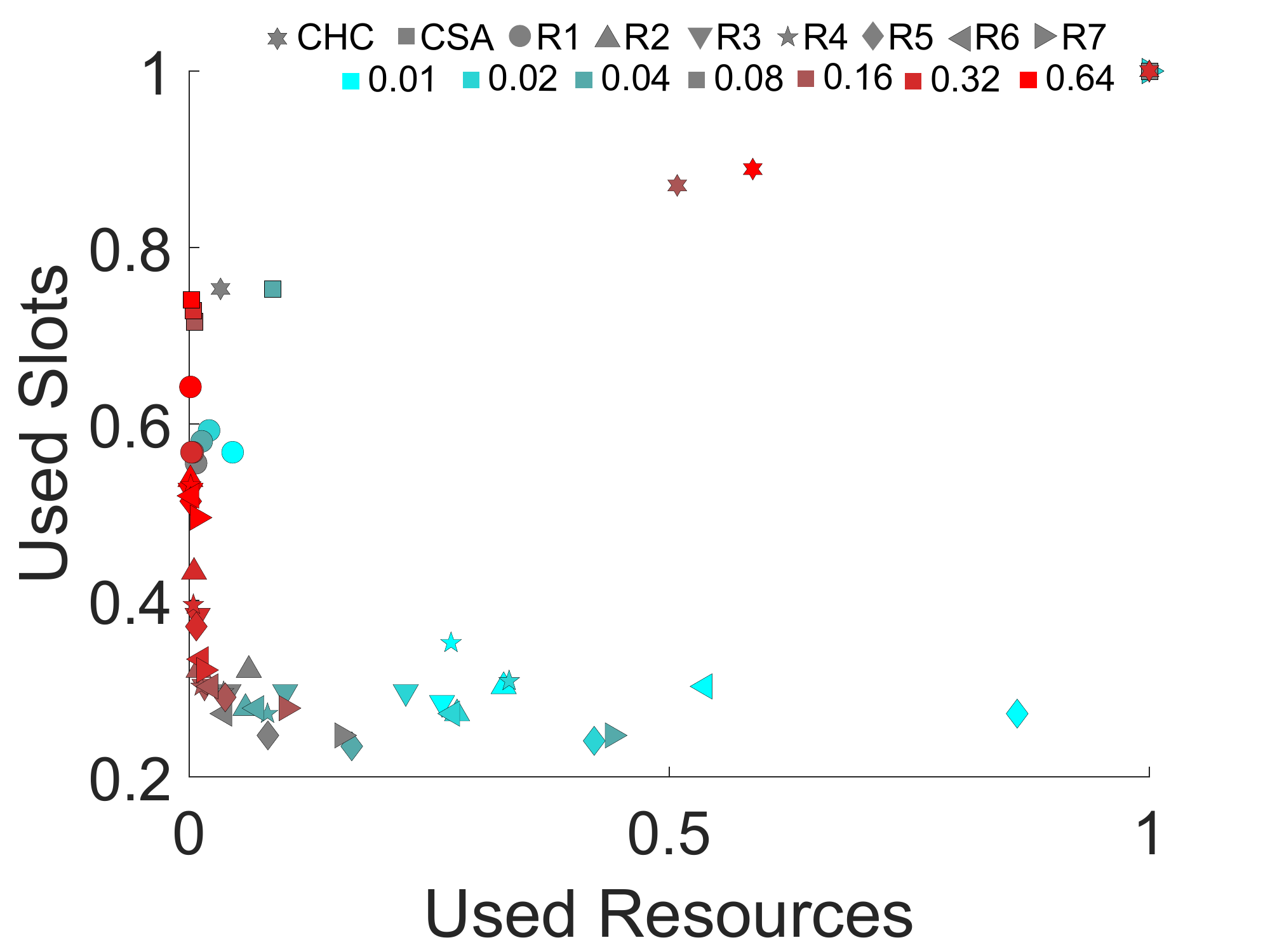}\label{fig:scatter9nodesRcd08cp0125}}

\subfloat[36 nodes ($cd = 0.5, cp = 1$)]{\includegraphics[width=0.45\columnwidth]{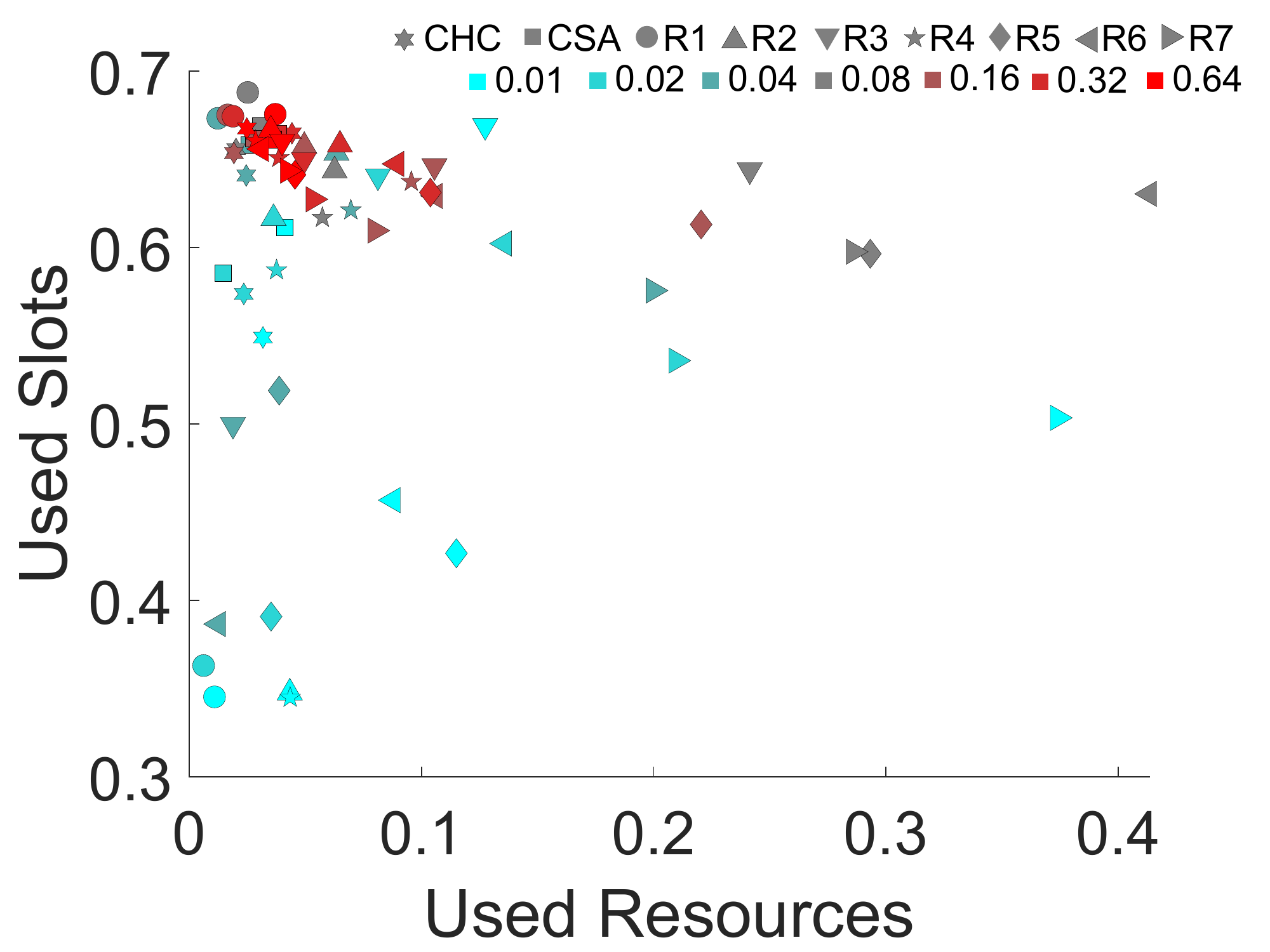}\label{fig:scatter36nodesRcd05cp1}}
\subfloat[36 nodes ($cd = 0.5, cp = 0.125$)]{\includegraphics[width=0.45\columnwidth]{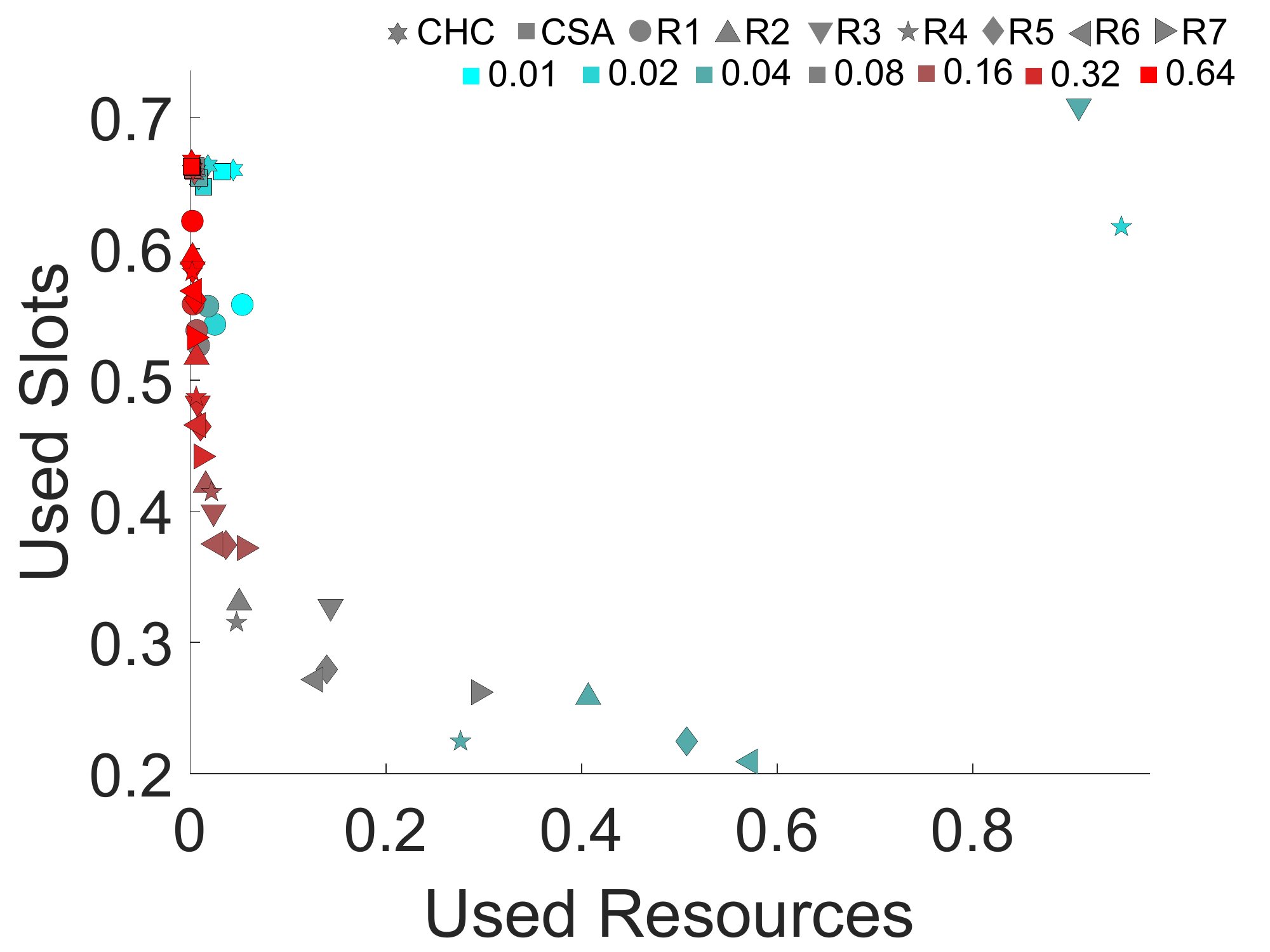}\label{fig:scatter36nodesRcd05cp0125}}

\subfloat[81 nodes ($cd = 0.3, cp = 1$)]{\includegraphics[width=0.45\columnwidth]{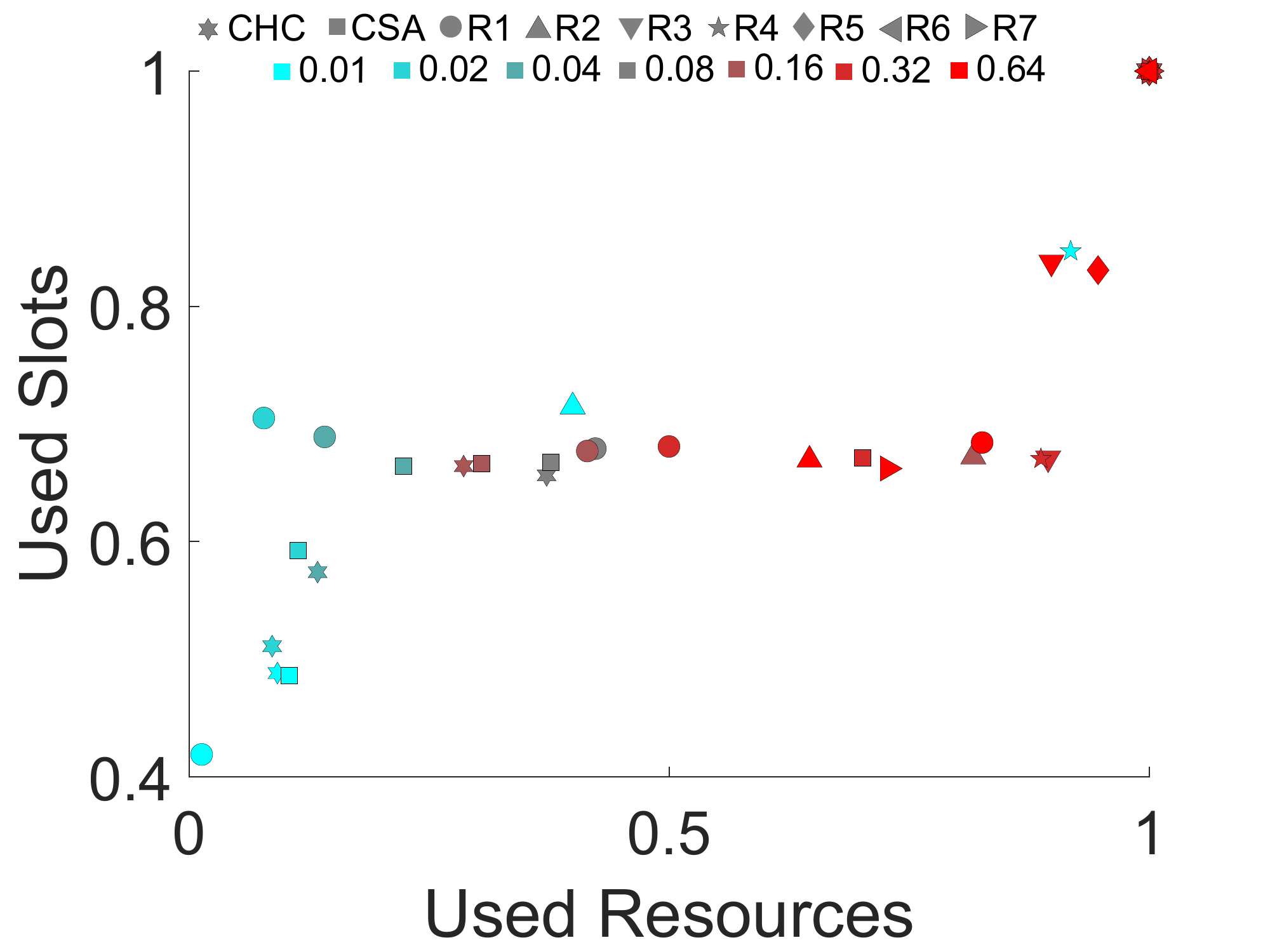}\label{fig:scatter81nodesRcd03cp1}}
\subfloat[81 nodes ($cd = 0.3, cp = 0.125$)]{\includegraphics[width=0.45\columnwidth]{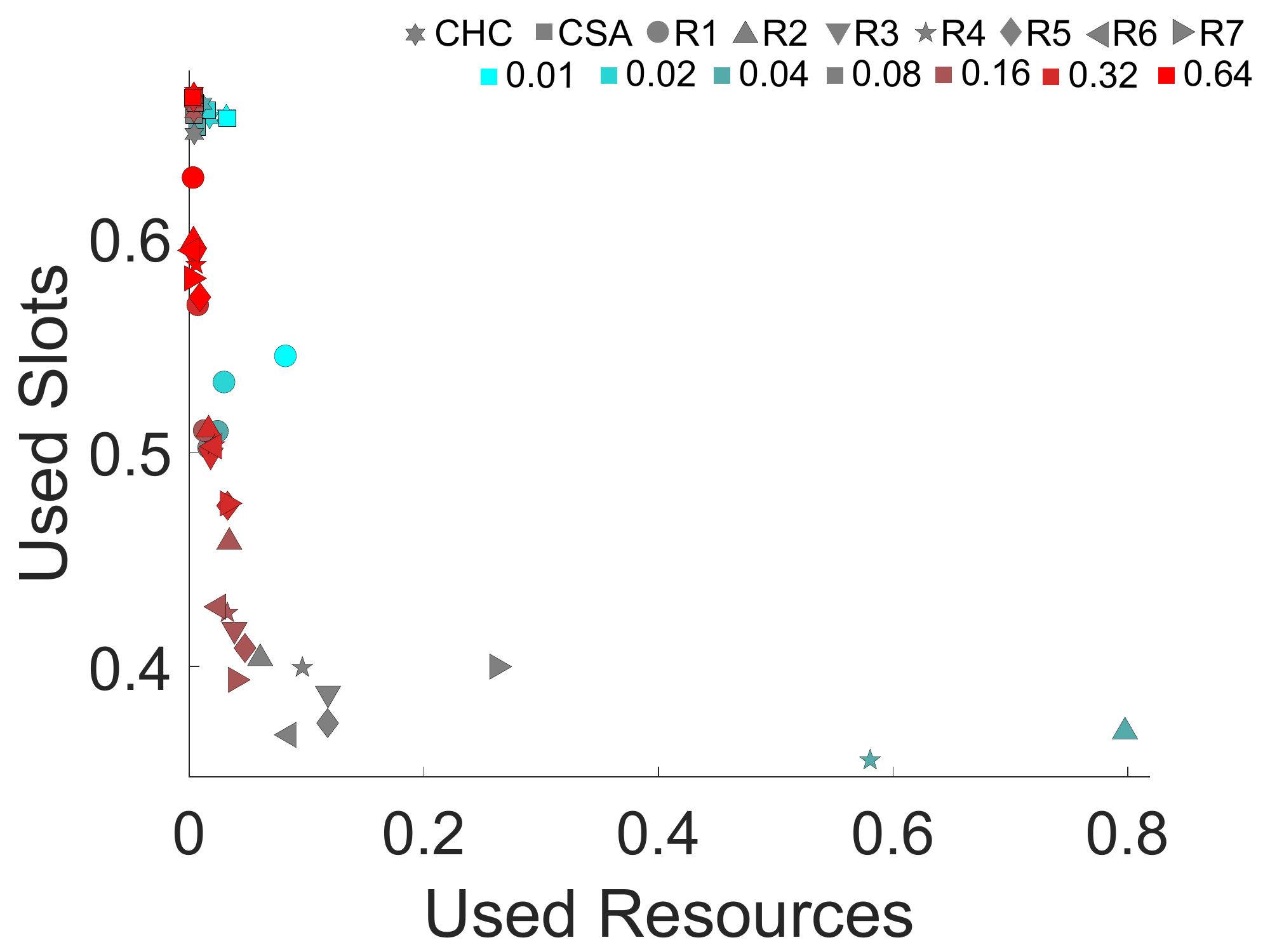}\label{fig:scatter81nodesRcd03cp0125}}
\end{subfigures}
%\end{center}
\caption{Ratio of used slots vs ratio of used resources (median across $28$ runs) of CHC, CSA and the proposed \ALGABRV~algorithm (with seven different rules $\times$ seven different mutation rates) on six selected random network configurations (the remaining ones are reported in the Supplementary Material).}\label{fig:scatterPlotsRand}
\vspace{-0.3cm}
\end{figure*}

%\subsection{Trade-off between slots and resources}
Another important aspect to analyze is the effectiveness, in terms of function evaluations, of the proposed approach. Figure~\ref{fig:scatterPlotsRand} %, \ref{fig:scatterPlotsRand36} and \ref{fig:scatterPlotsRand81} 
shows the results of CHC, CSA and the proposed \ALGABRV~with the seven rules shown in Table~\ref{tab:reinforcementRuleTable} in terms of used resources and used slots on six selected random network configurations. Here, \emph{used resources} refer to the ratio of function evaluations needed to find a solution with $100\%$ delivery rate, while \emph{used slots} refer to the ratio of non-empty slots in the solution found. The results (median across $28$ runs) of different algorithms and mutation rates are shown using different shapes and colors\footnote{The results for the other configurations of the grid and random networks and all the complete numerical data are provided as Supplementary Material.}. %The results of the other algorithms that use two objective are not comparable in these plots since they use maximum resources.

Overall, CHC and CSA are able to find a solution quickly. However, they often find solutions with a large ratio of used slots. On the other hand, \ALGABRV~is able to find solutions with a smaller ratio of used slots, although different reward/punishment assignments produce solutions with different ratios of used slots. In particular, we observe that rules which are more restrictive in terms of reward/punishment of unnecessary slot use (i.e., 5, 6, 7) tend to find solutions with a smaller ratio of used slots. However, they tend to use more fitness evaluations. Furthermore, we observe that higher mutation rates provide more randomness, resulting in similar results w.r.t. the CHC and CSA algorithms. In those cases, the solutions are found quite quickly but the ratio of slot use is high. Slightly higher mutation rates appear to be helpful for more restrictive rules rather than for the less restrictive ones. Overall, the best performing mutation rate ranges in $0.01-0.04$. However, the exact mutation rate that performs the best varies across different network configurations and rules.

%%%%%%%%% ROBUSTNESS %%%%%%%%%%%%%%%%

\subsection{Evaluating robustness}

We tested \ALGABRV~in four scenarios which require adaptation of previously found solutions to new conditions of the network. These scenarios include addition, removal, relocation and reinitialization of the nodes in the network. In case of addition, $10$ nodes were included into the networks at random locations, with random time frames. In case of removal, $6$ randomly selected nodes were removed from the networks. In case of relocation, $18$ randomly selected nodes were reassigned to new randomly initialized locations. In the case of reinitialization, the time frame of $18$ randomly selected nodes were reinitialized randomly. 

The experimental process involves first finding solutions for the initial networks, then applying each of the perturbations, independently.
After each perturbation, we establish links based on the $cp$ and $cd$ parameters, as described in Section~\ref{subsec:networkconf}. Then, we run the optimization process for $10000$ evaluations to find solutions adapted to the new conditions\footnote{Examples of delivery rate and fitness trends of the nodes during the evolutionary processes before and after the perturbations can be found in the Supplementary Material.}.

Table~\ref{tab:robustness} shows the results of \ALGABRV~after perturbations. The ``Used resources'' sub-table indicates the median of the ratio of the function evaluations used to find a $100\%$ delivery rate solution for the perturbed networks, while the ``Used slots'' sub-table indicates the ratio of occupied slots in the corresponding solutions. ``Initial solution'' indicates the median of the ratio of used resources and median of ratio of used slots found for the initial network prior to the perturbation. We tested the algorithm on random networks with $36$ nodes, using the best rules found in Table~\ref{tab:resultsComparison} (shown in bold for each network configuration).

Concerning the resource use, we observe that the algorithm can find solutions to the perturbed networks in a short time, usually using less than 10\% of the allocated function evaluations. However, adaptation requires more evaluations in the case of the network with the highest sparsity ($cp=0.125$).

Concerning the ratio of used slots, this seems to increase in general after perturbations. However, we observe that removal and relocation do not appear to have a high impact, especially in sparse networks. On the other hand, addition and reinitialization produces an increased slot use in all networks. This is likely due to the random reinitialization of the time frames occurring during the addition and reinitialization processes.

%NOTE: saved tgn file in DEE folder
\begin{table}[!ht]
\vspace{-0.3cm}
\caption{Results found by \ALGABRV~after perturbing the initial solution by addition, removal, relocation and reinitialization.}\label{tab:robustness}
\vspace{-0.3cm}
\begin{center}
\resizebox{0.7\columnwidth}{!}{
\begin{tabular}{|l|c|c|c|c|c|}
\hline
 & \multicolumn{5}{c|}{\textbf{Used resources}} 
 \\ \cline{2-6} & & \multicolumn{4}{c|}{\textbf{Adaptation after initial solution}}
 \\ \cline{3-6} 
 \multirow{-3}{*}{\textbf{Problem}} & \multirow{-2}{*}{\textbf{\begin{tabular}[c]{@{}c@{}}Initial \\ solution\end{tabular}}} & \textbf{Add} & \textbf{Remove} & \textbf{Relocate} & \textbf{Reinit.} \\ \hline
\textbf{36cd05cp1} & 0.04 & 0.27 & 0.01 & 0.05 & 0.03\\ \hline
\textbf{36cd05cp05} & 0.17 & 0.02 & 0.06 & 0.07 & 0.02\\ \hline
\textbf{36cd05cp025} & 0.37 & 0.06 & 0.58 &0.26 & 0.06\\ \hline
\textbf{36cd05cp0125} & 0.57 & 0.25 & 0.35 & 0.59 & 0.86\\ \hline\hline

& \multicolumn{5}{c|}{\textbf{Used slots}} \\ \cline{2-6} 
& & \multicolumn{4}{c|}{\textbf{Adaptation after initial solution}} 
\\ \cline{3-6}
\multirow{-3}{*}{\textbf{Problem}} & \multirow{-2}{*}{\textbf{\begin{tabular}[c]{@{}c@{}}Initial \\ solution\end{tabular}}} & \textbf{Add} & \textbf{Remove} & \textbf{Relocate} & \textbf{Reinit.} \\ \hline
\textbf{36cd05cp1} & 0.34 & 0.64 & 0.49 & 0.64 & 0.67\\ \hline
\textbf{36cd05cp05} & 0.25 & 0.49 & 0.32 &0.33 & 0.53\\ \hline
\textbf{36cd05cp025} & 0.21 & 0.37 & 0.21 & 0.21 & 0.54\\ \hline
\textbf{36cd05cp0125} &0.21 &0.36 & 0.20& 0.20 & 0.40\\ \hline
\end{tabular}
}
\end{center}
\vspace{-0.6cm}
\end{table}

%%%%%%%%%%%%%%%%%%%%%%%%%%%%%%%%%%%%%%%%%%%%%%%%%%%%%%%%%%%%

%%%%%%%%%%%%%%%%%%%%%%%%%%% Conclusions %%%%%%%%%%%%%%%%%%%%%%%%%%%%%%%%%

\section{Conclusions}
\label{sec:conclusions}

We proposed a decentralized, online evolutionary optimization algorithm, referred to as \ALGNAME~(\ALGABRV), for optimizing a TDMA MAC protocol on a distributed network of nodes. In the proposed approach, each node evolves its time frame locally. To do that, we devised a set of reinforcement rules to assign predefined scores to the actions of the nodes and compute their local fitness accordingly. Overall, we tested seven different reinforcement rules. We found that the \ALGABRV~algorithm was able to evolve, in all the tested network scenarios, efficient TDMA MAC protocols with $\%100$ delivery rate. Moreover, even though \ALGNAME~was not configured to explicit minimize the node activity (i.e., the number of used time slots), different reinforcement rules allowed the emergence of various protocols with different quality in terms of node activity.

We compared our algorithm with seven centralized single and multi-objective approaches, in which the optimization is performed at a global network level concatenating the time frames of all nodes in the network. In the multi-objective cases, we introduced a second objective function to minimize explicitly the node activity. Based on our comparisons, the benchmark algorithms with the explicit second objective showed better performance in terms of node activity only on the low-dimensional scenarios. On the other hand, \ALGABRV~showed better performance when the network size increased.

%The \ALGABRV~algorithm is a continuous distributed evolutionary optimization approach where the information about the performance of the whole network (i.e. the packet delivery rate) is not available to the local nodes. Thus, it can be useful to implement distributed stopping criteria when a solution is found. We implemented such a criterion by propagating a stopping signal to the network starting from the target node when 100\% delivery rate achieved. However, in our experiments we stopped the optimization process to reduce the runtime of the algorithms. %In future work, it would be interesting to explore distributed approaches to stop or resume the evolutionary optimization processes autonomously.

An interesting direction for future investigations will be to extend the algorithm to other rules, and applying it to different network layers where online adaptation might provide a benefit. Furthermore, it will be interesting to verify the proposed protocols in hardware.

%%%%%%%%%%%%%%%%%%%%%%%%%%%%%%%%%%%%%%%%%%%%%%%%%%%%%%%%%%%%
%%%%%%%%%%%%%%%%%%%%%%%%%%%%%%%%%%%%%%%%%%%%%%%%%%%%%%%%%%%%

\clearpage

%\bibliographystyle{elsarticle-harv}
%\bibliographystyle{elsarticle-num}
%\bibliography{bibliography}

\appendix

\section{Parameter analysis}

\begin{figure}[!ht]
\begin{subfigures}
\subfloat[{\scriptsize \% Used resources (3$\times$3 nodes)}]{\includegraphics[width=0.45\columnwidth]{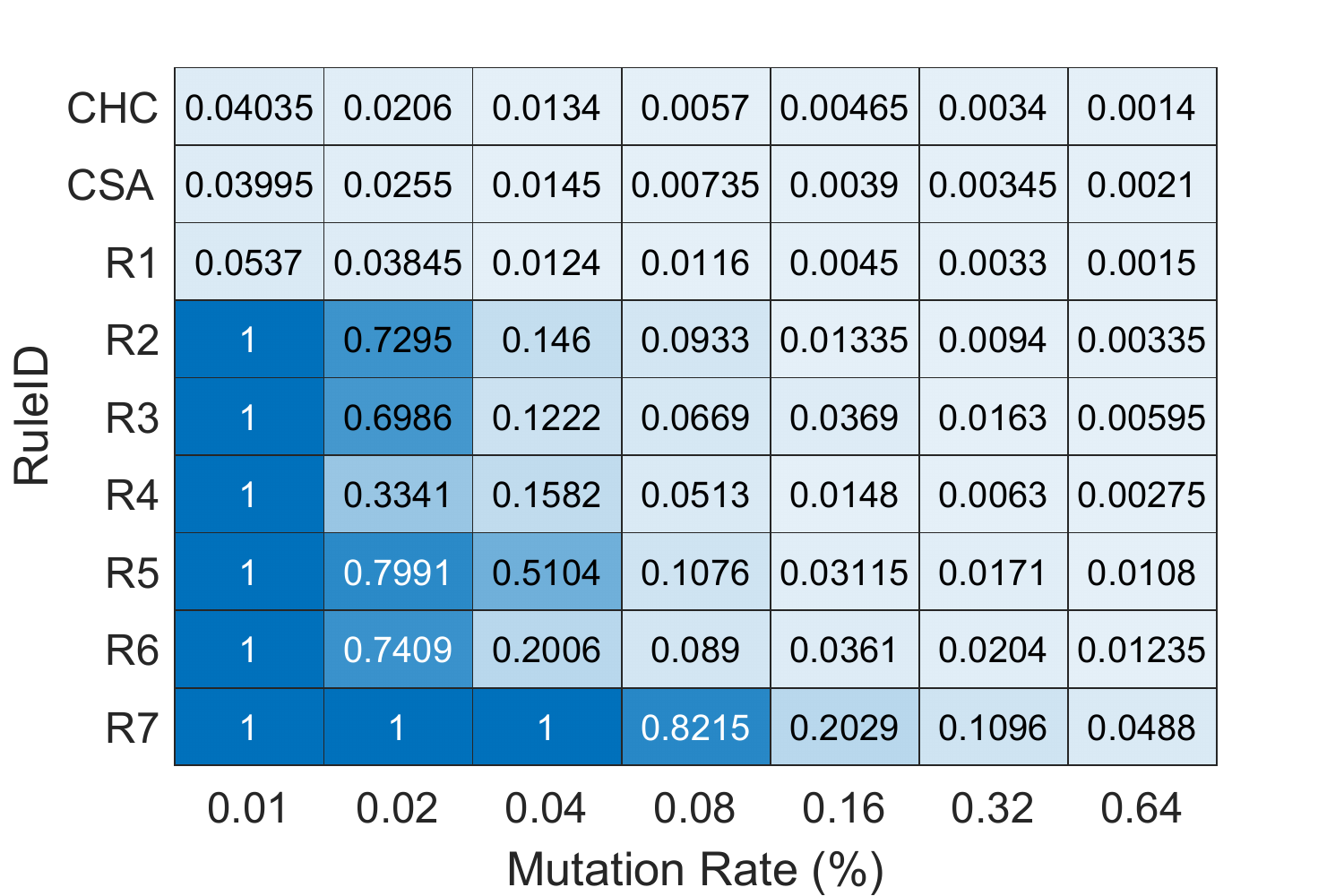}\label{fig:DEEinxSolution9}}
\subfloat[{\scriptsize \% Used slots (3$\times$3 nodes)}]{\includegraphics[width=0.45\columnwidth]{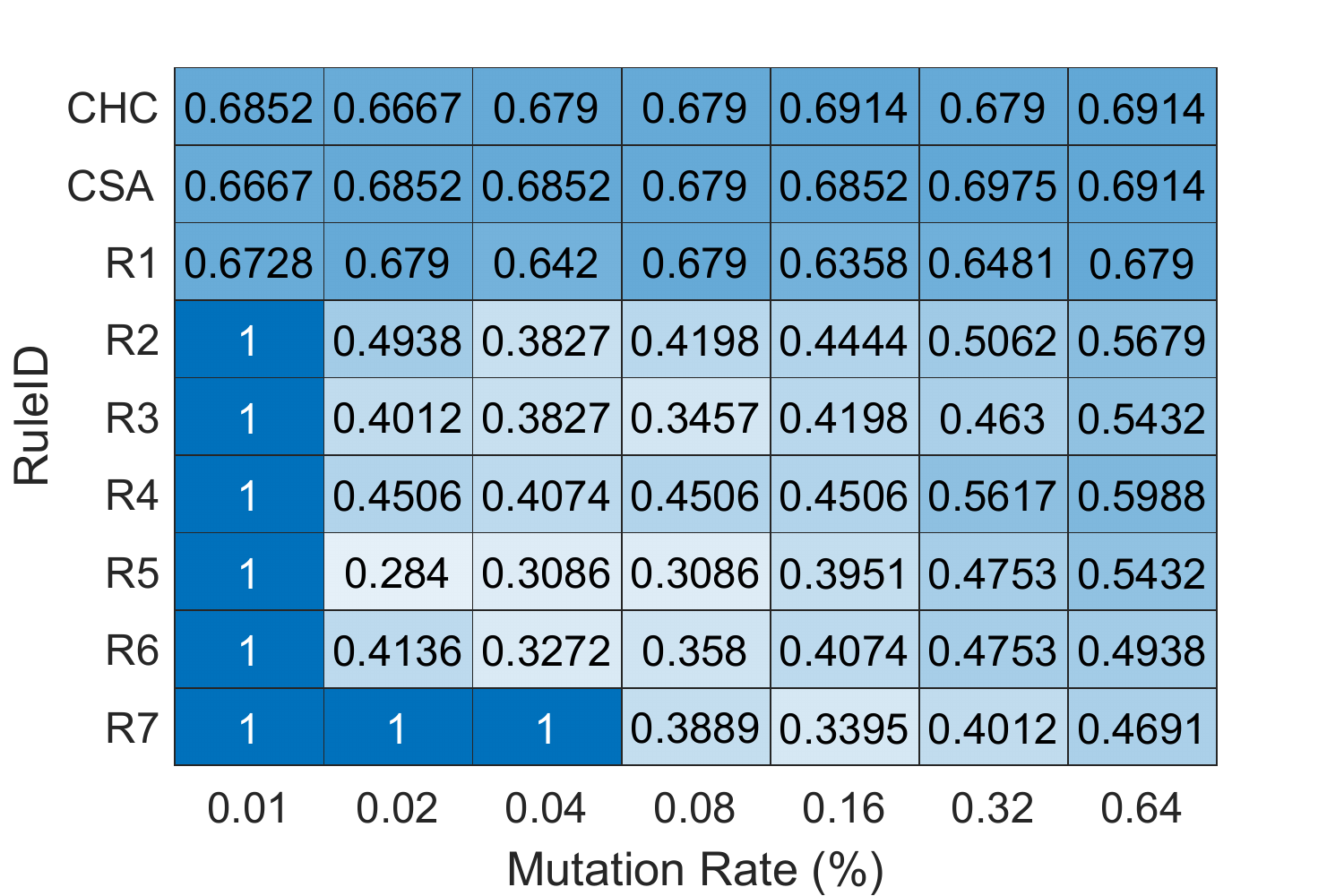}\label{fig:DEEusedSlots9}}

\subfloat[{\scriptsize \% Used resources (6$\times$6 nodes)}]{\includegraphics[width=0.45\columnwidth]{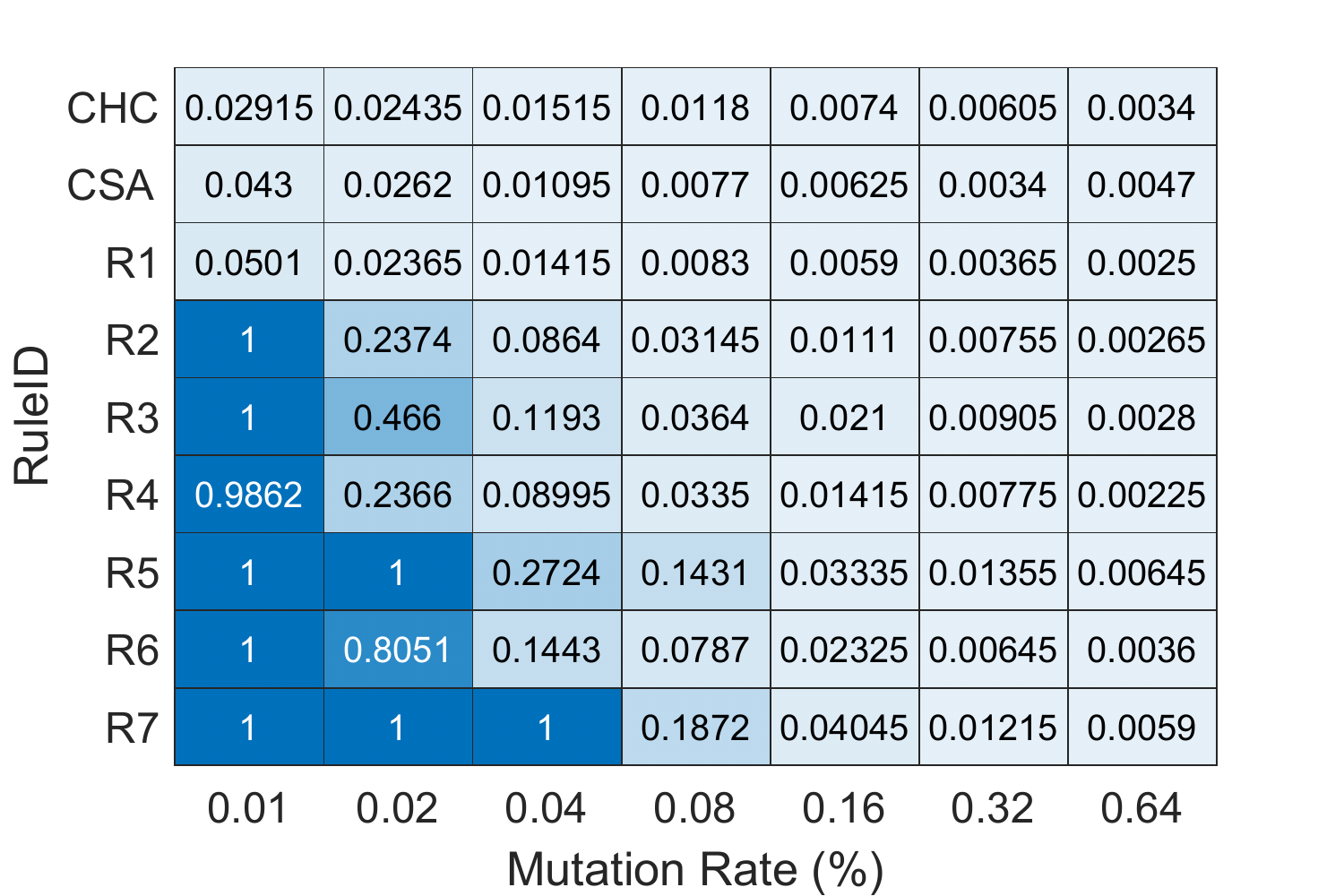}\label{fig:DEEinxSolution36}}
\subfloat[{\scriptsize \% Used slots (6$\times$6 nodes)}]{\includegraphics[width=0.45\columnwidth]{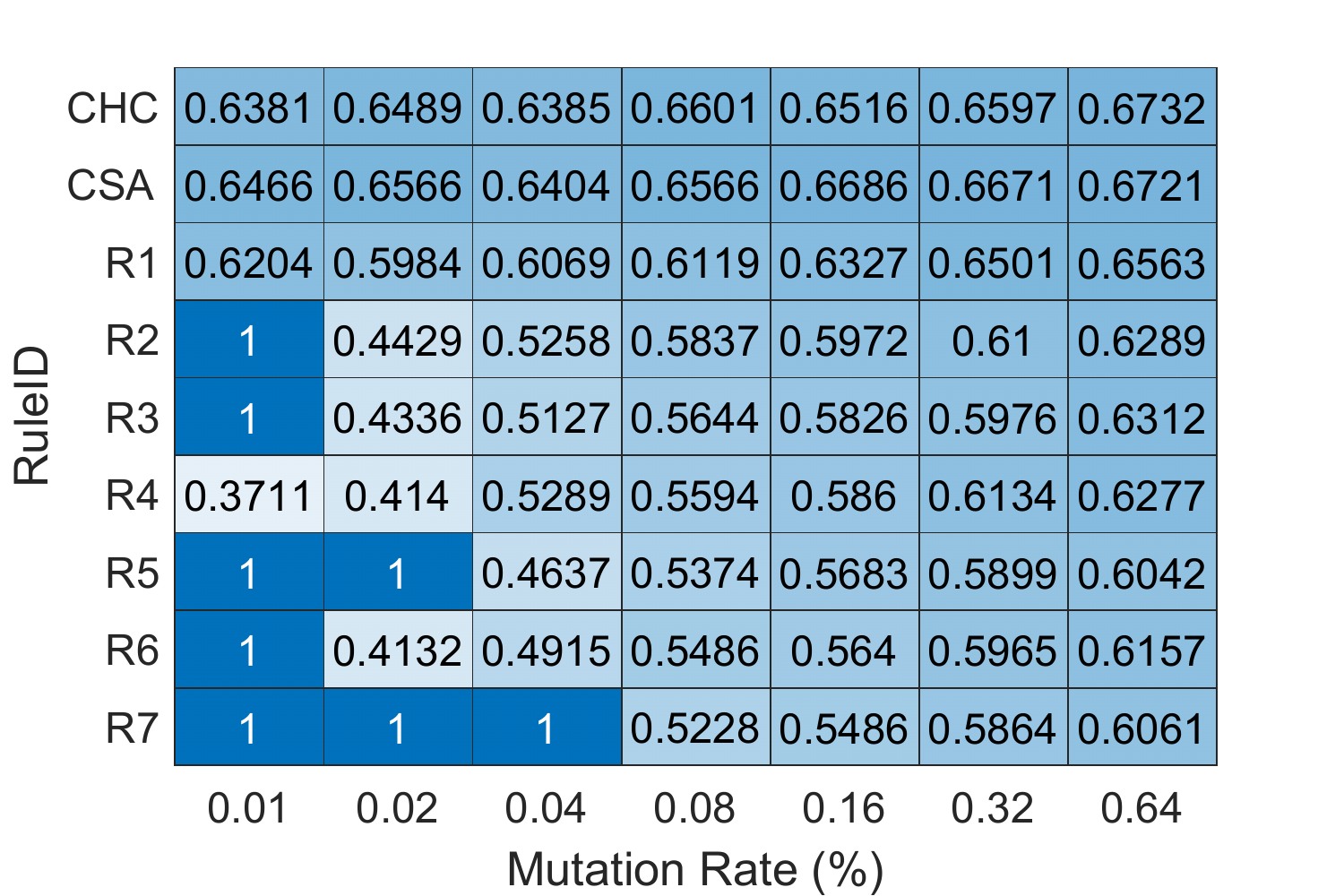}\label{fig:DEEusedSlots36}}

\subfloat[{\scriptsize \% Used resources (9$\times$9 nodes)}]{\includegraphics[width=0.45\columnwidth]{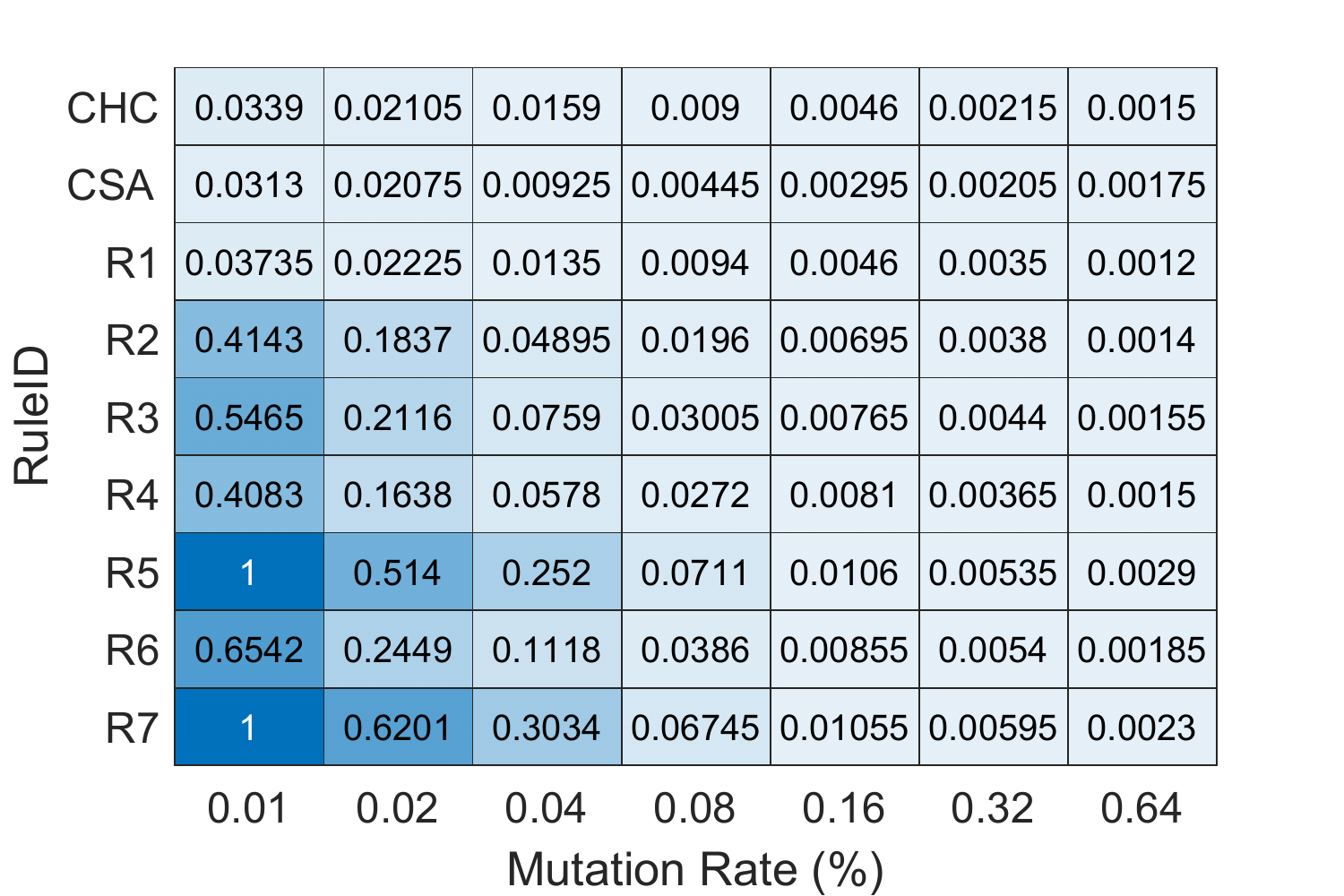}\label{fig:DEEinxSolution81}}
\subfloat[{\scriptsize \% Used slots (9$\times$9 nodes)}]{\includegraphics[width=0.45\columnwidth]{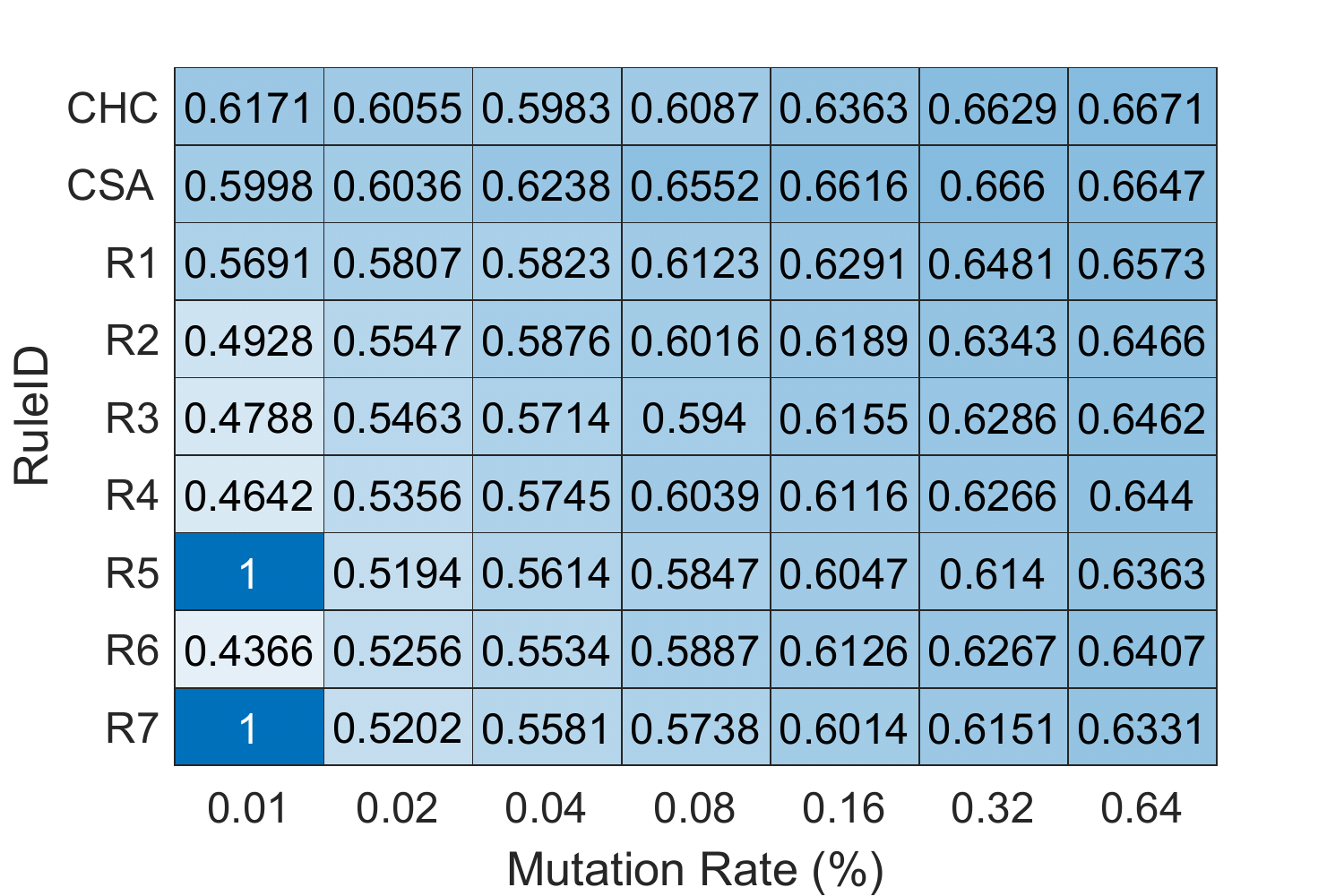}\label{fig:DEEusedSlots81}}
\end{subfigures}
\caption{Variation of the evolved TDMA MAC protocol performance for Centralized Hill Climbing (CHC), Centralized Simulated Annealing (CSA) and \ALGNAME~(\ALGABRV) with 7 different rules on grid networks with 3$\times$3, 6$\times$6 and 9$\times$9 nodes.}\label{fig:results-grid-nets}
\end{figure}

\begin{figure}[!ht]
\begin{subfigures}
\subfloat[{\scriptsize \% Used resources ($cd$=0.8, $cp$=1)}]{\includegraphics[width=0.35\columnwidth]{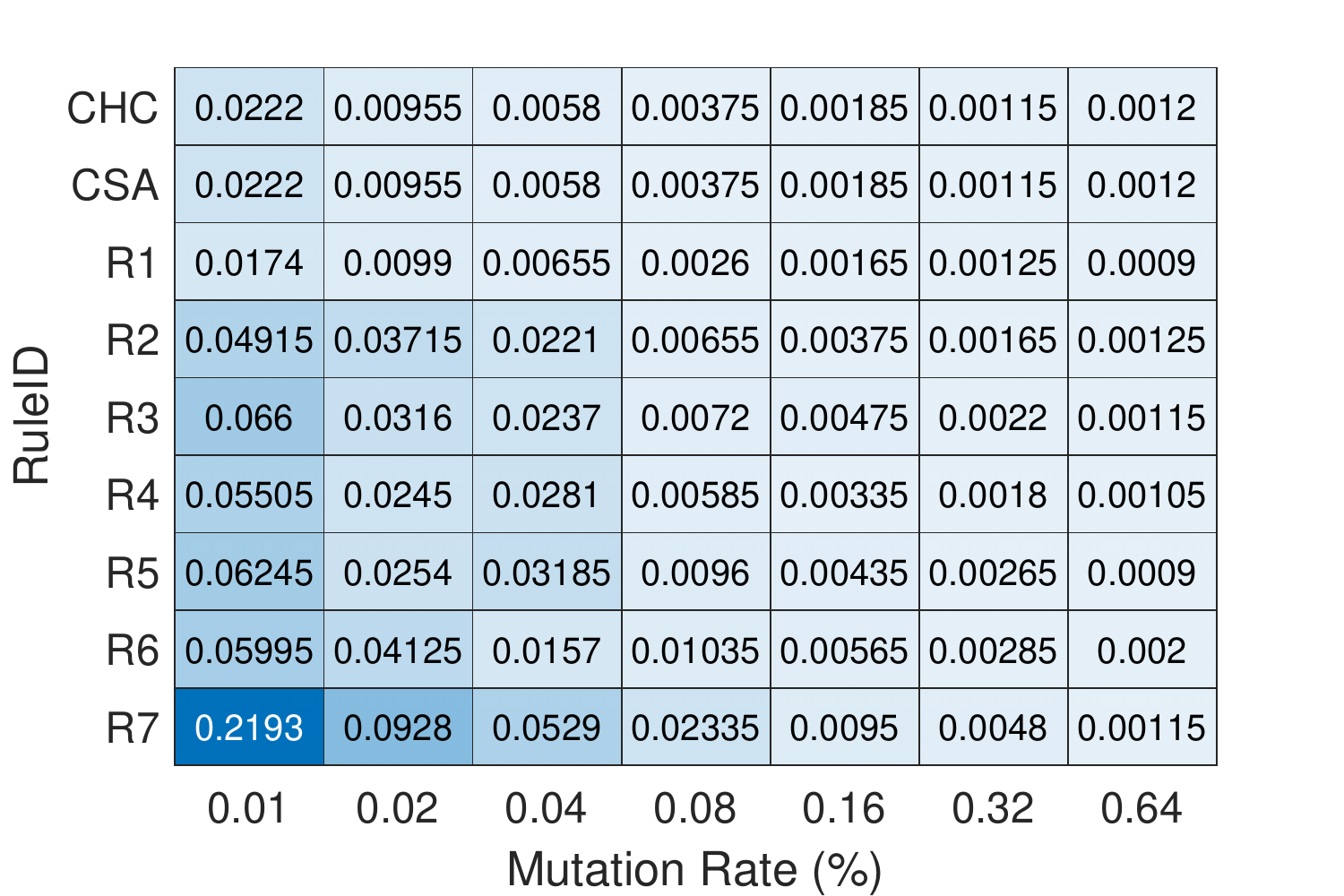}\label{fig:heatMap9nodesSolutionRcd08cp1}}
\subfloat[{\scriptsize \% Used slots ($cd$=0.8, $cp$=1)}]{\includegraphics[width=0.35\columnwidth]{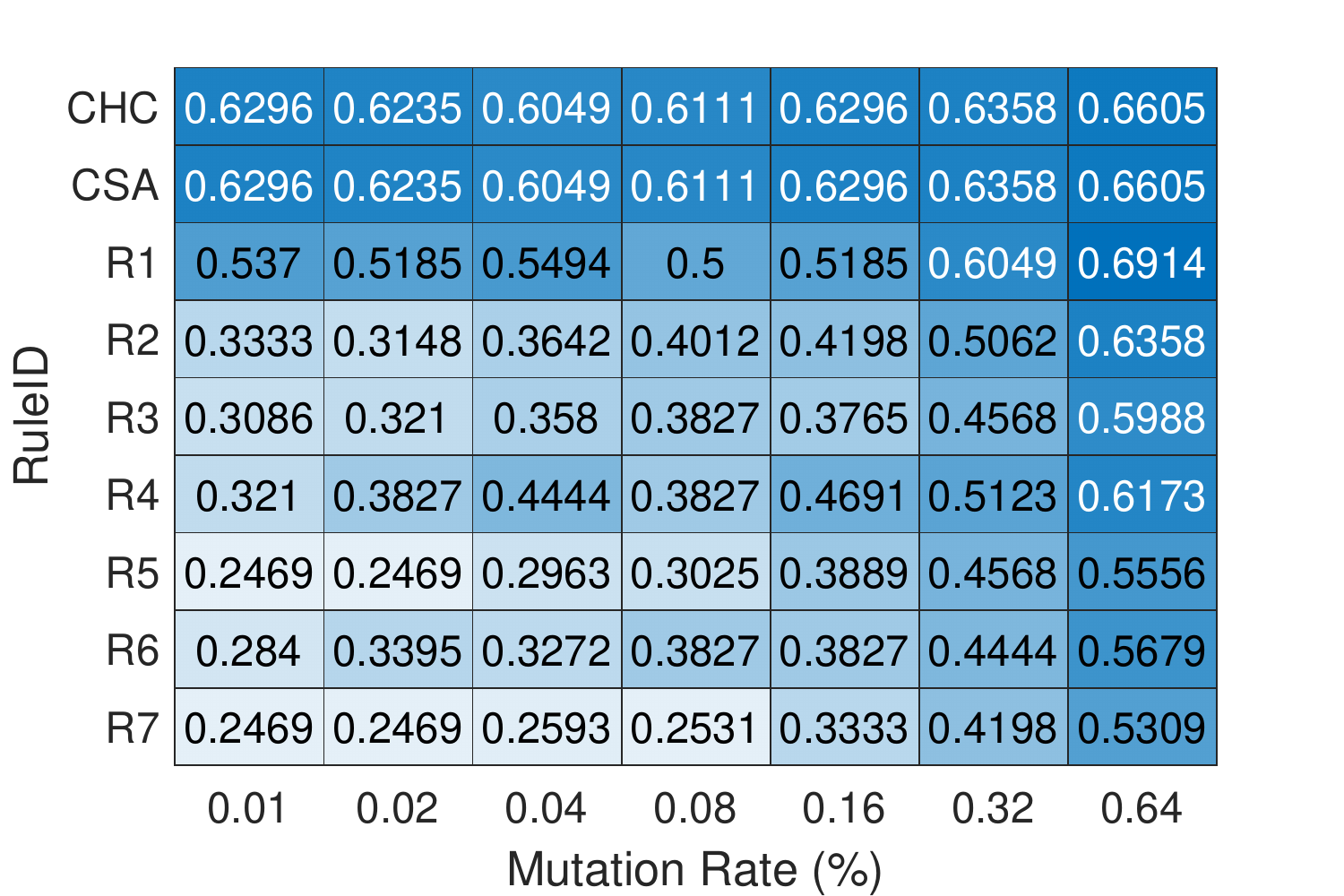}\label{fig:heatMap9nodesSlotsRcd08cp1}}

\subfloat[{\scriptsize \% Used resources ($cd$=0.8, $cp$=0.5)}]{\includegraphics[width=0.35\columnwidth]{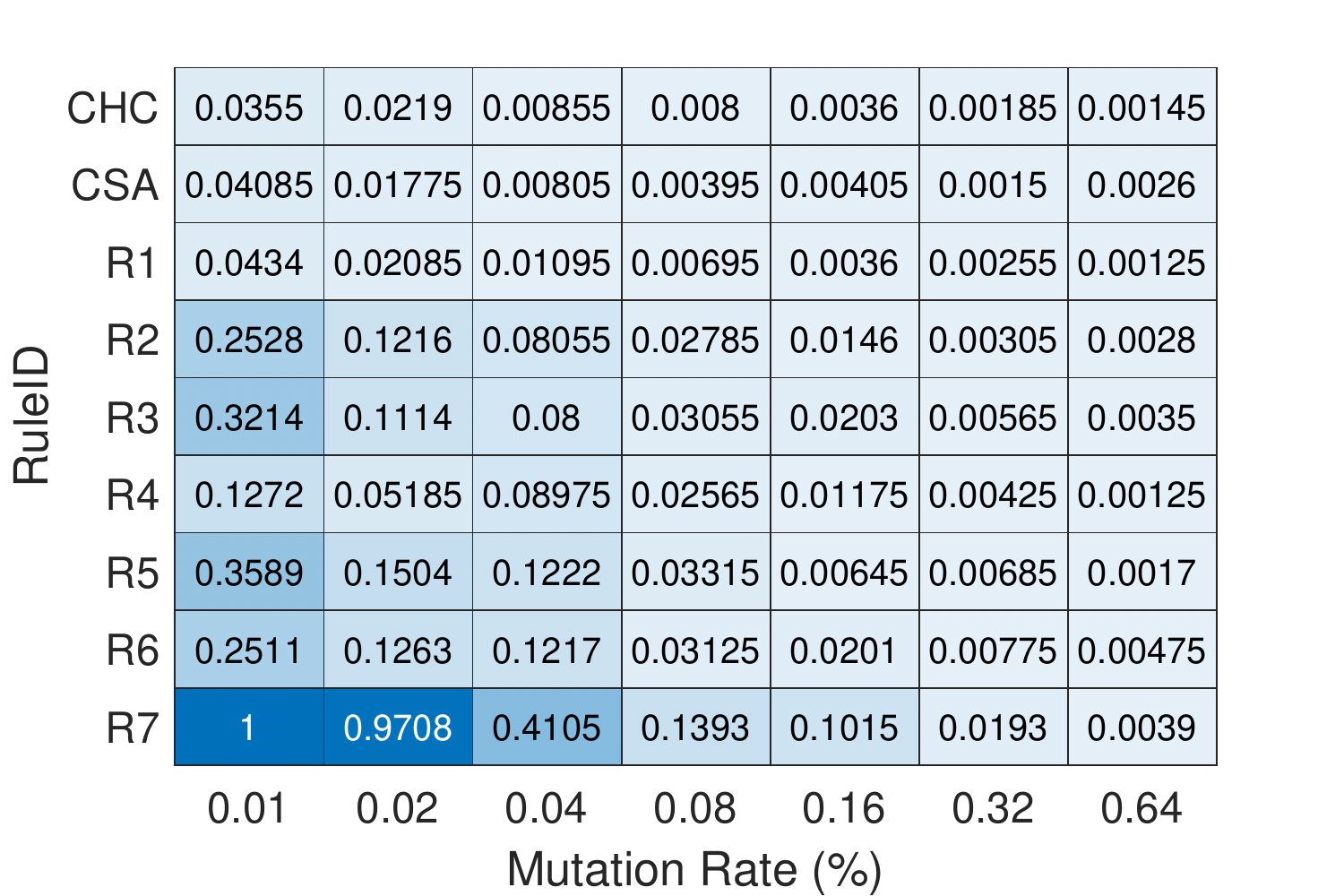}\label{fig:heatMap9nodesSolutionRcd08cp05}}
\subfloat[{\scriptsize \% Used slots ($cd$=0.8, $cp$=0.5)}]{\includegraphics[width=0.35\columnwidth]{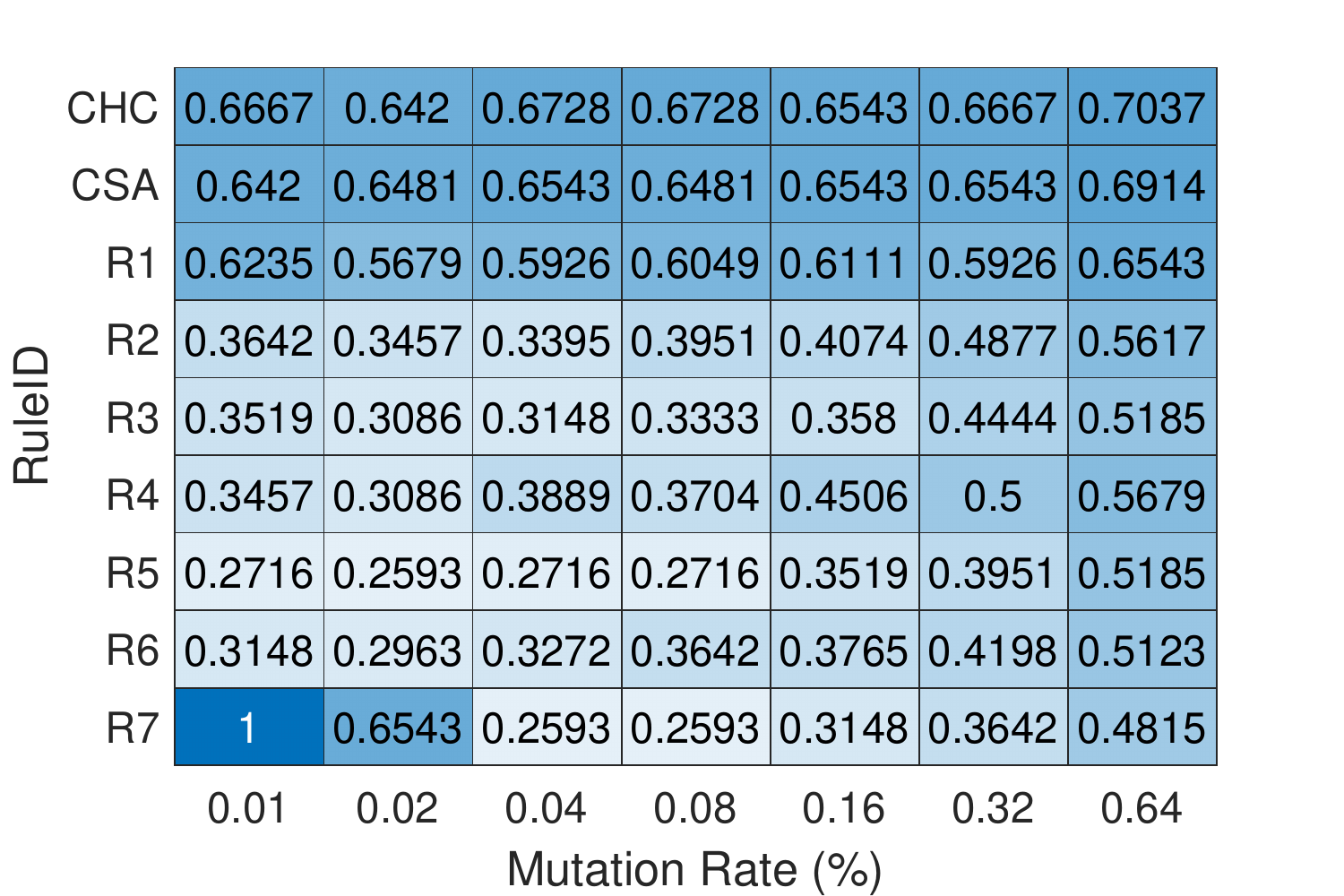}\label{fig:heatMap36nodesSlotsRcd08cp05}}

\subfloat[{\scriptsize \% Used resources ($cd$=0.8, $cp$=0.25)}]{\includegraphics[width=0.35\columnwidth]{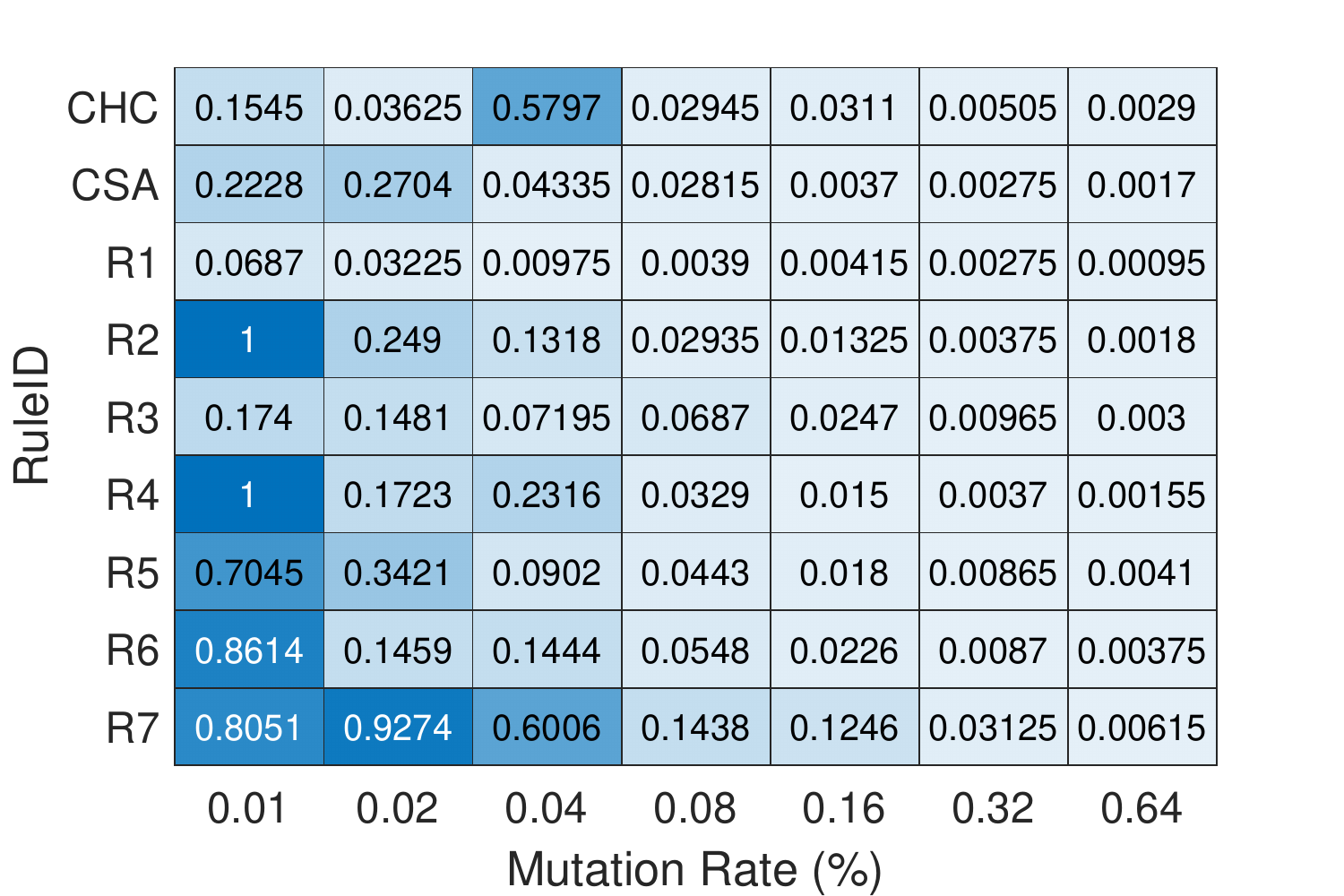}\label{fig:heatMap9nodesSolutionRcd08cp025}}
\subfloat[{\scriptsize \% Used slots ($cd$=0.8, $cp$=0.25)}]{\includegraphics[width=0.35\columnwidth]{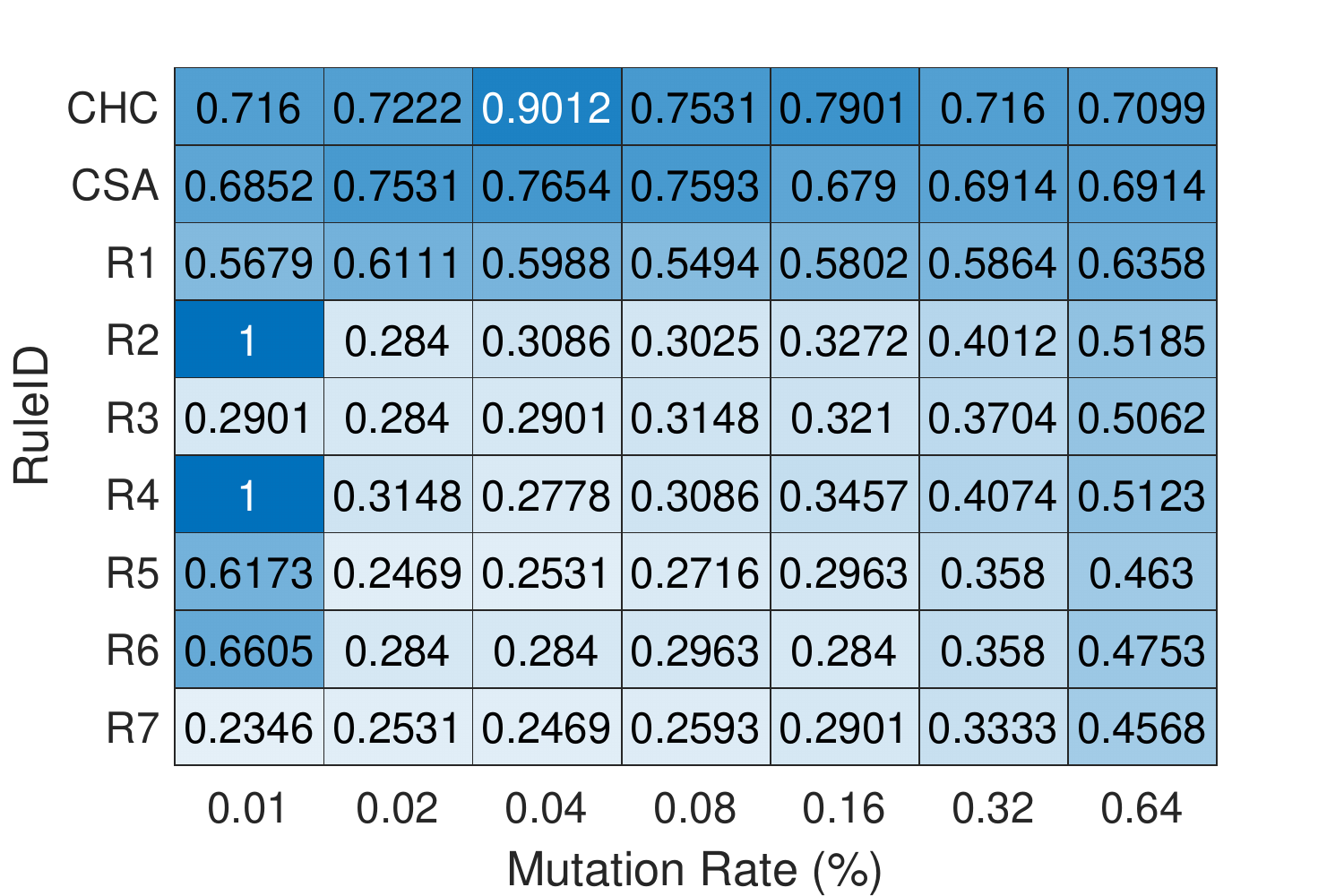}\label{fig:heatMap36nodesSlotsRcd08cp1}}

\subfloat[{\scriptsize \% Used resources ($cd$=0.8, $cp$=0.125)}]{\includegraphics[width=0.35\columnwidth]{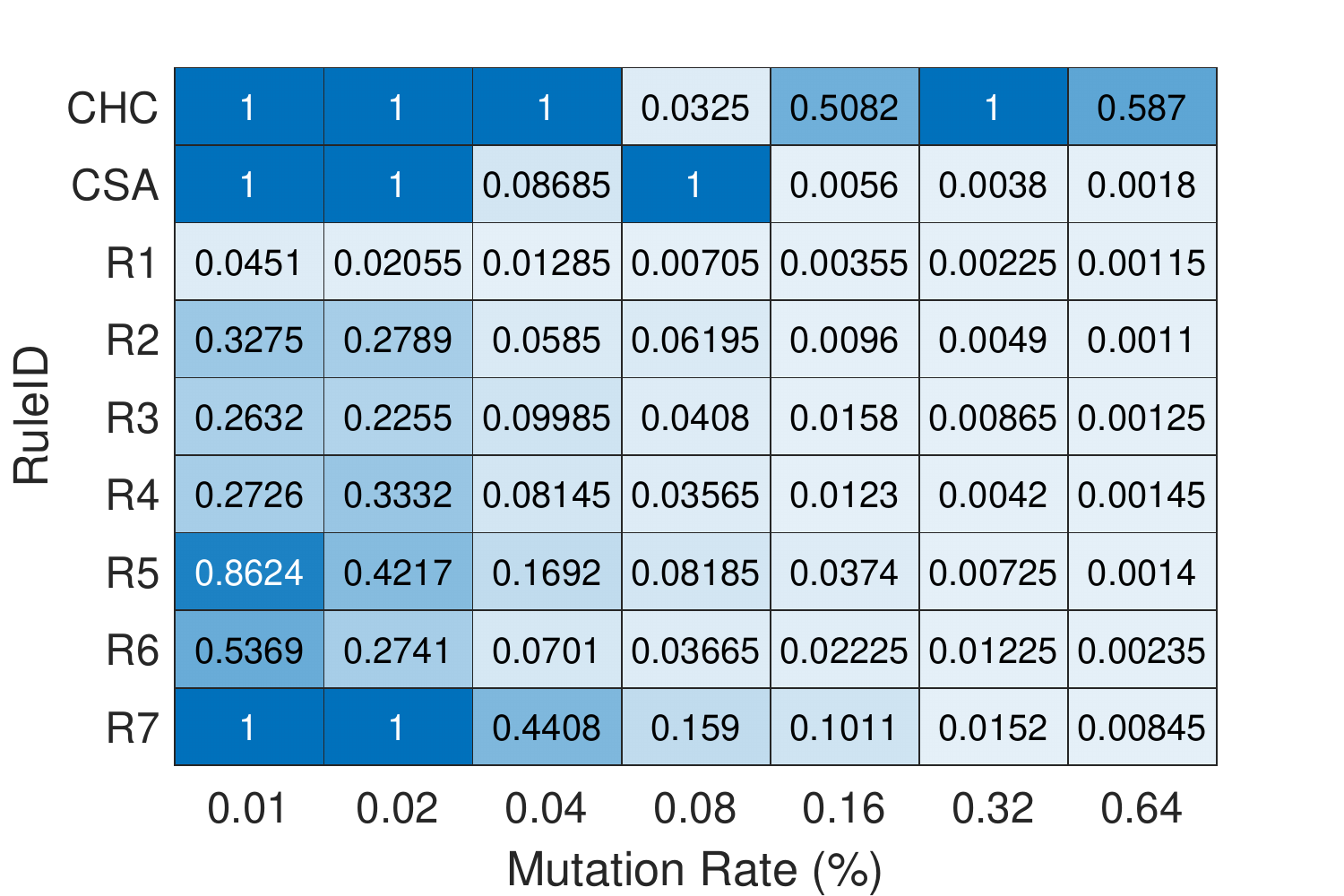}\label{fig:heatMap9nodesSolutionRcd08cp0125}}
\subfloat[{\scriptsize \% Used slots ($cd$=0.8, $cp$=0.125)}]{\includegraphics[width=0.35\columnwidth]{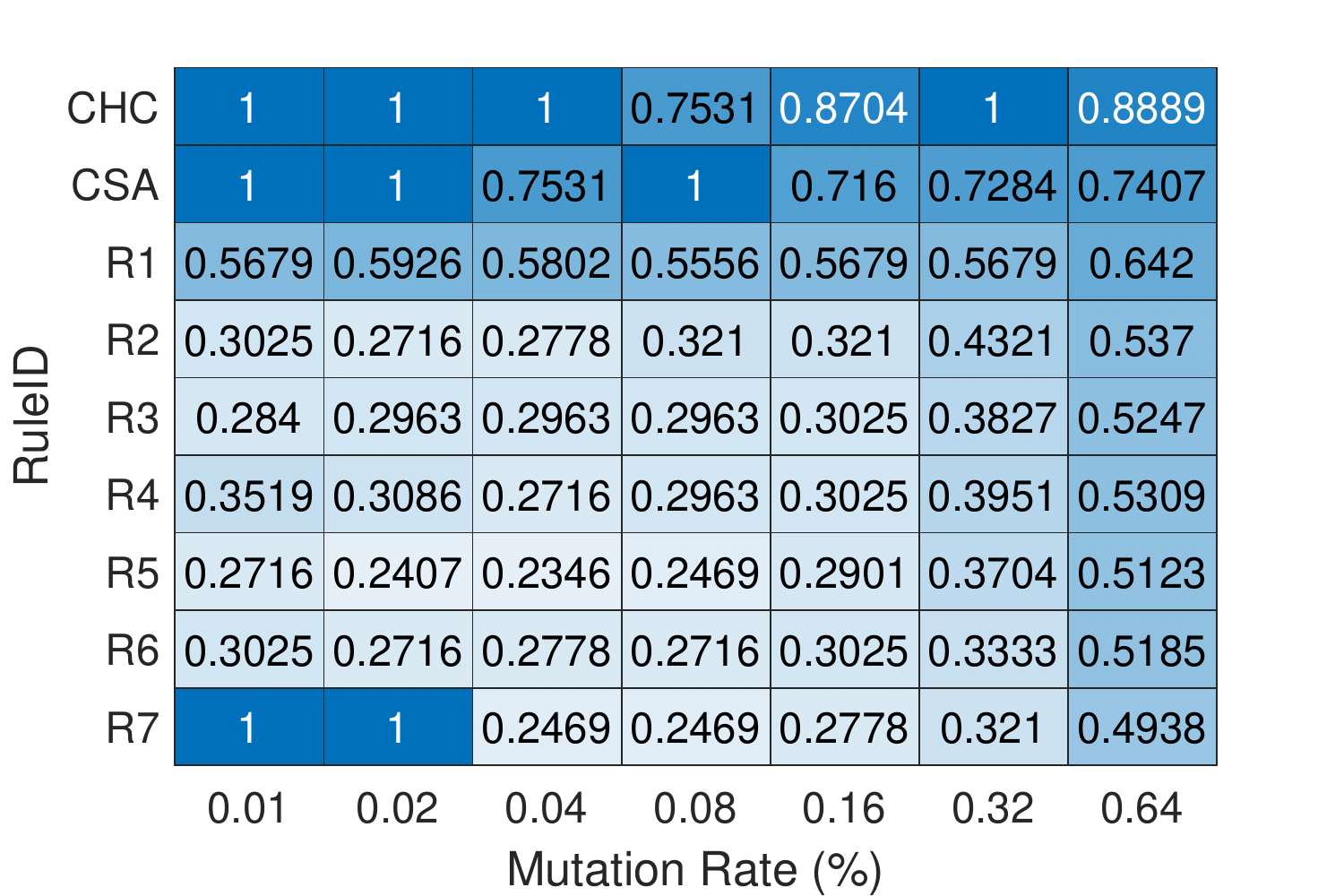}\label{fig:heatMap9nodesSlotsRcd08cp0125}}
\end{subfigures}
\caption{Variation of the evolved TDMA MAC protocol performance for Centralized Hill Climbing (CHC), Centralized Simulated Annealing (CSA) and \ALGNAME~(\ALGABRV) with 7 different rules on random networks with $9$ nodes and various levels of connection distance ($cd$) and connection probability ($cp$).} \label{fig:results-random-nets9}
\end{figure}

\begin{figure}[!ht]
\vspace{-3cm}
\begin{subfigures}
\subfloat[{\scriptsize \% Used resources ($cd$=0.5, $cp$=1)}]{\includegraphics[width=0.4\columnwidth]{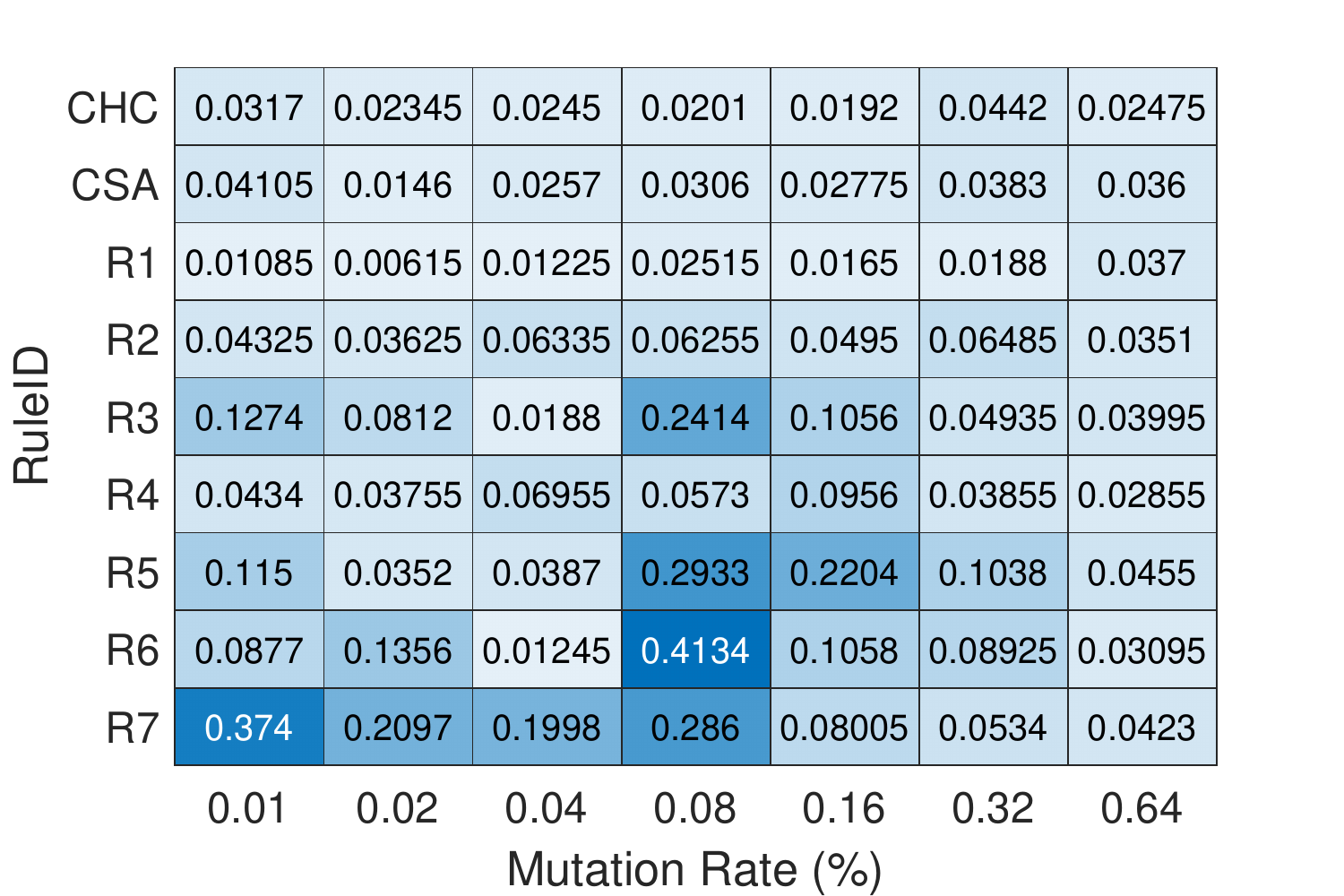}\label{fig:heatMap36nodesSolutionRcd05cp1}}
\subfloat[{\scriptsize \% Used slots ($cd$=0.5, $cp$=1)}]{\includegraphics[width=0.4\columnwidth]{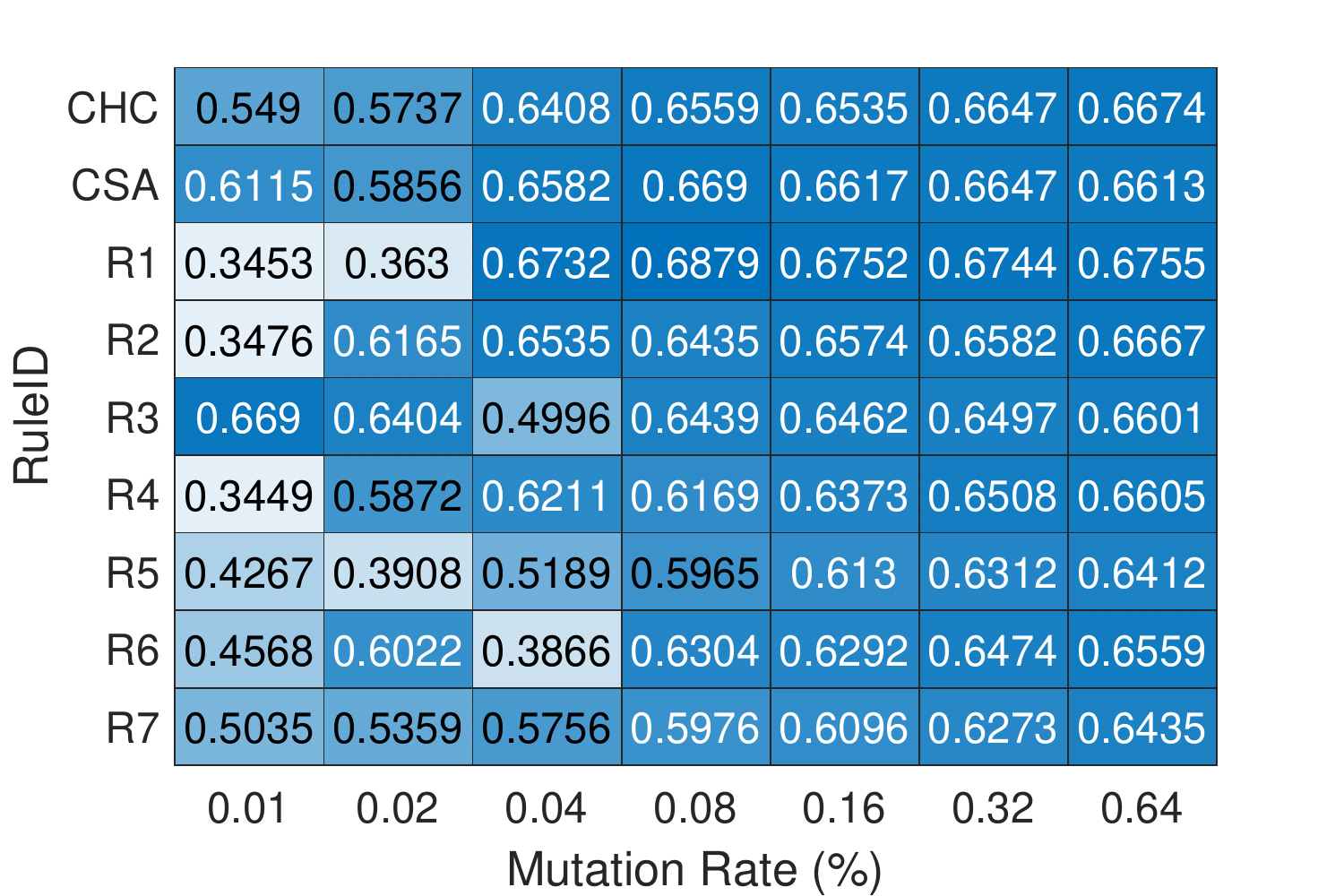}\label{fig:heatMap36nodesSlotsRcd05cp1}}

\subfloat[{\scriptsize \% Used resources ($cd$=0.5, $cp$=0.5)}]{\includegraphics[width=0.4\columnwidth]{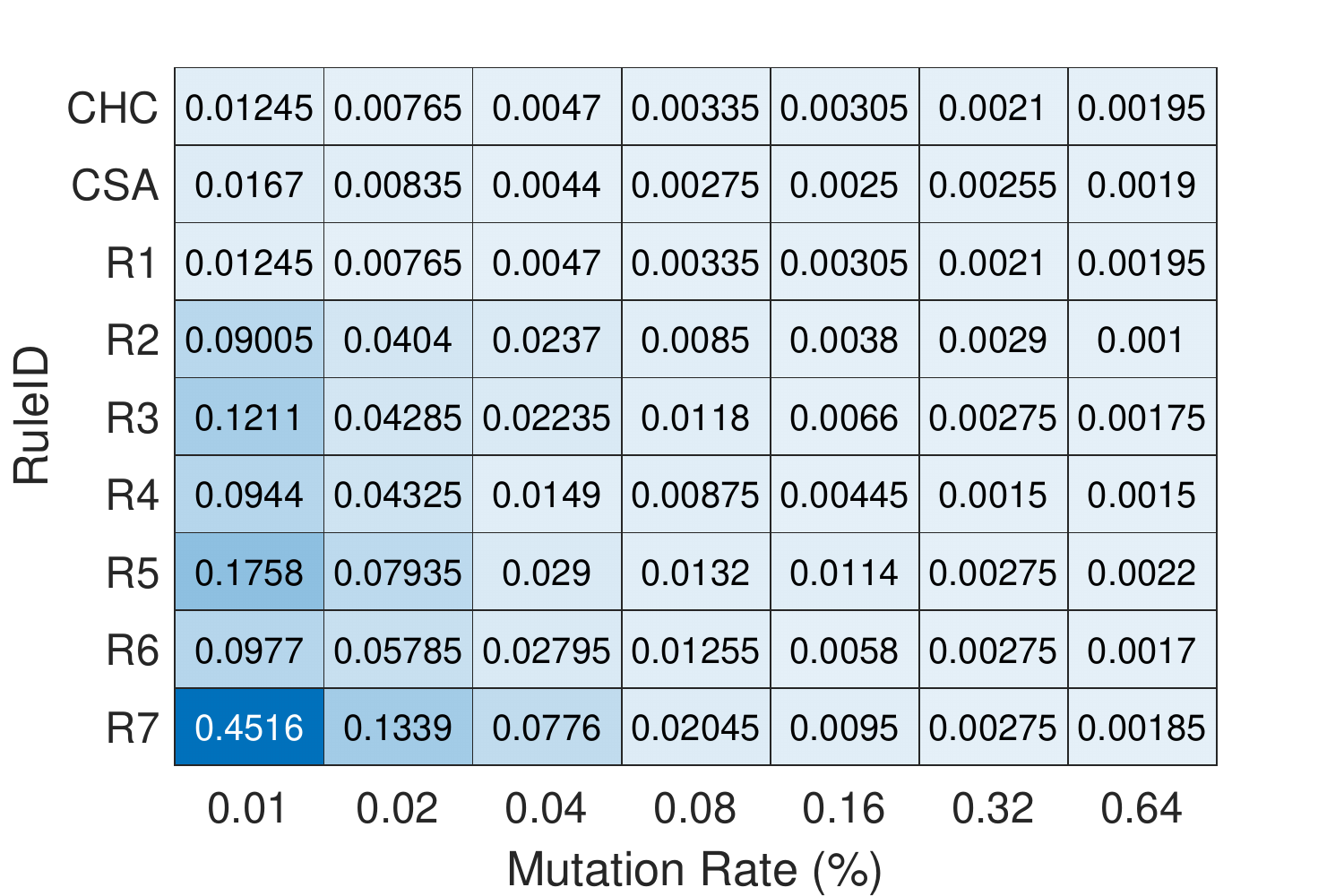}\label{fig:heatMap36nodesSolutionRcd05cp05}}
\subfloat[{\scriptsize \% Used slots ($cd$=0.5, $cp$=0.5)}]{\includegraphics[width=0.4\columnwidth]{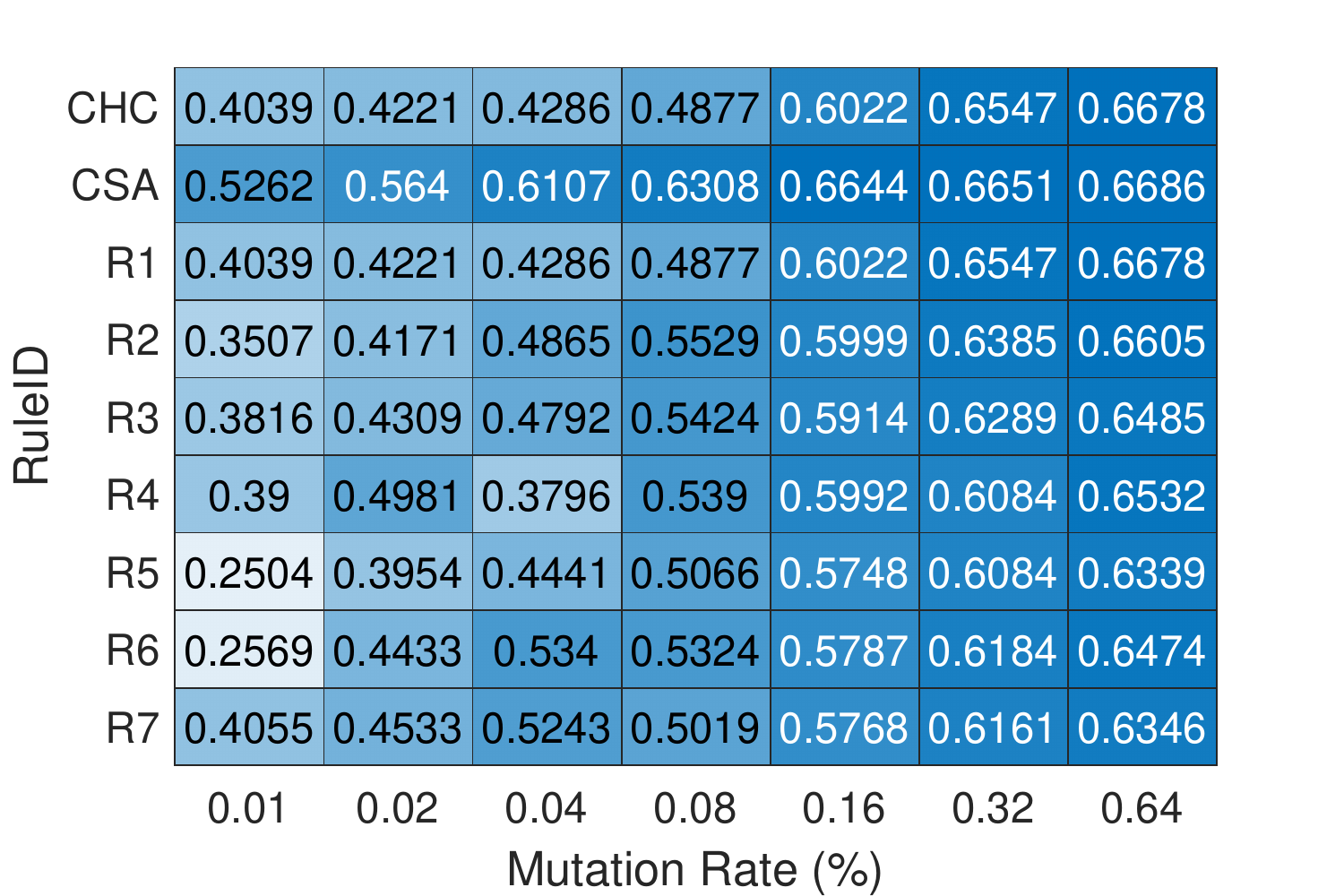}\label{fig:heatMap36nodesSlotsRcd05cp05}}

\subfloat[{\scriptsize \% Used resources ($cd$=0.5, $cp$=0.25)}]{\includegraphics[width=0.4\columnwidth]{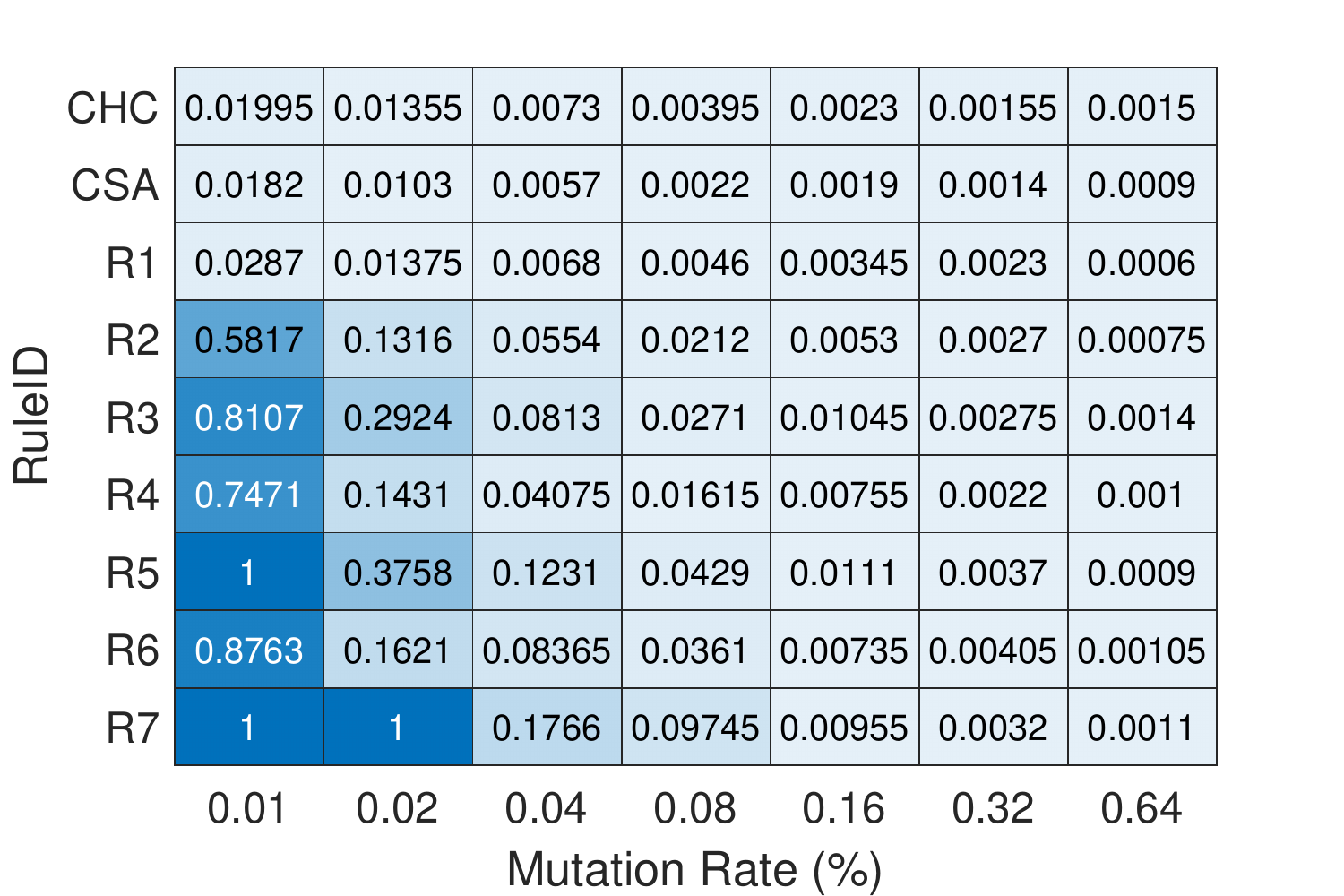}\label{fig:heatMap36nodesSolutionRcd05cp025}}
\subfloat[{\scriptsize \% Used slots ($cd$=0.5, $cp$=0.25)}]{\includegraphics[width=0.4\columnwidth]{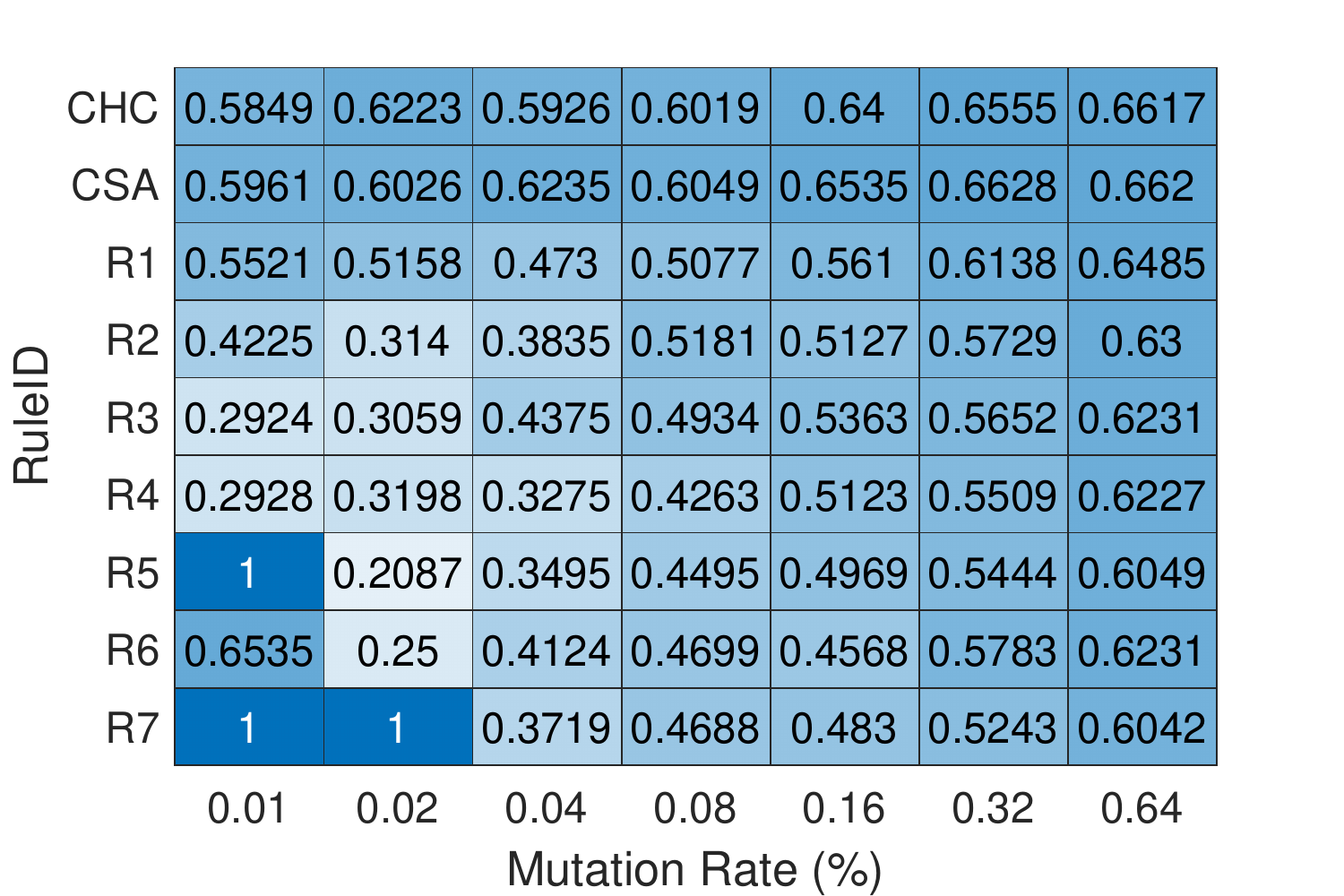}\label{fig:heatMap36nodesSlotsRcd05cp025}}

\subfloat[{\scriptsize \% Used resources ($cd$=0.5, $cp$=0.125)}]{\includegraphics[width=0.4\columnwidth]{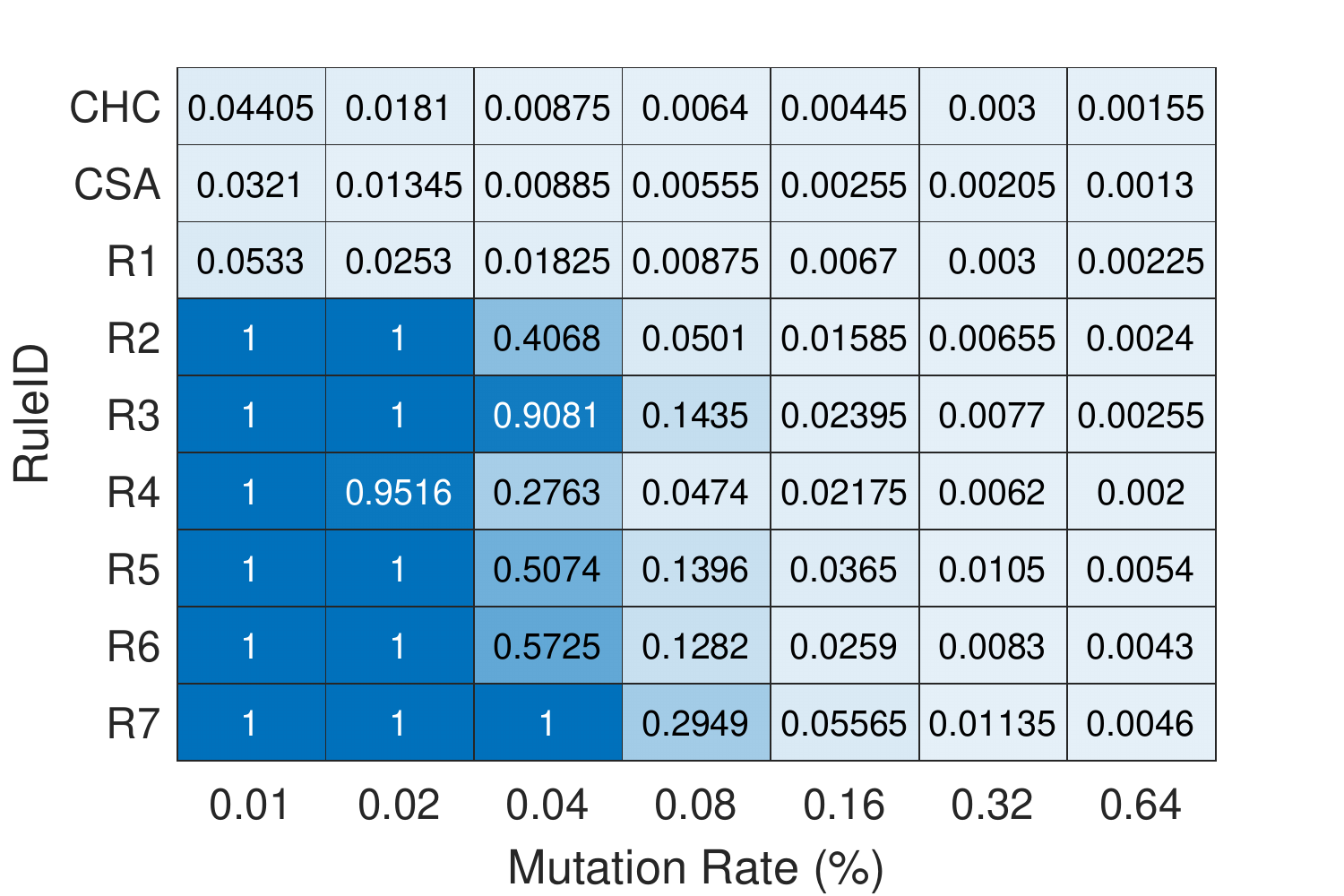}\label{fig:heatMap36nodesSolutionRcd05cp0125}}
\subfloat[{\scriptsize \% Used slots ($cd$=0.5, $cp$=0.125)}]{\includegraphics[width=0.4\columnwidth]{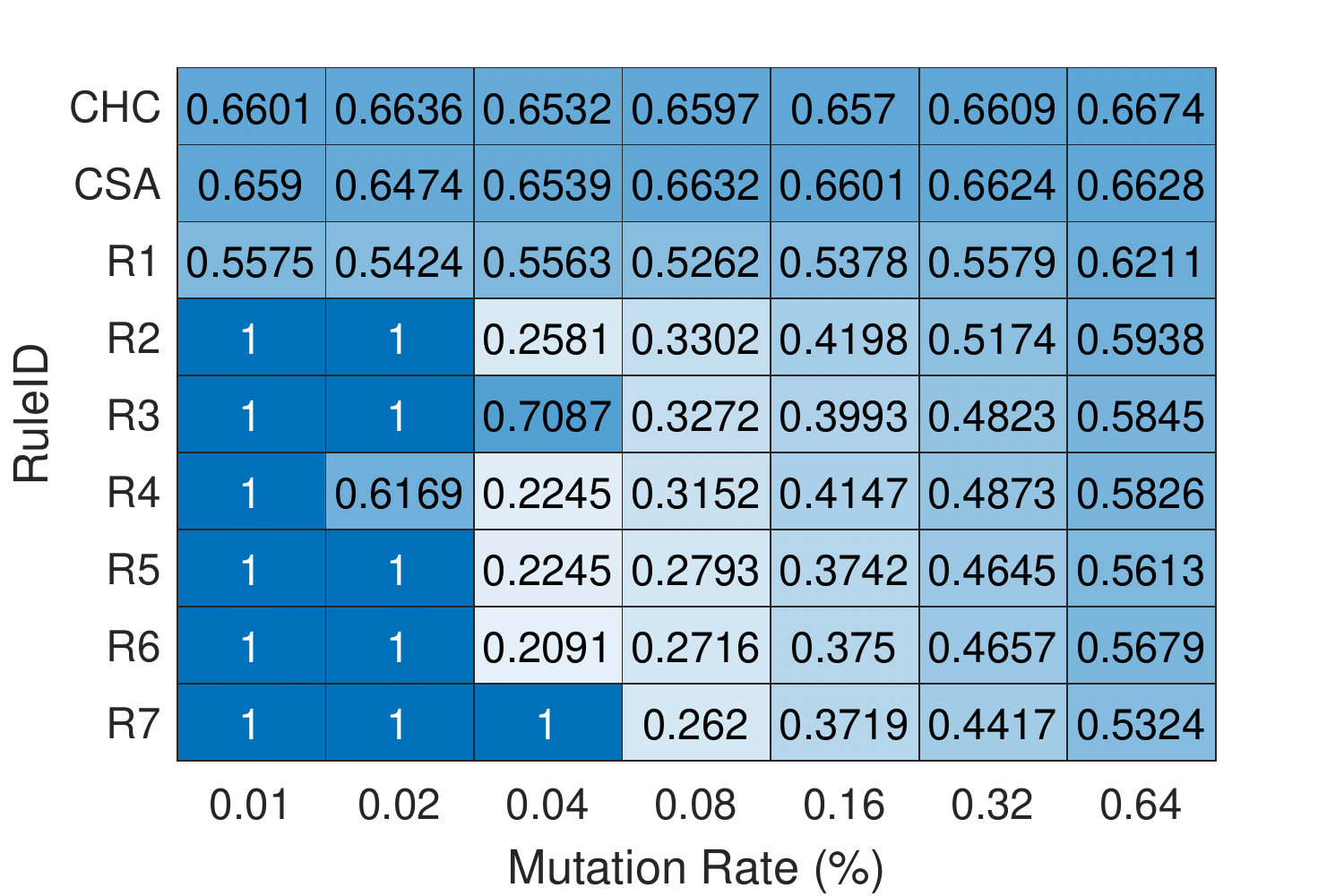}\label{fig:heatMap36nodesSlotsRcd05cp0125}}
\end{subfigures}
\caption{Variation of the evolved TDMA MAC protocol performance for Centralized Hill Climbing (CHC), Centralized Simulated Annealing (CSA) and \ALGNAME~(\ALGABRV) with 7 different rules on random networks with $36$ nodes and various levels of connection distance ($cd$) and connection probability ($cp$).} \label{fig:results-random-nets36}
\end{figure}

\begin{figure}[!t]
\vspace{-3cm}
\begin{subfigures}
\subfloat[{\scriptsize \% Used resources ($cd$=0.3, $cp$=1)}]{\includegraphics[width=0.4\columnwidth]{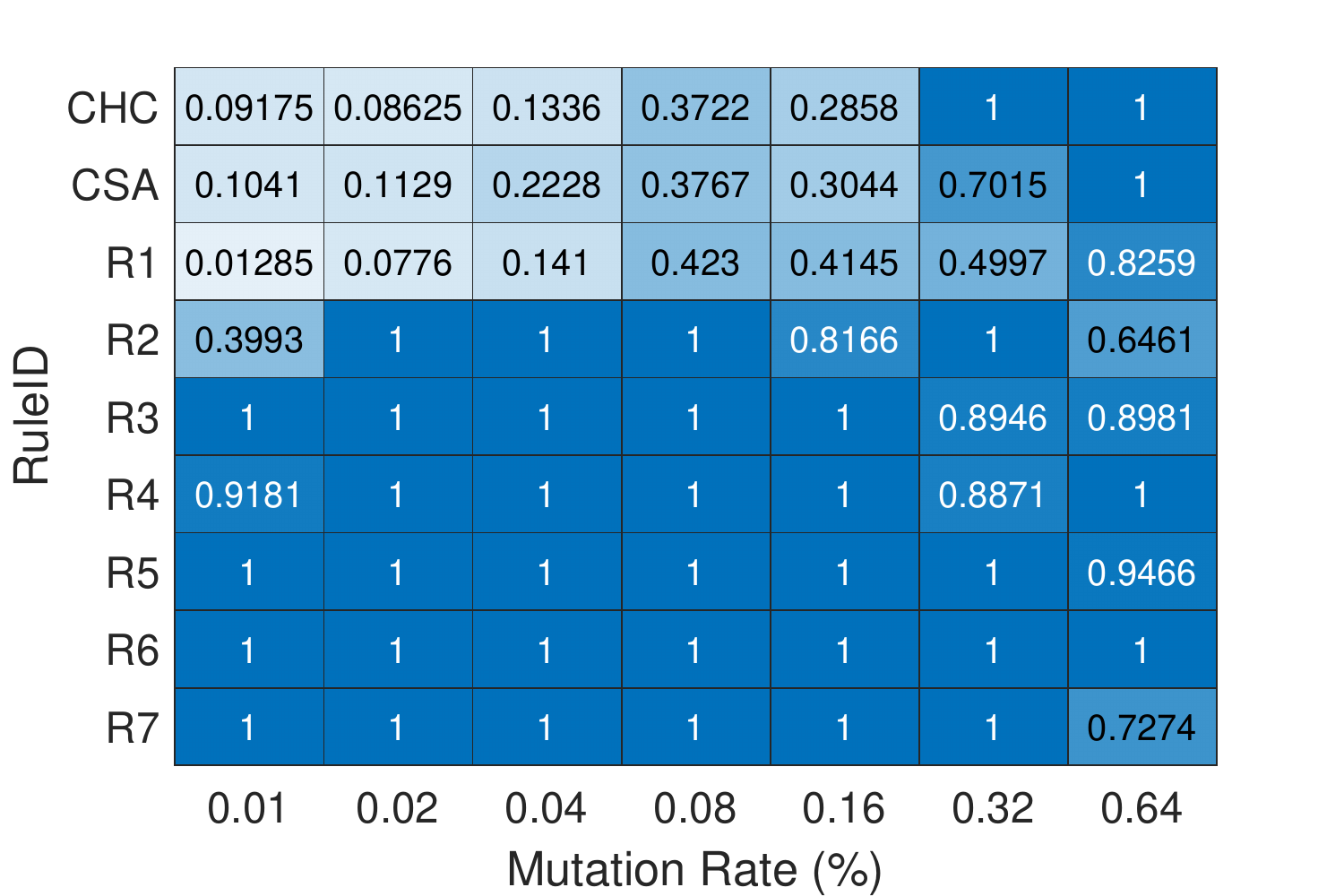}\label{fig:heatMap81nodesSolutionRcd03cp1}}
\subfloat[{\scriptsize \% Used slots ($cd$=0.3, $cp$=1)}]{\includegraphics[width=0.4\columnwidth]{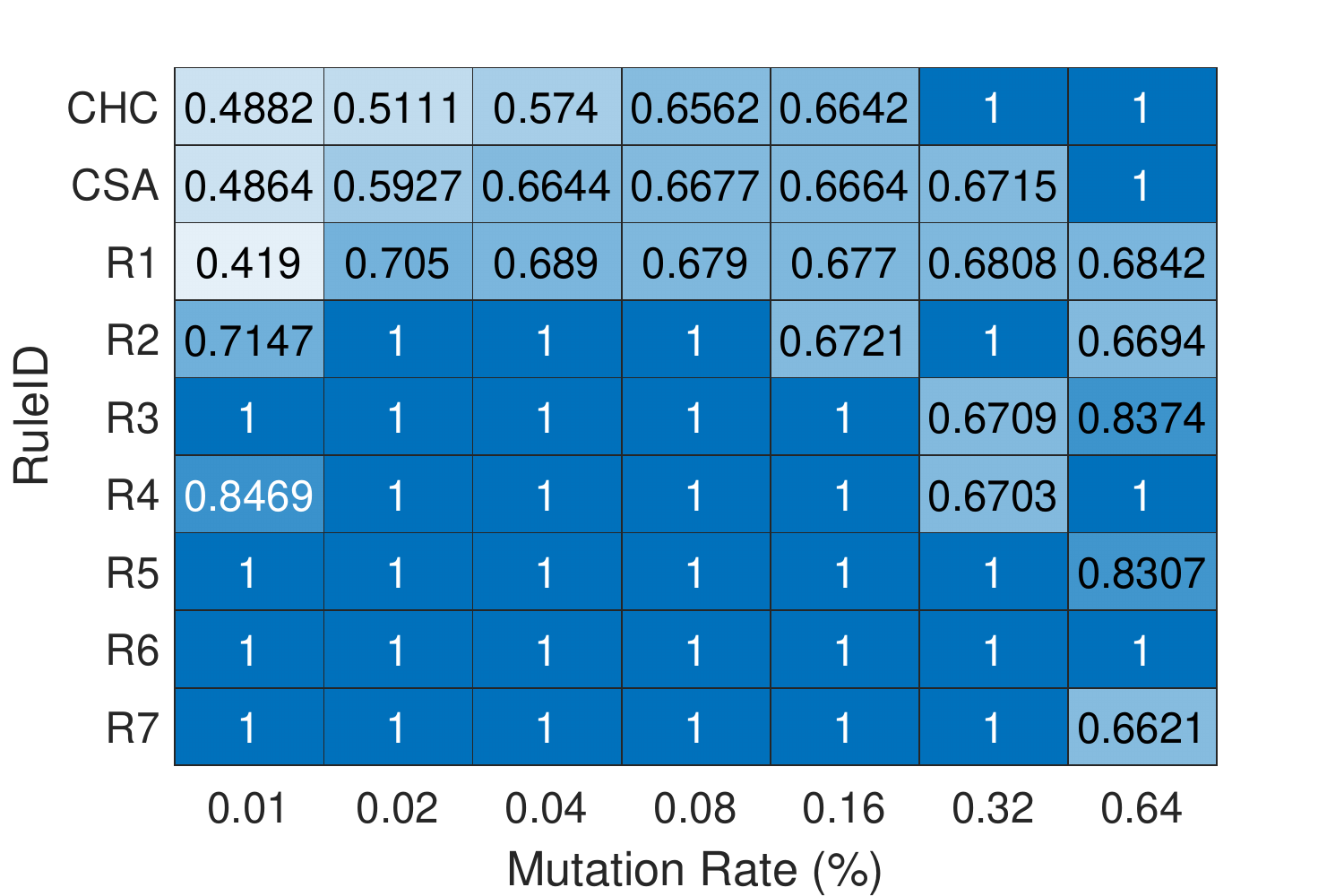}\label{fig:heatMap81nodesSlotsRcd03cp1}}

\subfloat[{\scriptsize \% Used resources ($cd$=0.3, $cp$=0.5)}]{\includegraphics[width=0.4\columnwidth]{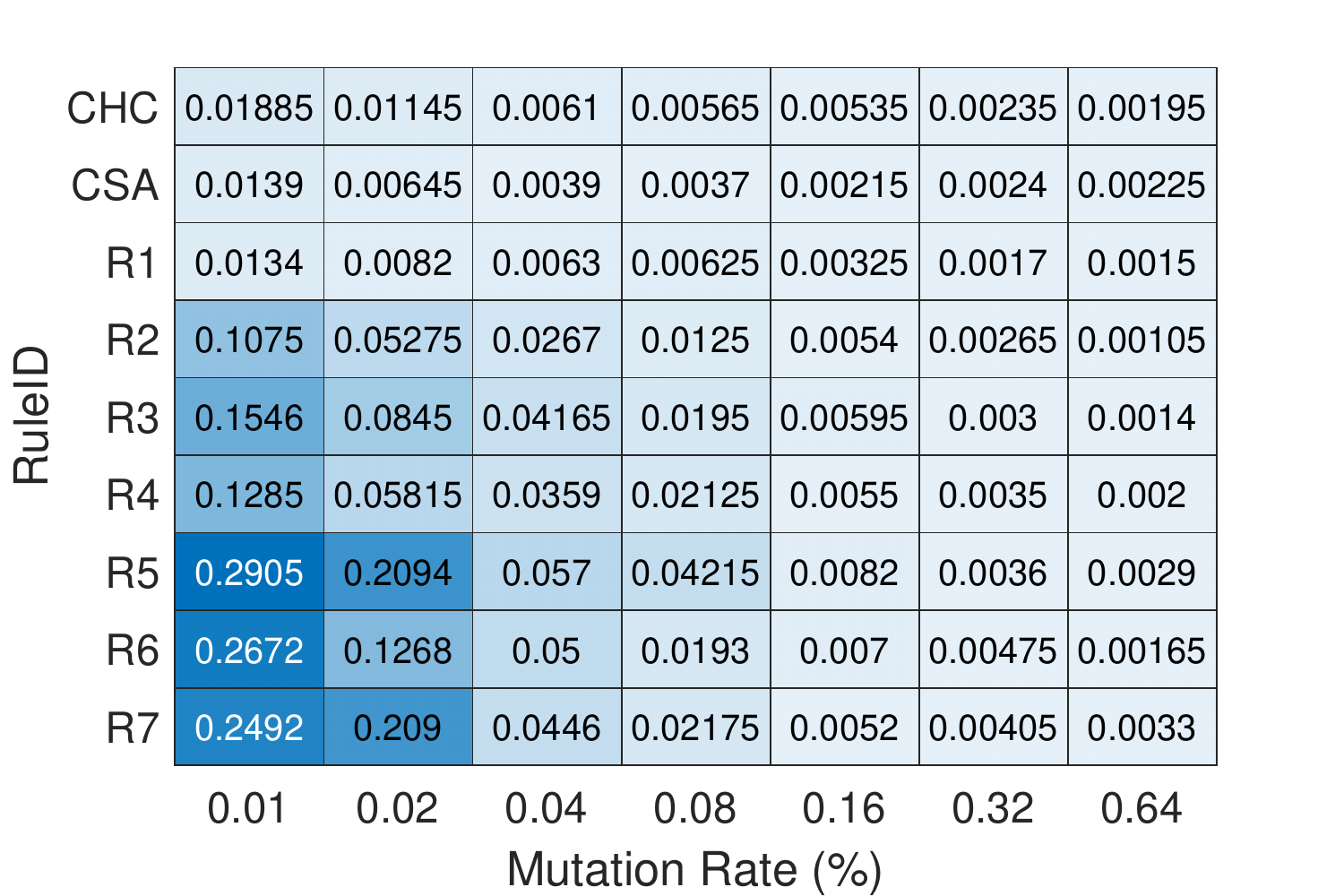}\label{fig:heatMap81nodesSolutionRcd03cp05}}
\subfloat[{\scriptsize \% Used slots ($cd$=0.3, $cp$=0.5)}]{\includegraphics[width=0.4\columnwidth]{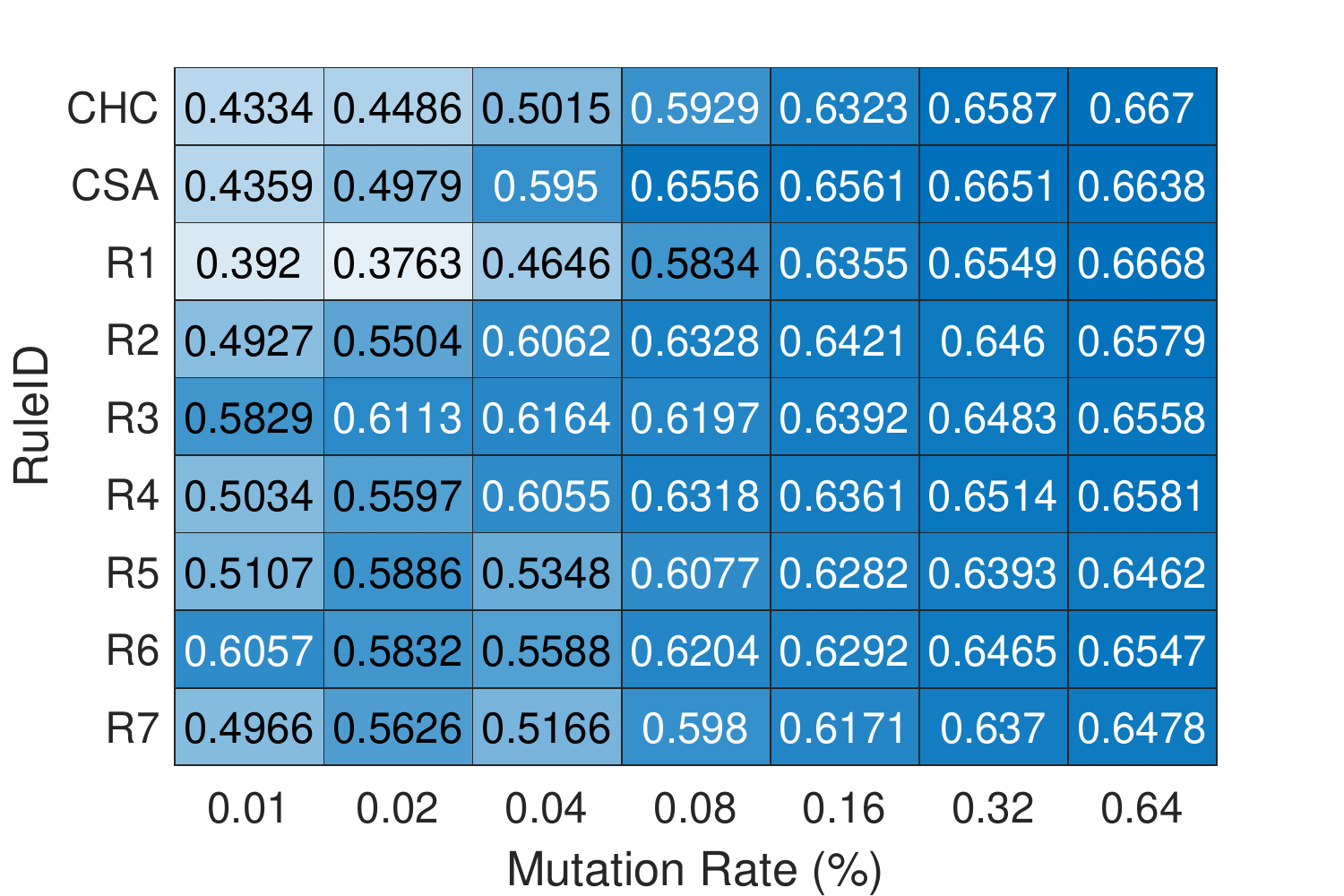}\label{fig:heatMap81nodesSlotsRcd03cp05}}

\subfloat[{\scriptsize \% Used resources ($cd$=0.3, $cp$=0.25)}]{\includegraphics[width=0.4\columnwidth]{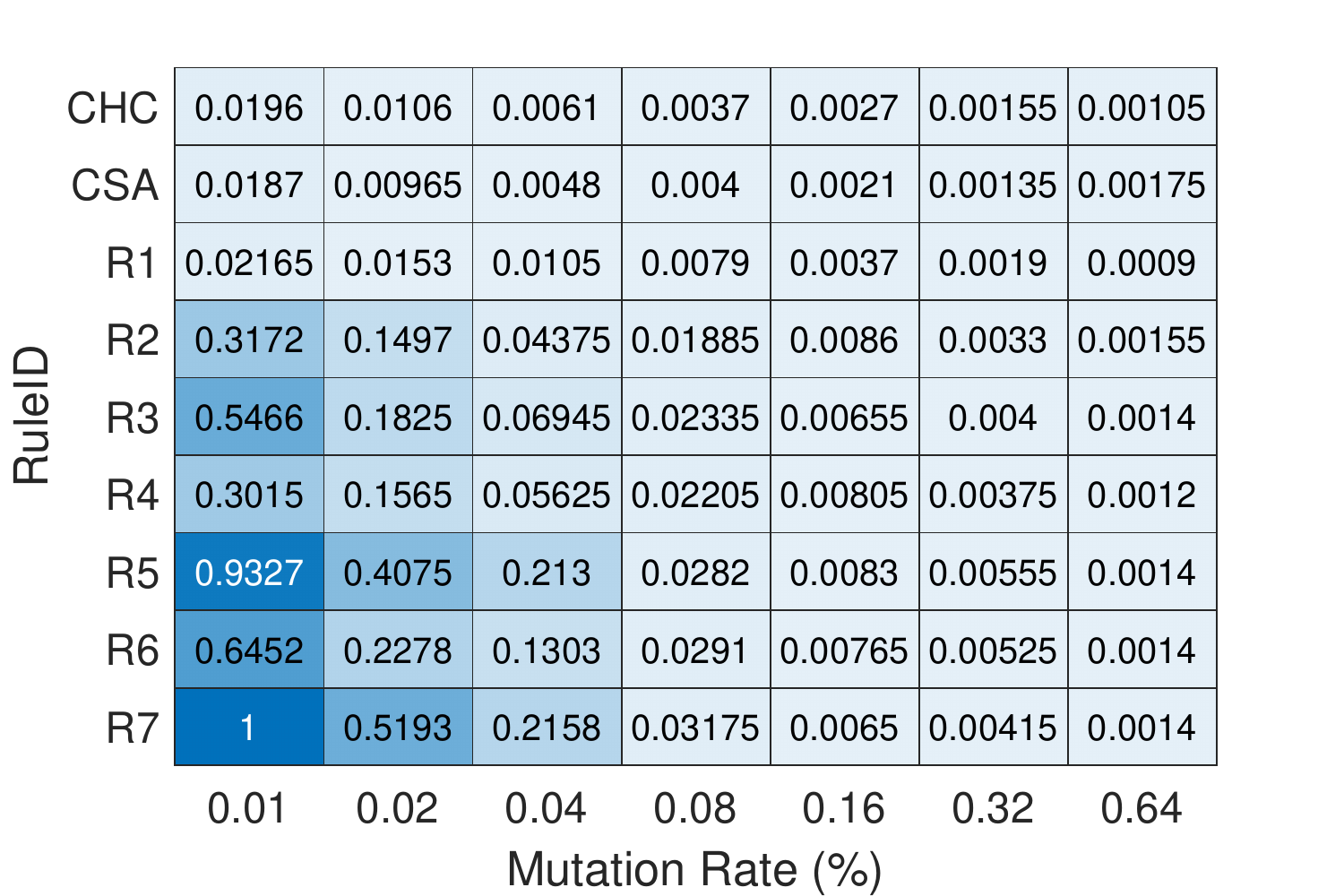}\label{fig:heatMap81nodesSolutionRcd03cp025}}
\subfloat[{\scriptsize \% Used slots ($cd$=0.3, $cp$=0.25)}]{\includegraphics[width=0.4\columnwidth]{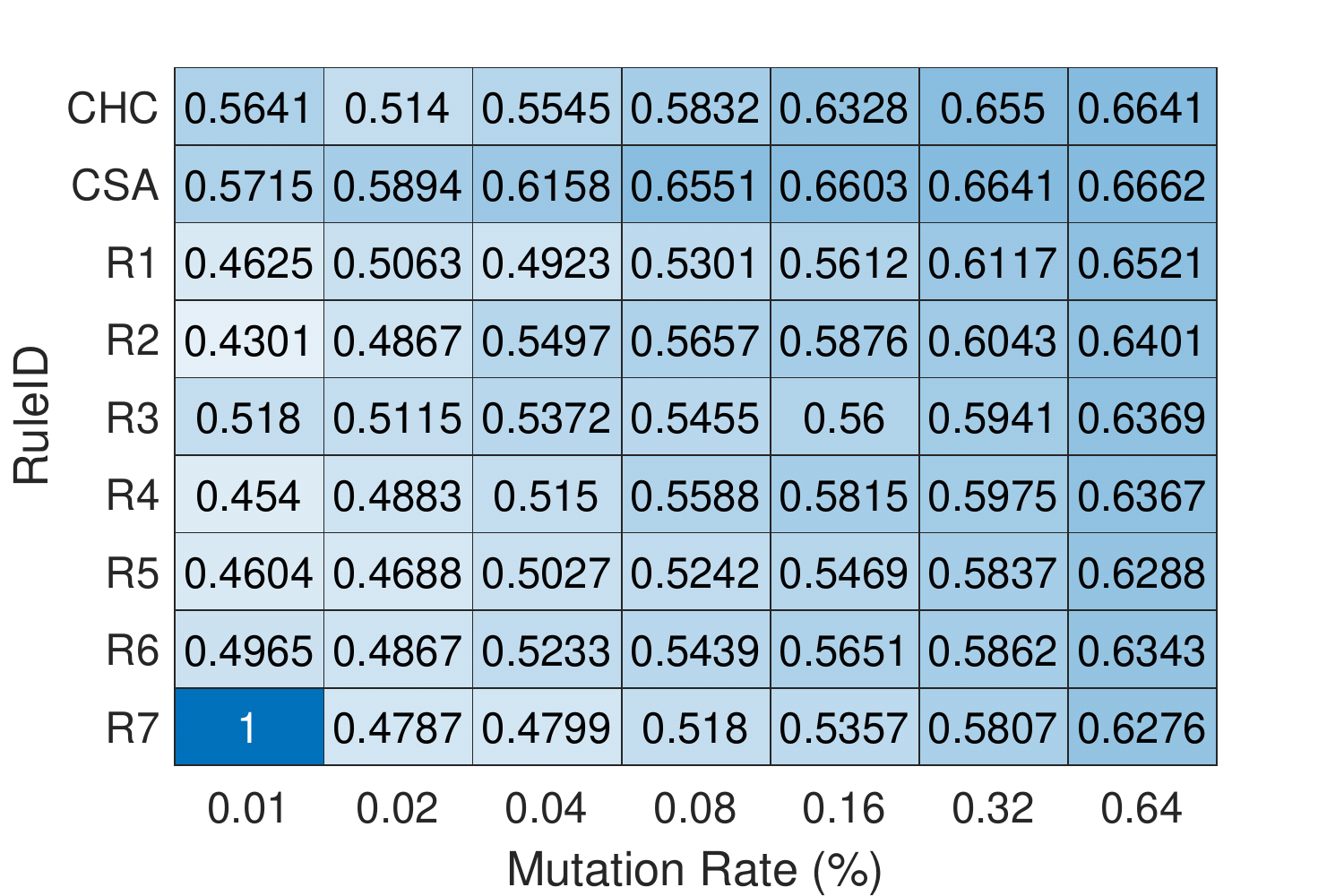}\label{fig:heatMap81nodesSlotsRcd03cp025}}

\subfloat[{\scriptsize \% Used resources ($cd$=0.3, $cp$=0.125)}]{\includegraphics[width=0.4\columnwidth]{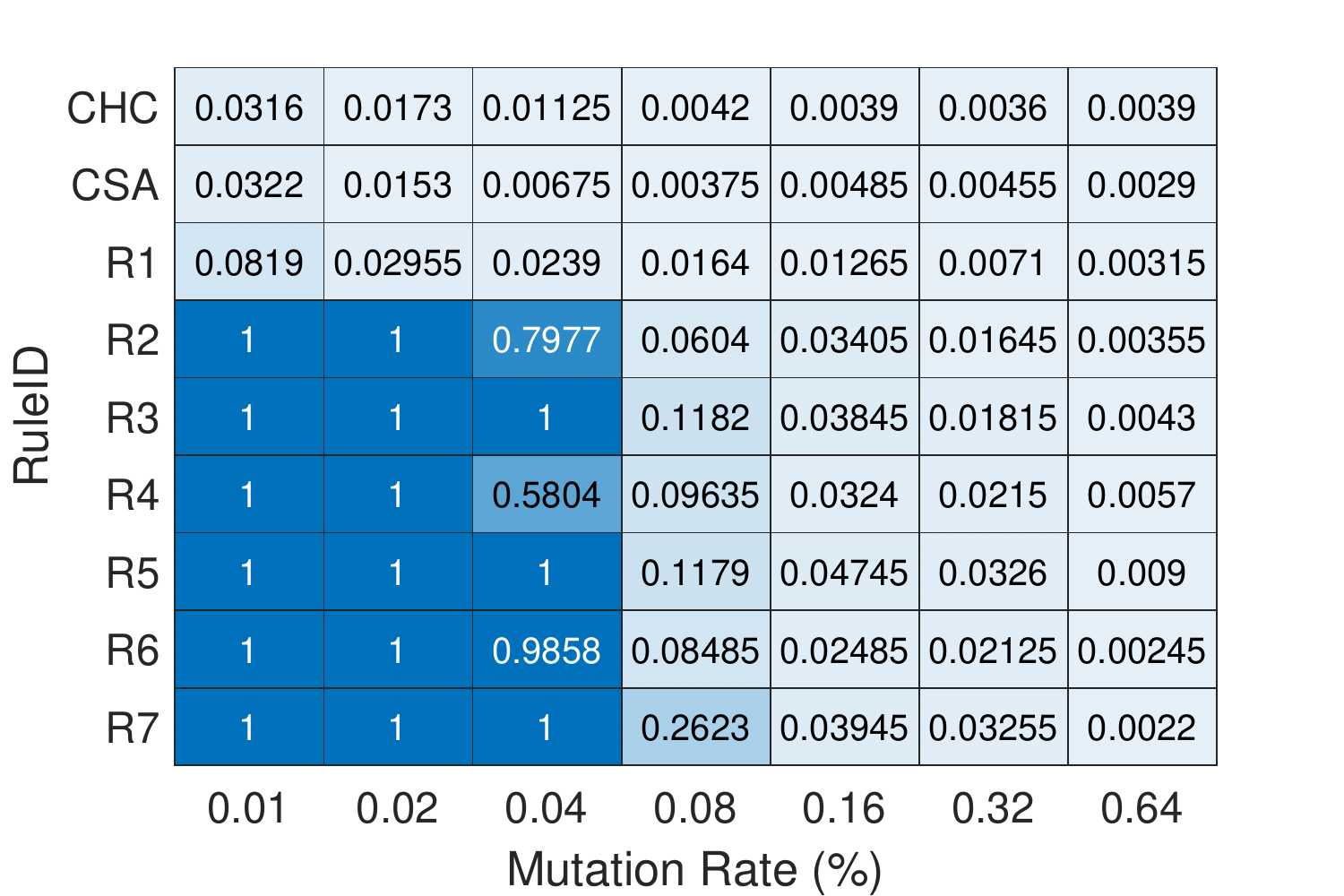}\label{fig:heatMap81nodesSolutionRcd03cp0125}}
\subfloat[{\scriptsize \% Used slots ($cd$=0.3, $cp$=0.125)}]{\includegraphics[width=0.4\columnwidth]{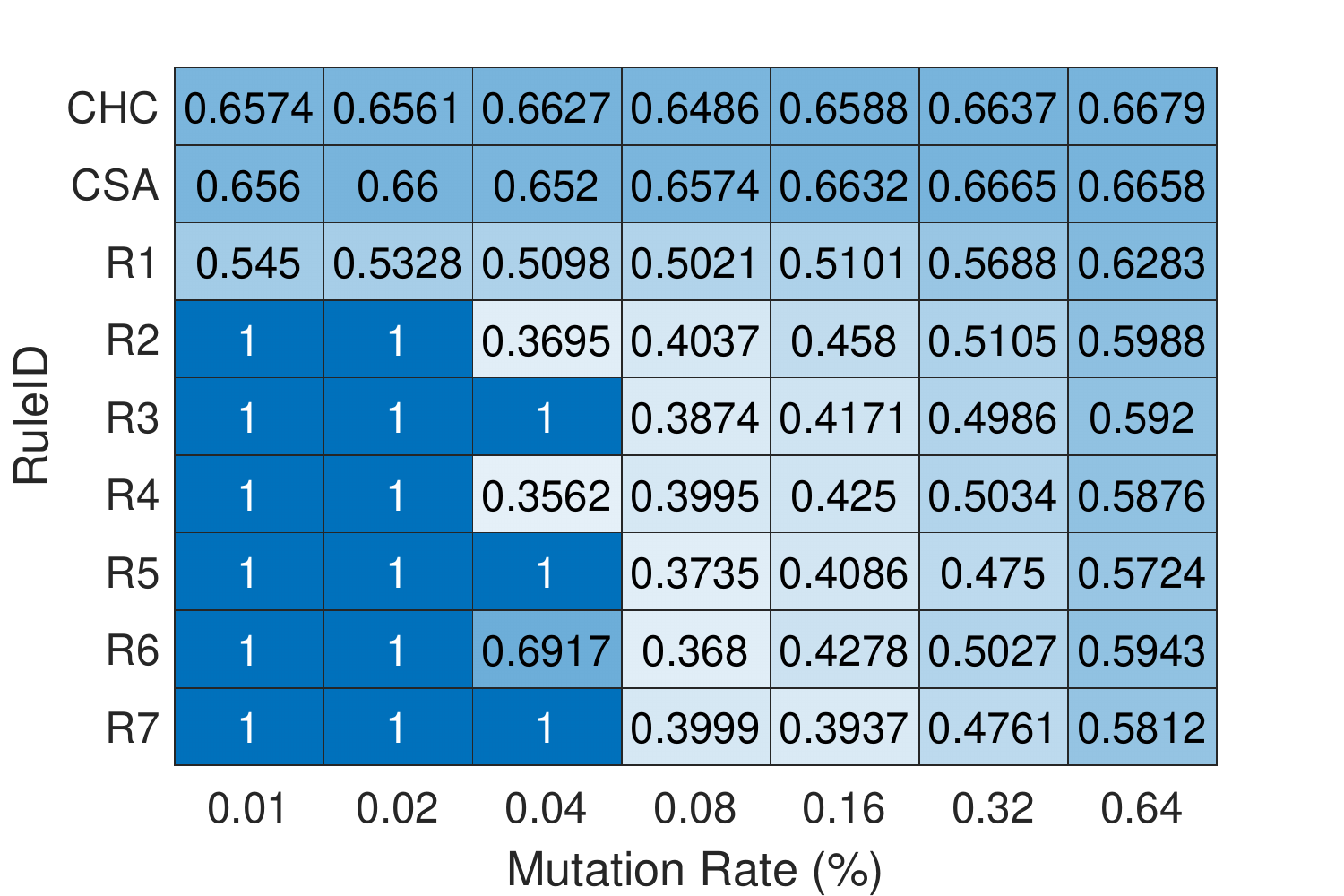}\label{fig:heatMap81nodesSlotsRcd03cp0125}}
\end{subfigures}
\caption{Variation of the evolved TDMA MAC protocol performance for Centralized Hill Climbing (CHC), Centralized Simulated Annealing (CSA) and \ALGNAME~(\ALGABRV) with 7 different rules on random networks with $81$ nodes and various levels of connection distance ($cd$) and connection probability ($cp$).} \label{fig:results-random-nets81}
\end{figure}

We report in the heatmaps shown in Figures \ref{fig:results-grid-nets}, \ref{fig:results-random-nets9}, \ref{fig:results-random-nets36} and \ref{fig:results-random-nets81} the variation of the evolved TDMA MAC protocol performance (as \% of used resources and used slots, average values across 28 runs of each algorithmic setting at the end of the optimization process, i.e., either as soon as a viable protocol configuration capable to obtain 100\% delivery rate is found, or after 10000 evaluations) w.r.t. the mutation rate (\%) and the rule ID used in \ALGABRV, as well as the two baseline single-objective centralized algorithms, namely Centralized Hill Climbing (CHC) and Centralized Simulated Annealing (CSA), for the cases of grid networks (9, 36 and 81 nodes), as well as random networks with 9, 36, and 81 nodes, respectively. For the random networks, we consider different combinations of connection distance ($cd$) and connection probability ($cp$) values. In the heatmaps, darker (lighter) color means worse (better) performance.

The main findings that can be inferred from the figures can be summarized as follows:
\begin{itemize}
 \item Concerning the used resources (first column of each figure), apart from the case of random networks with 81 nodes and $cd$=0.3, $cp$=1 (Figure \ref{fig:heatMap81nodesSolutionRcd03cp1}), in all the tested settings low mutation rates tend to lead to higher \% resource consumption. This is particularly evident in the grid topologies when \ALGABRV~is configured to use rules from R2 to R7. Another interesting case is random networks with 9 nodes and $cd$=0.8, $cp$=0.125 (Figure \ref{fig:heatMap9nodesSolutionRcd08cp0125}), where CHC and CSA perform quite poorly for the lowest values of mutation rate. Apart from the two aforementioned peculiar cases, when higher mutation rates are used the resource consumption tends to be lower, with most of the algorithms reaching roughly similar values.
 \item Concerning the used slots (second column of each figure), in most cases there is a trend similar to that observed for the used resources (the lower the mutation rate, the worse the performance), with some exceptions represented by e.g. the random networks with higher values of $cp$, especially with 36 and 81 nodes (Figures \ref{fig:heatMap36nodesSlotsRcd05cp1}, \ref{fig:heatMap36nodesSlotsRcd05cp05}, \ref{fig:heatMap81nodesSlotsRcd03cp1}, and \ref{fig:heatMap81nodesSlotsRcd03cp05}). Overall, apart from these four cases \ALGABRV~appears to use in general less slots than both centralized approaches. 
\end{itemize}
Apart from these two trends, this analysis reveals that in general the optimal mutation rate depends on the network size and the specific \ALGABRV~rule adopted for the online protocol evolution. This observation suggests, for instance, a possible extension of \ALGABRV~to use (self-)adaptive mutation rates.

\vspace{-3cm}

%\newpage

%Figure~\ref{fig:scatterPlotsGrid} shows the results of CHC, CSA and rule variants of \ALGABRV~on the networks with grid and random topologies consisting of 9, 36 and 81 nodes.

\begin{figure*}[!htpb]
%\begin{center}
\begin{subfigures}
\subfloat[$9$ nodes ($3\times 3$ grid)]{\includegraphics[width=0.3\columnwidth]{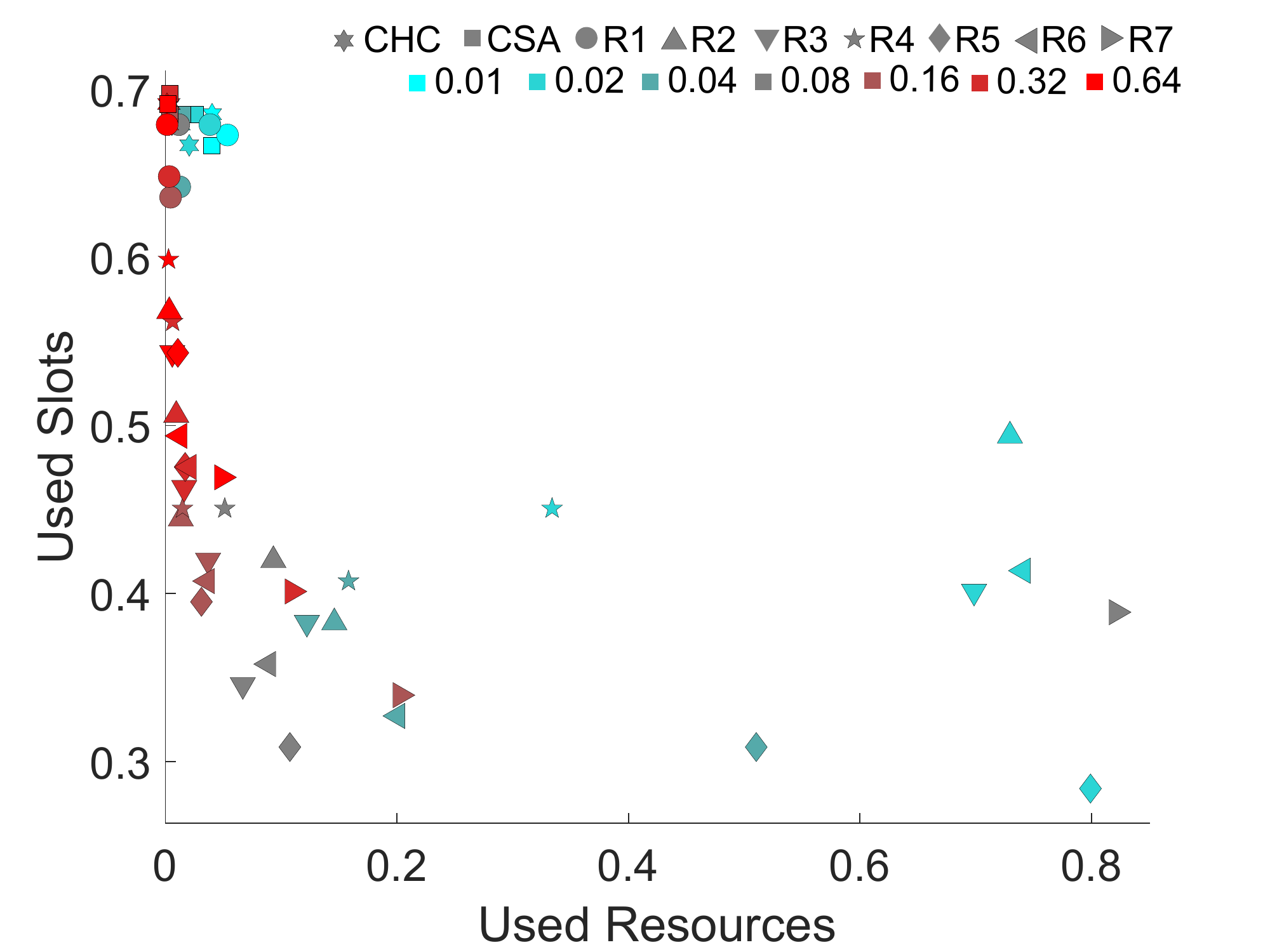}\label{fig:scatter9nodesMean}} 
\subfloat[$36$ nodes ($6\times 6$ grid)]{\includegraphics[width=0.3\columnwidth]{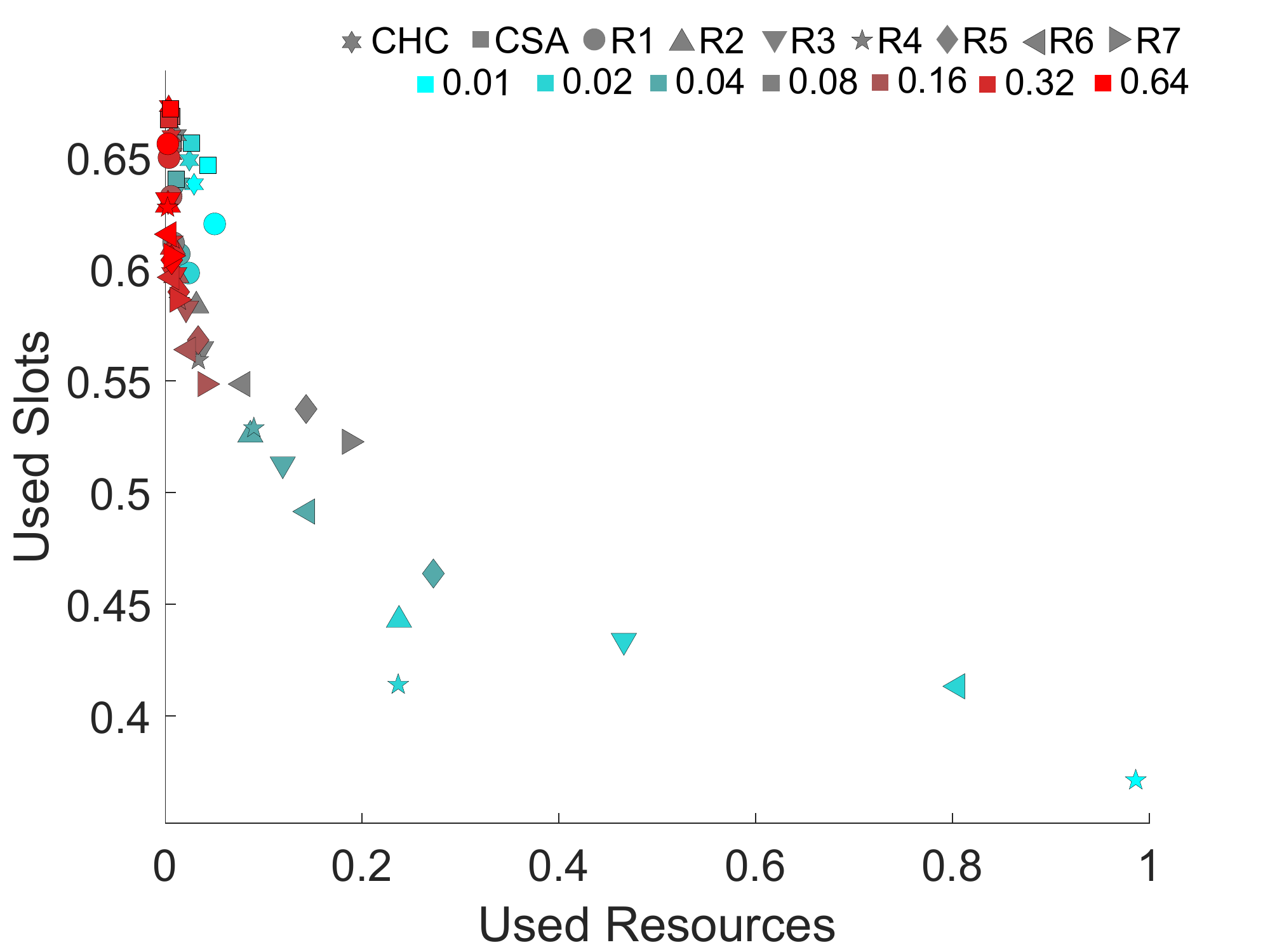}\label{fig:scatter36nodesMean}}
\subfloat[$81$ nodes ($9\times 9$ grid)]{\includegraphics[width=0.3\columnwidth]{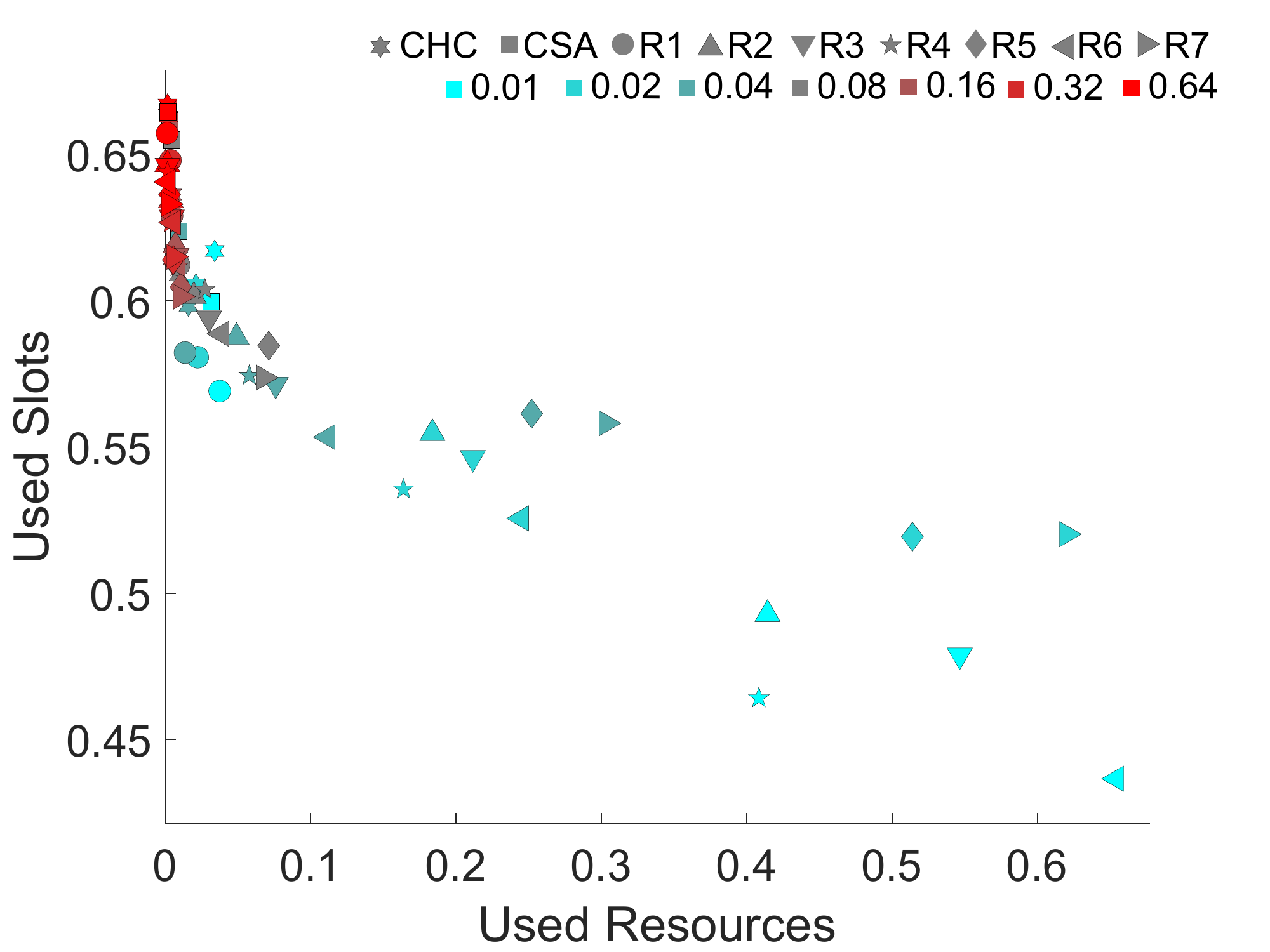}\label{fig:scatter81nodesMean}}

\subfloat[9 nodes ($cd = 0.8, cp = 0.5$)]{\includegraphics[width=0.3\columnwidth]{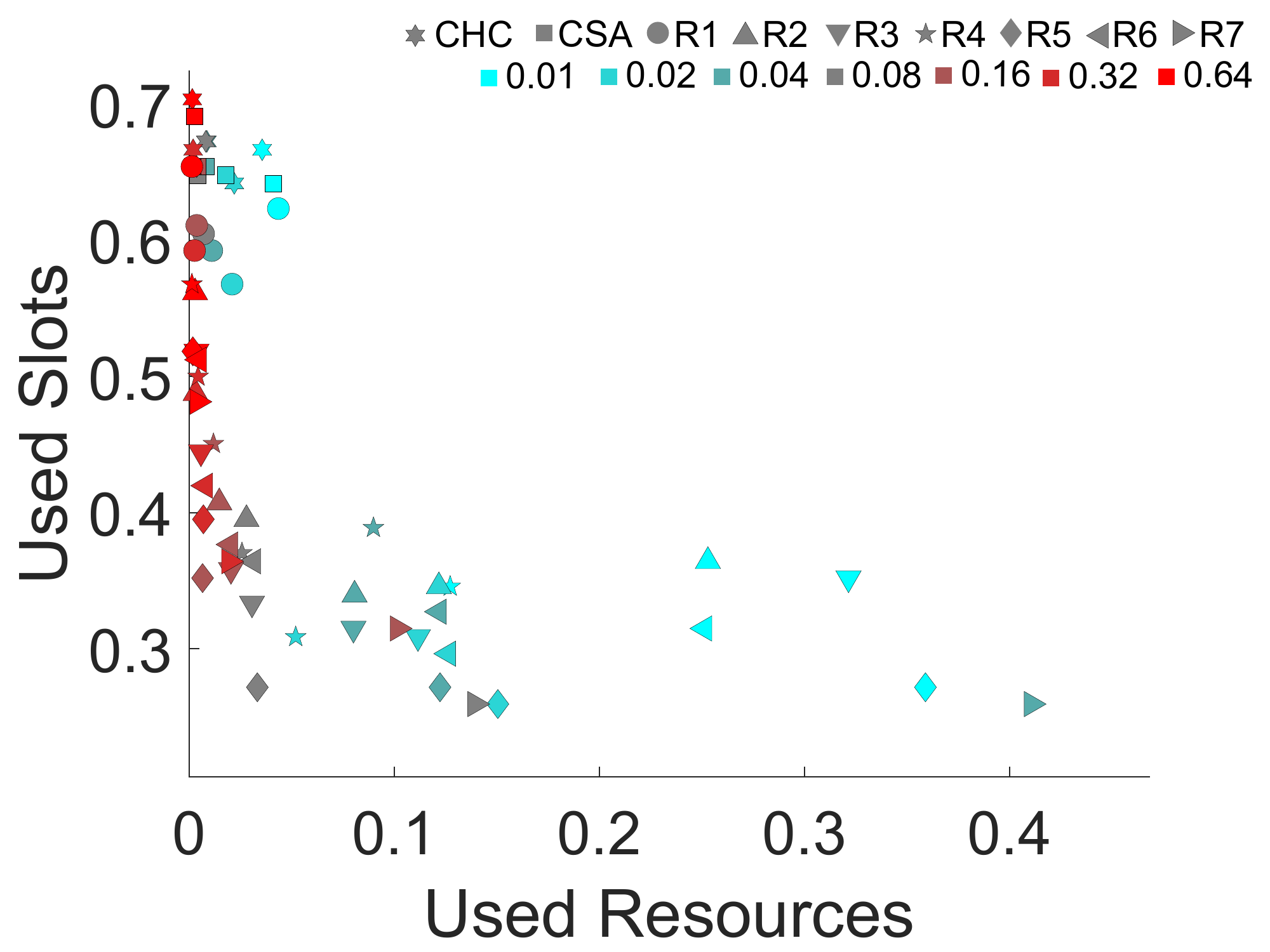}\label{fig:scatter9nodesRcd08cp05}}
\subfloat[36 nodes ($cd = 0.5, cp = 0.5$)]{\includegraphics[width=0.3\columnwidth]{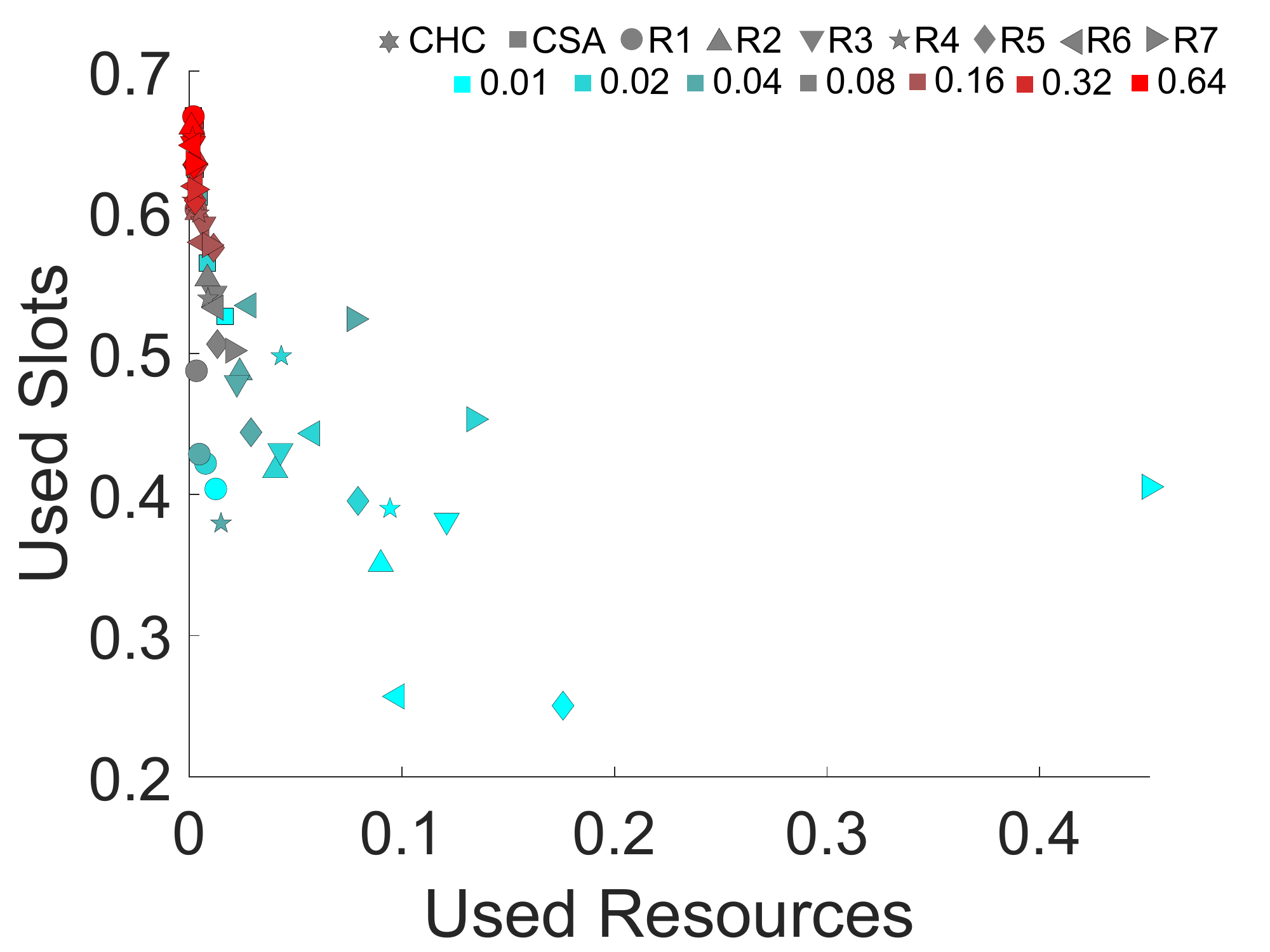}\label{fig:scatter36nodesRcd05cp05}}
\subfloat[81 nodes ($cd = 0.3, cp = 0.5$)]{\includegraphics[width=0.3\columnwidth]{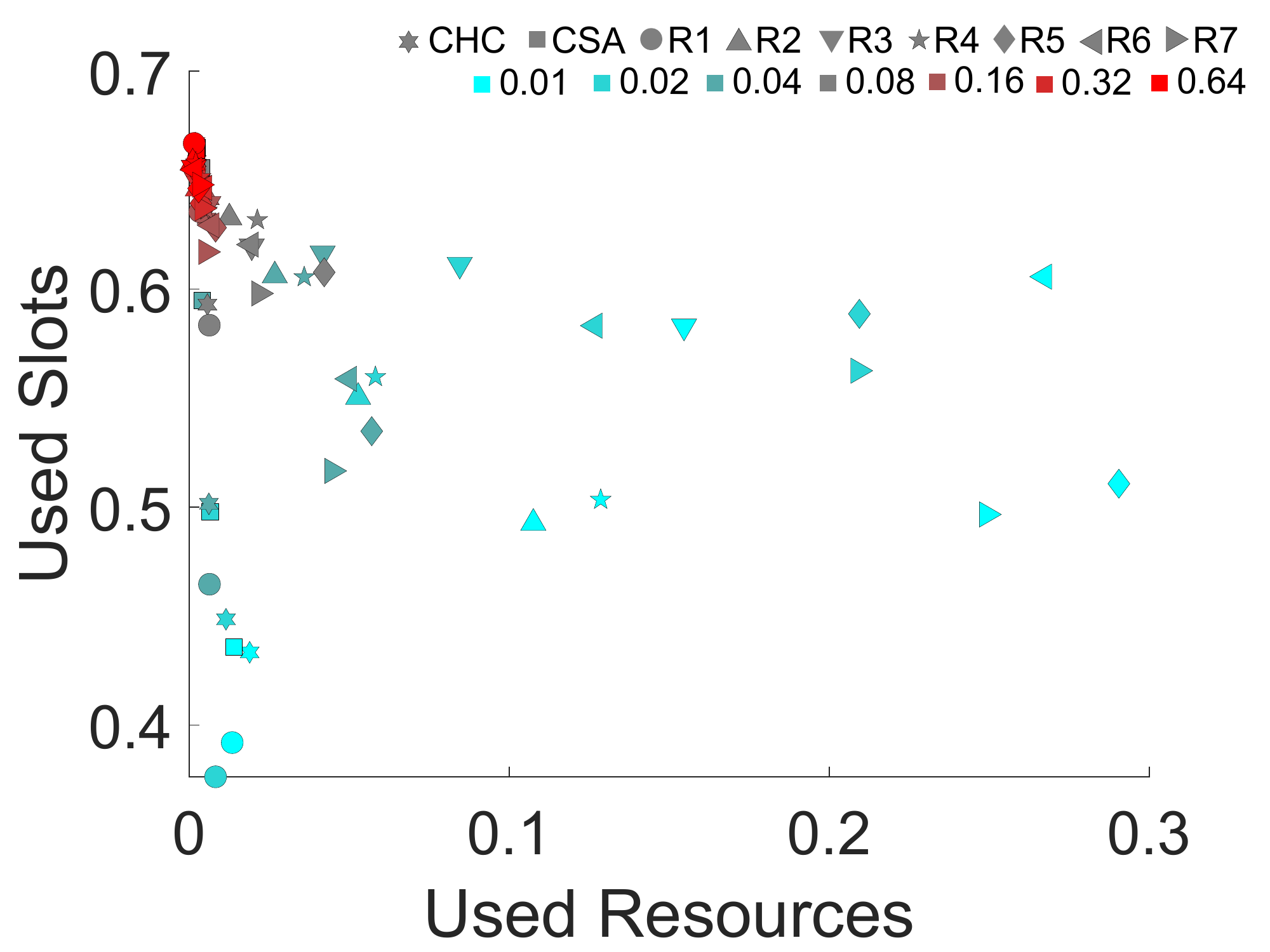}\label{fig:scatter81nodesRcd03cp05}}

\subfloat[9 nodes ($cd = 0.8, cp = 0.25$)]{\includegraphics[width=0.3\columnwidth]{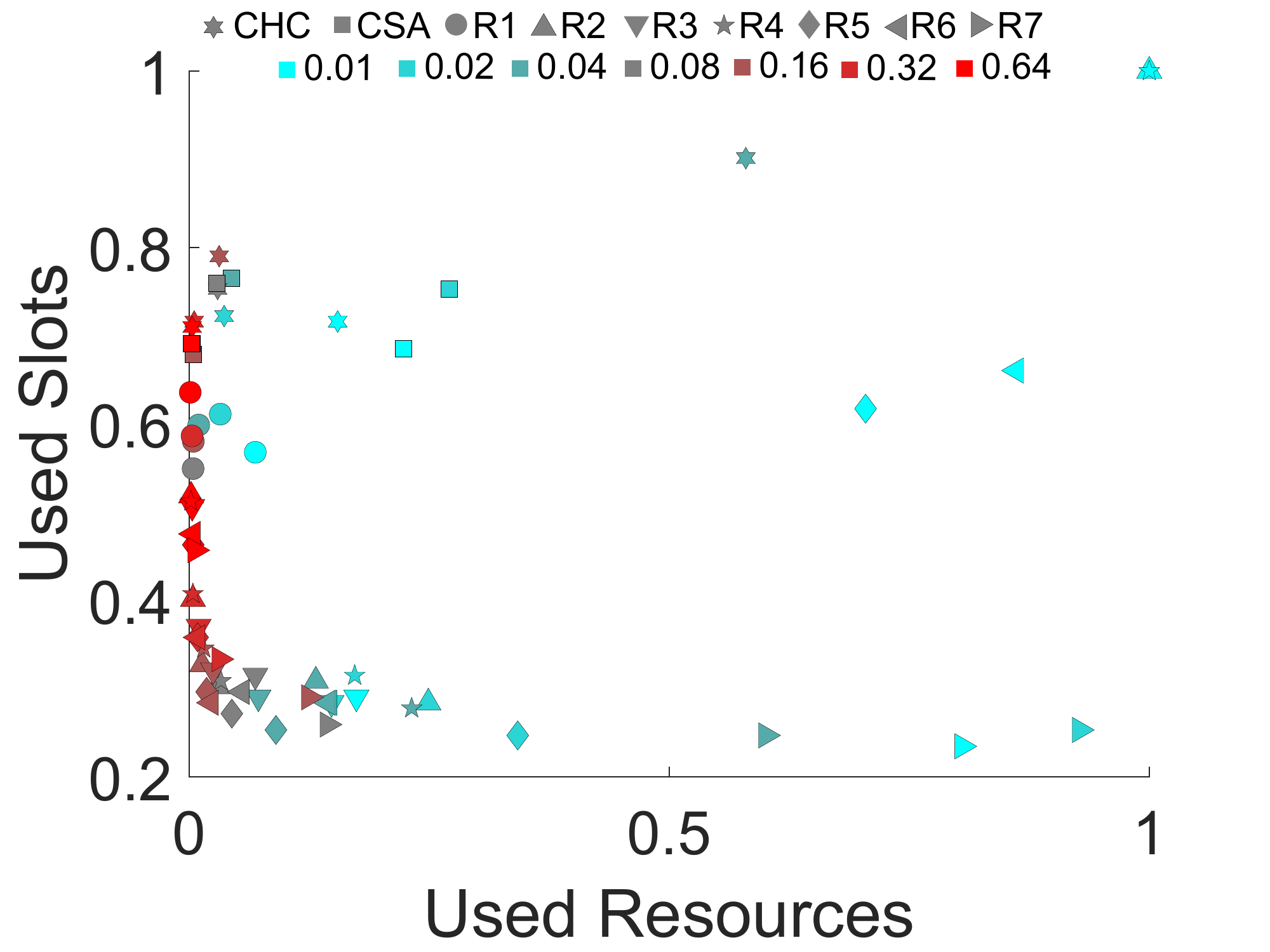}\label{fig:scatter9nodesRcd08cp025}}
\subfloat[36 nodes ($cd = 0.5, cp = 0.25$)]{\includegraphics[width=0.3\columnwidth]{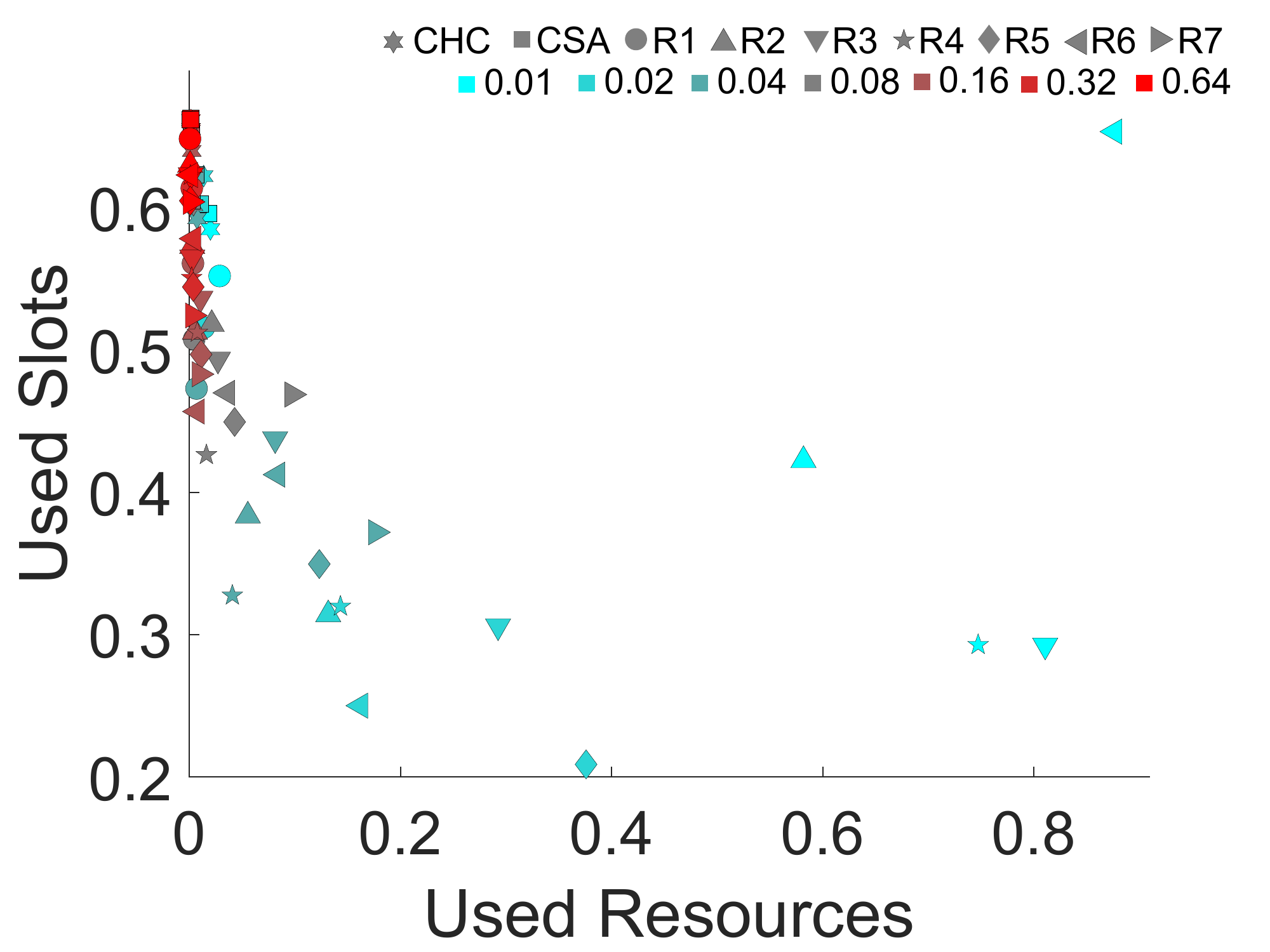}\label{fig:scatter36nodesRcd05cp025}}
\subfloat[81 nodes ($cd = 0.3, cp = 0.25$)]{\includegraphics[width=0.3\columnwidth]{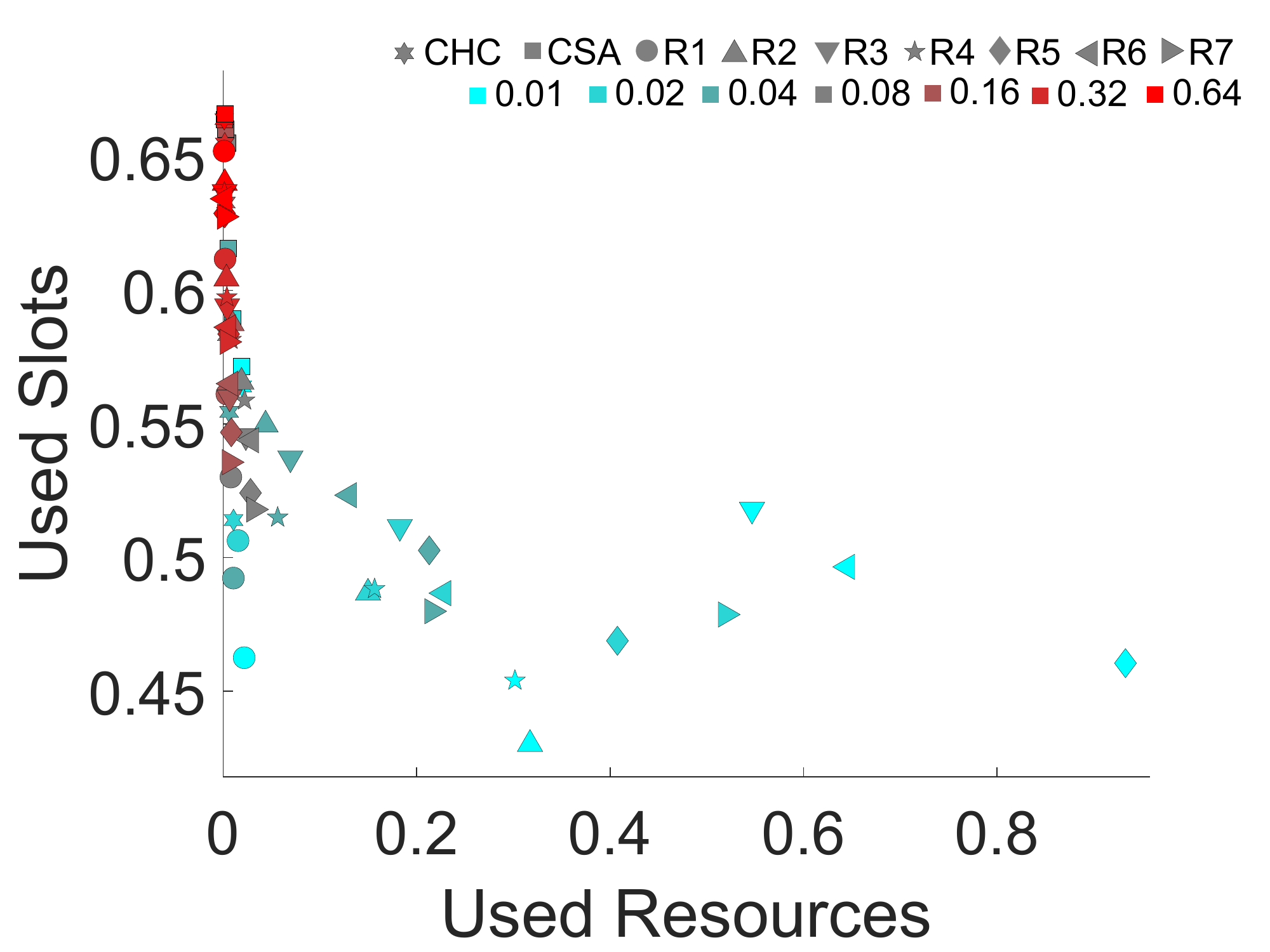}\label{fig:scatter81nodesRcd03cp025}}

\end{subfigures}
%\end{center}
\caption{Ratio of used slots vs ratio of used resources (median across $28$ runs) of CHC, CSA and the proposed \ALGABRV~algorithm (with seven different rules $\times$ seven different mutation rates) on grid topologies and six selected random network configurations (the remaining ones are reported in the main text).} \label{fig:scatterPlotsGrid}
\end{figure*}

\clearpage

\section{Statistical Analysis}

Table~\ref{tab:resultsComparisonSTD} shows the standard deviations of the results (based on slot use across 28 independent runs) produced by the variants of \ALGABRV~and the other algorithms under comparison.

\begin{table*}[!ht]
\caption{Comparison of the algorithms on grid and random networks with 9 to 81 nodes. The problem size is computed as $N$ (no. of nodes) $\times$ $S$ (no. of slots), which yields to $N^2$ since we set $N=S$. The value in each cell shows the ratio of used slots (std. dev. across 28 runs at the end of the optimization process) for the solutions found by the algorithms (``-'' indicates that no solution is found).}\label{tab:resultsComparisonSTD}
\begin{center}
\resizebox{\textwidth}{!}{
\begin{tabular}{lc?c|c?c|c|c?c|c?c|c|c|c|c|c|c|}
%\hline
\cline{3-16} & & \multicolumn{2}{c?}{\textbf{Group 1}} & \multicolumn{3}{c?}{\textbf{Group 2}} & \multicolumn{2}{c?}{\textbf{Group 3}} & \multicolumn{7}{c|}{\textbf{\ALGABRV}}\\ \hline
\multicolumn{1}{|l|}{\textbf{Problem}} & \textbf{Size} & \textbf{NSGA-II} & \textbf{MSEA} & \textbf{GA2O} & \textbf{CHC2O} & \textbf{CSA2O} & \textbf{CHC} & \textbf{CSA} & \textbf{R1} & \textbf{R2} & \textbf{R3} & \textbf{R4} & \textbf{R5} & \textbf{R6} & \textbf{R7} \\ \hline
\rowcolor[HTML]{EFEFEF} 
\multicolumn{1}{|l|}{\textbf{Grid9}} & \textbf{81} & 0.033 & 0.042 & 0.004 & 0.014 & 0.185 & 0.085 & 0.058 & 0.070 & 0.136 & 0.071 & 0.068 & 0.349 & 0.213 & 0.132 \\ \hline
\multicolumn{1}{|l|}{\textbf{9cp1}} & \textbf{81} & 0.012& 0.034& 0.015& 0.012& 0.222& 0.122& 0.114& 0.158& 0.129& 0.087& 0.160& 0.058& 0.048& 0.239 \\ \hline
\rowcolor[HTML]{EFEFEF} 
\multicolumn{1}{|l|}{\textbf{9cp05}} & \textbf{81} & 0.026& 0.067& 0.021& 0.018& 0.190& 0.118& 0.138& 0.117& 0.249& 0.276& 0.312& 0.347& 0.253& 0.348 \\ \hline
\multicolumn{1}{|l|}{\textbf{9cp025}} & \textbf{81} & -& -& 0.021& 0.016& 0.170& 0.151& 0.167& 0.107& 0.336& 0.321& 0.287& 0.353& 0.304& 0.404 \\ \hline
\rowcolor[HTML]{EFEFEF} 
\multicolumn{1}{|l|}{\textbf{9cp0125}} & \textbf{81} &-& -& 0.024& 0.014& 0.024& 0.166& 0.178& 0.085& 0.288& 0.339& 0.261& 0.302& 0.312& 0.395 \\ \hline\hline
\multicolumn{1}{|l|}{\textbf{Grid36}} & \textbf{1296} & 0.026& 0.022& 0.007& 0.004& 0.007& 0.018& 0.036& 0.034& 0.040& 0.258& 0.385& 0.046& 0.352& 0.020 \\ \hline
\rowcolor[HTML]{EFEFEF} 
\multicolumn{1}{|l|}{\textbf{36cp1}} & \textbf{1296} & 0.020& 0.016& 0.023& 0.027& 0.024& 0.075& 0.129& 0.187& 0.232& 0.284& 0.216& 0.270& 0.254& 0.292\\ \hline
\multicolumn{1}{|l|}{\textbf{36cp05}} & \textbf{1296} &0.014& 0.014& 0.015& 0.020& 0.015& 0.111& 0.078& 0.094& 0.157& 0.187& 0.168& 0.200& 0.165& 0.318 \\ \hline
\rowcolor[HTML]{EFEFEF} 
\multicolumn{1}{|l|}{\textbf{36cp025}}& \textbf{1296} & 0.010& 0.017& 0.041& 0.072& 0.137& 0.073& 0.091& 0.097& 0.181& 0.396& 0.371& 0.349& 0.337& 0.322 \\ \hline
\multicolumn{1}{|l|}{\textbf{36cp0125}}& \textbf{1296} & 0.009& 0.022& -& 0.052& 0.052& 0.079& 0.038& 0.075& 0.361& 0.306& 0.371& 0.426& 0.404& 0.367 \\ \hline\hline
\rowcolor[HTML]{EFEFEF} 
\multicolumn{1}{|l|}{\textbf{Grid81}} & \textbf{6561} & 0.005& 0.006& 0.003& 0.005& 0.004& 0.042& 0.037& 0.032& 0.049& 0.105& 0.056& 0.136& 0.187& 0.208 \\ \hline
\multicolumn{1}{|l|}{\textbf{81cp1}} & \textbf{6561} & 0.019& 0.009& 0.012& 0.016& 0.020& 0.145& 0.196& 0.218& 0.163& 0.173& 0.171& 0.176& -& 0.176 \\ \hline
\rowcolor[HTML]{EFEFEF} 
\multicolumn{1}{|l|}{\textbf{81cp05}} & \textbf{6561} & 0.005& 0.006& 0.011& 0.016& 0.014& 0.061& 0.096& 0.099& 0.154& 0.197& 0.172& 0.169& 0.140 & 0.178 \\ \hline
\multicolumn{1}{|l|}{\textbf{81cp025}} & \textbf{6561} & 0.003& 0.006& 0.035& 0.034& 0.035& 0.070& 0.096& 0.076& 0.296& 0.230& 0.312& 0.358& 0.236& 0.266\\ \hline
\rowcolor[HTML]{EFEFEF} 
\multicolumn{1}{|l|}{\textbf{81cp0125}} & \textbf{6561} & 0.002& 0.006& -& 0.019& 0.039& 0.063& 0.033& 0.082& 0.342& 0.233& 0.359& 0.312& 0.242& 0.256 \\ \hline
\end{tabular}
}
\end{center}
\end{table*}

\begin{figure}[!ht]
 \centering
 \includegraphics[width=0.6\columnwidth]{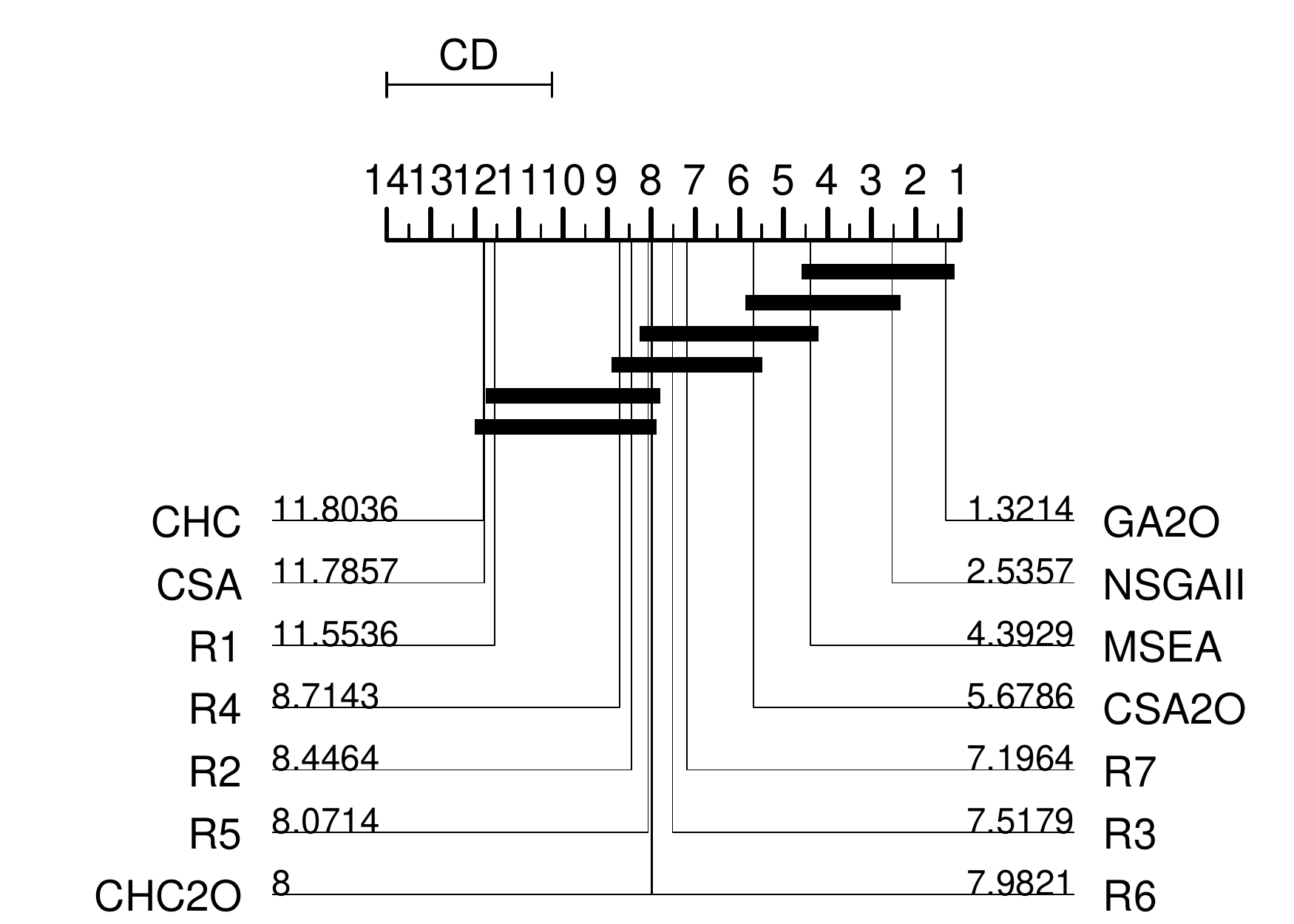}
 \caption{Critical differences of \ALGABRV~vs the compared algorithms on grid networks with 9 nodes.}
 \label{fig:cdGrid9}
\end{figure}

\begin{figure}[!ht]
 \centering
 \includegraphics[width=0.6\columnwidth]{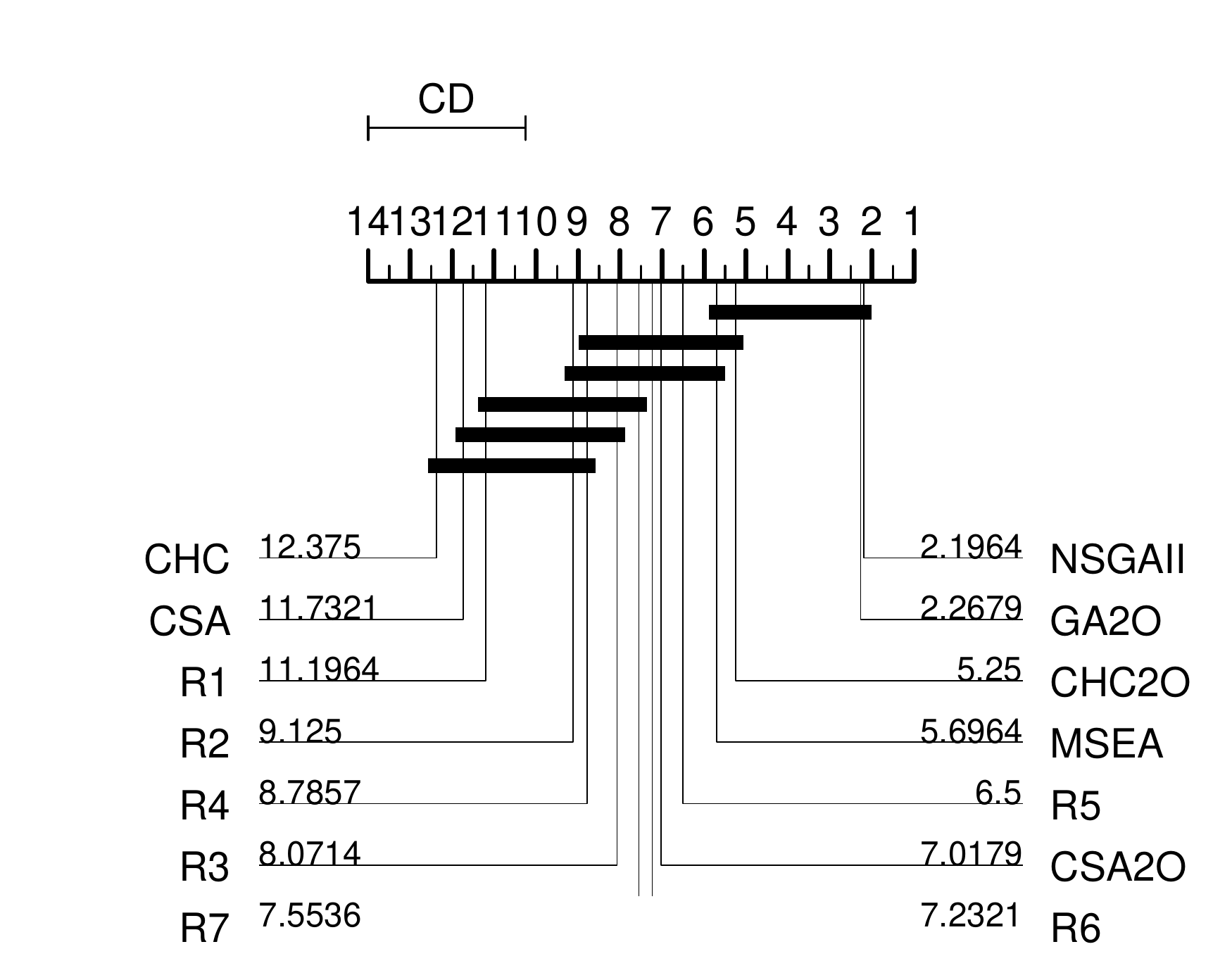}
 \caption{Critical differences of \ALGABRV~vs the compared algorithms on random networks with 9 nodes, $cd=0.8$ and $cp=1$.}
 \label{fig:cd9cp1}
\end{figure}

\begin{figure}[!ht]
 \centering
 \includegraphics[width=0.6\columnwidth]{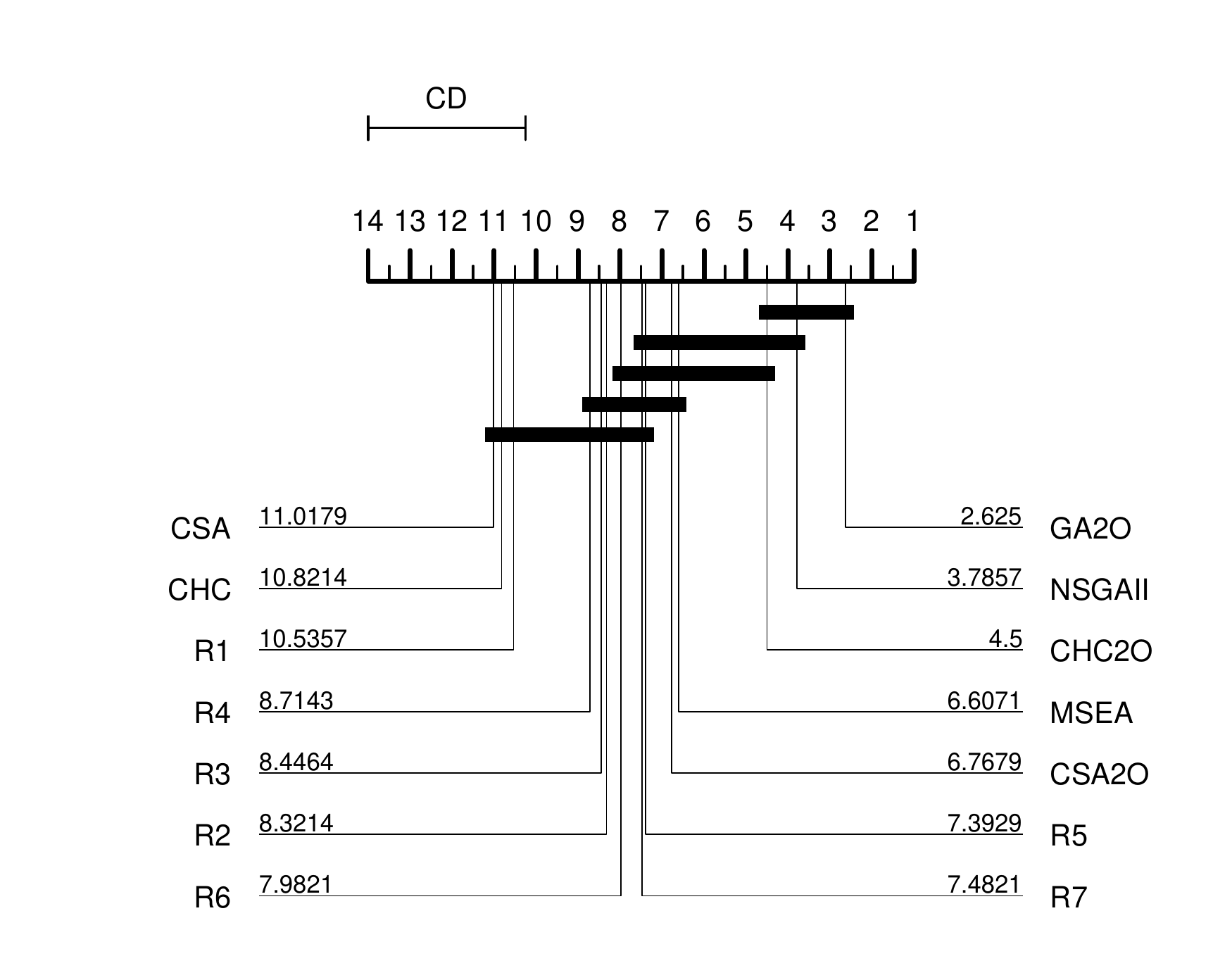}
 \caption{Critical differences of \ALGABRV~vs the compared algorithms on random networks with 9 nodes, $cd=0.8$ and $cp=0.5$.}
 \label{fig:cd9cp05}
\end{figure}

\begin{figure}[!ht]
 \centering
 \includegraphics[width=0.6\columnwidth]{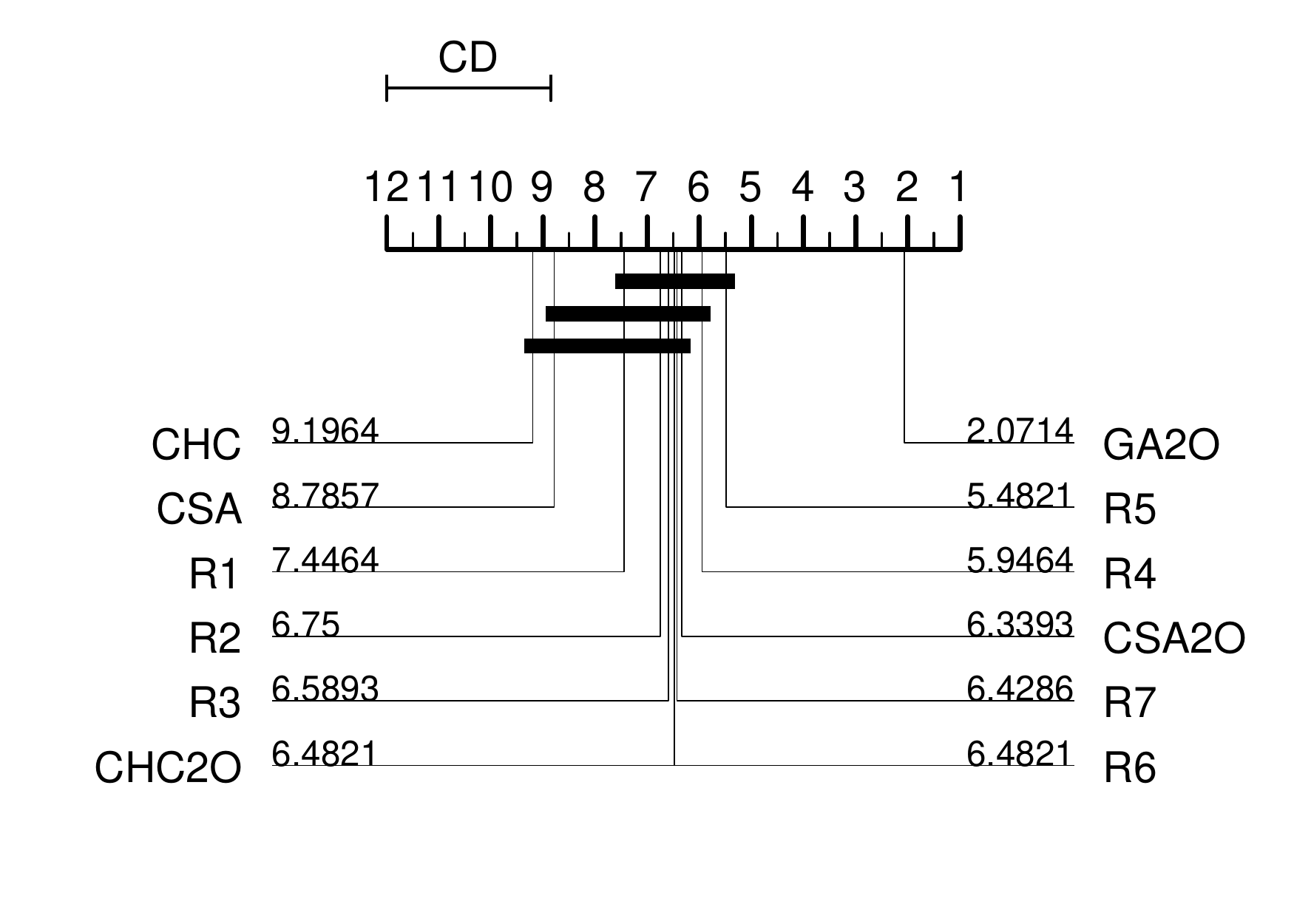}
 \caption{Critical differences of \ALGABRV~vs the compared algorithms on random networks with 9 nodes, $cd=0.8$ and $cp=0.25$.}
 \label{fig:cd9cp025}
\end{figure}

\begin{figure}[!ht]
 \centering
 \includegraphics[width=0.6\columnwidth]{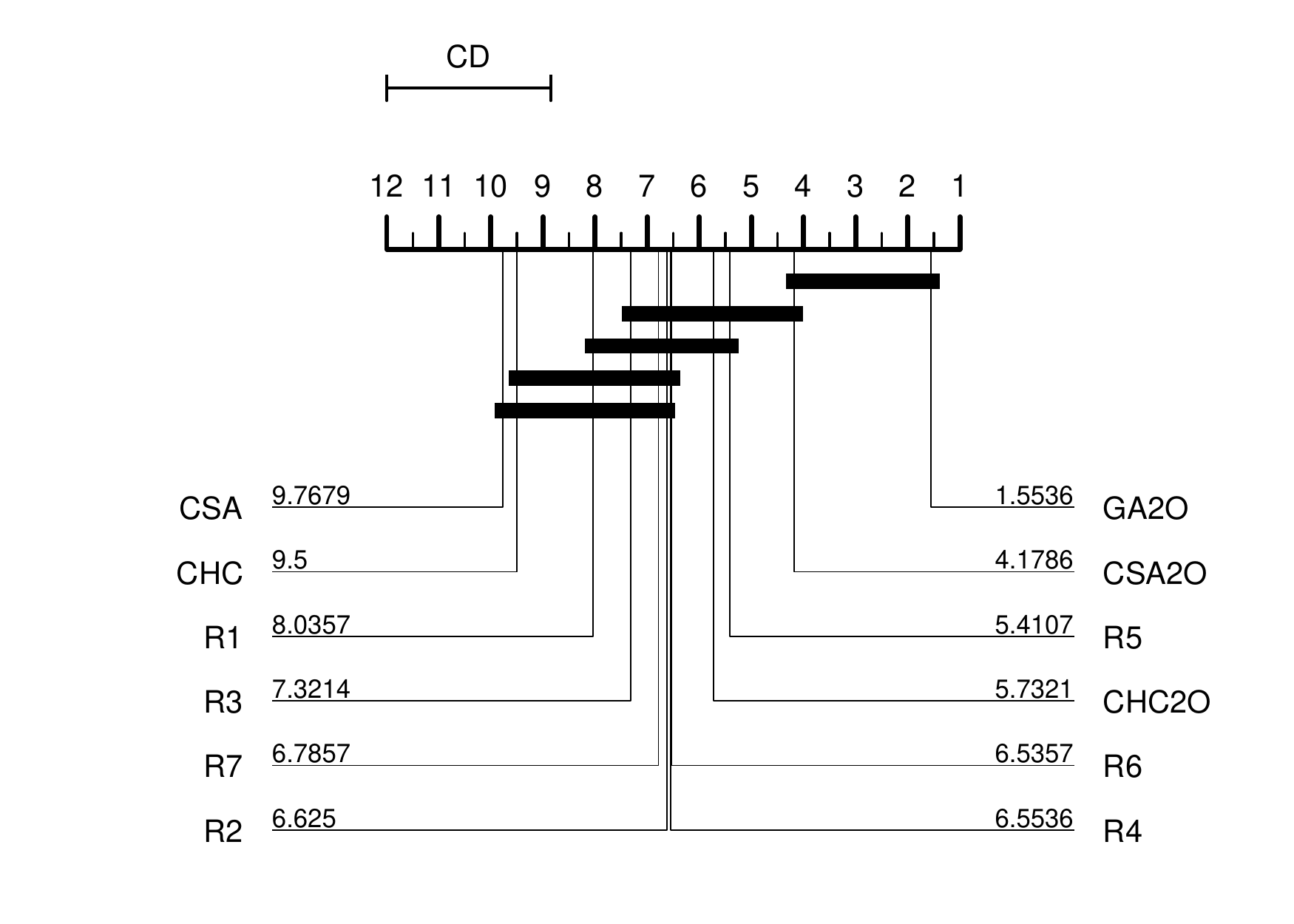}
 \caption{Critical differences of \ALGABRV~vs the compared algorithms on random networks with 9 nodes, $cd=0.8$ and $cp=0.125$.}
 \label{fig:cd9cp0125}
\end{figure}

\begin{figure}[!ht]
 \centering
 \includegraphics[width=0.6\columnwidth]{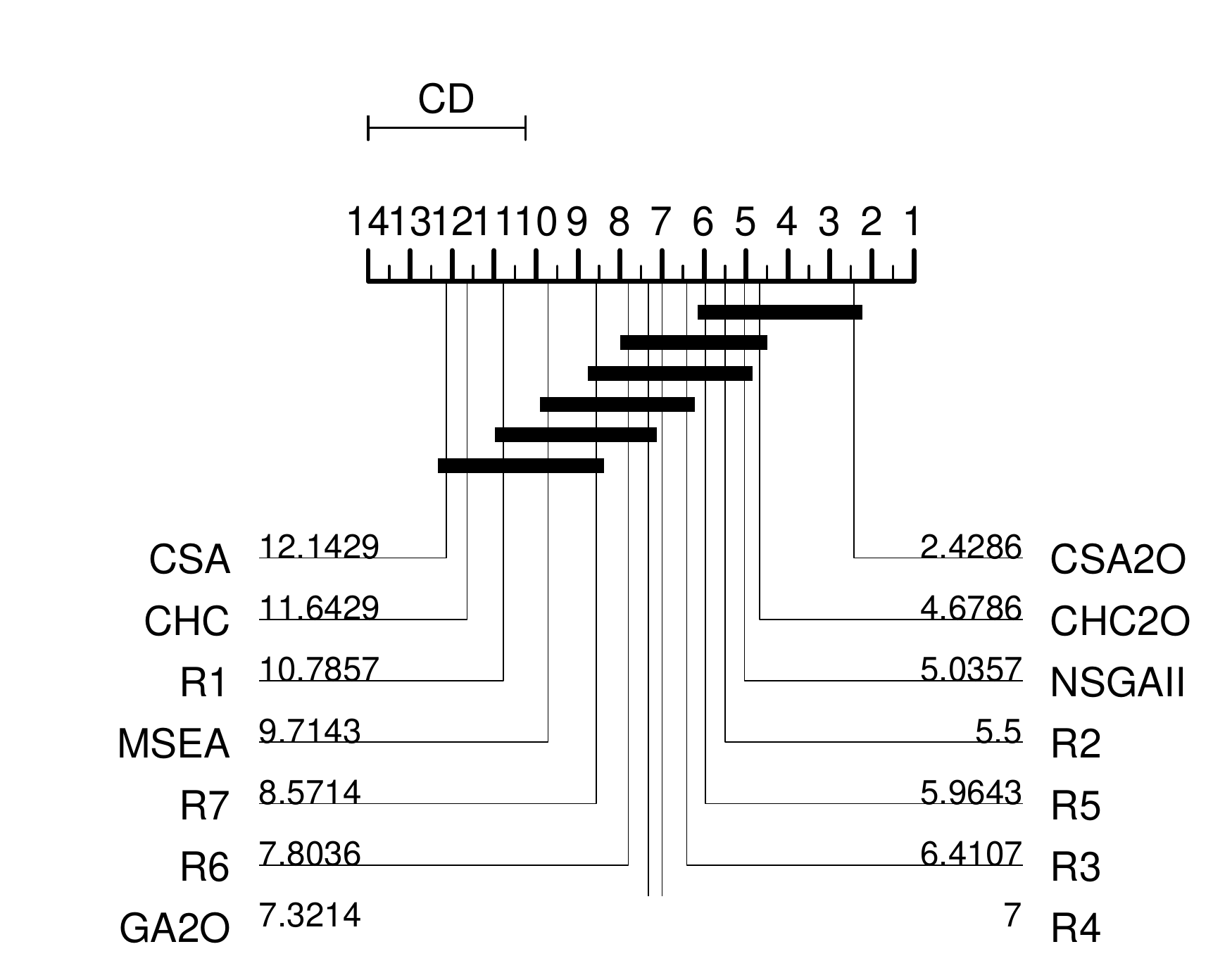}
 \caption{Critical differences of \ALGABRV~vs the compared algorithms on grid networks with 36 nodes.}
 \label{fig:cdGrid36}
\end{figure}

\begin{figure}[!ht]
 \centering
 \includegraphics[width=0.6\columnwidth]{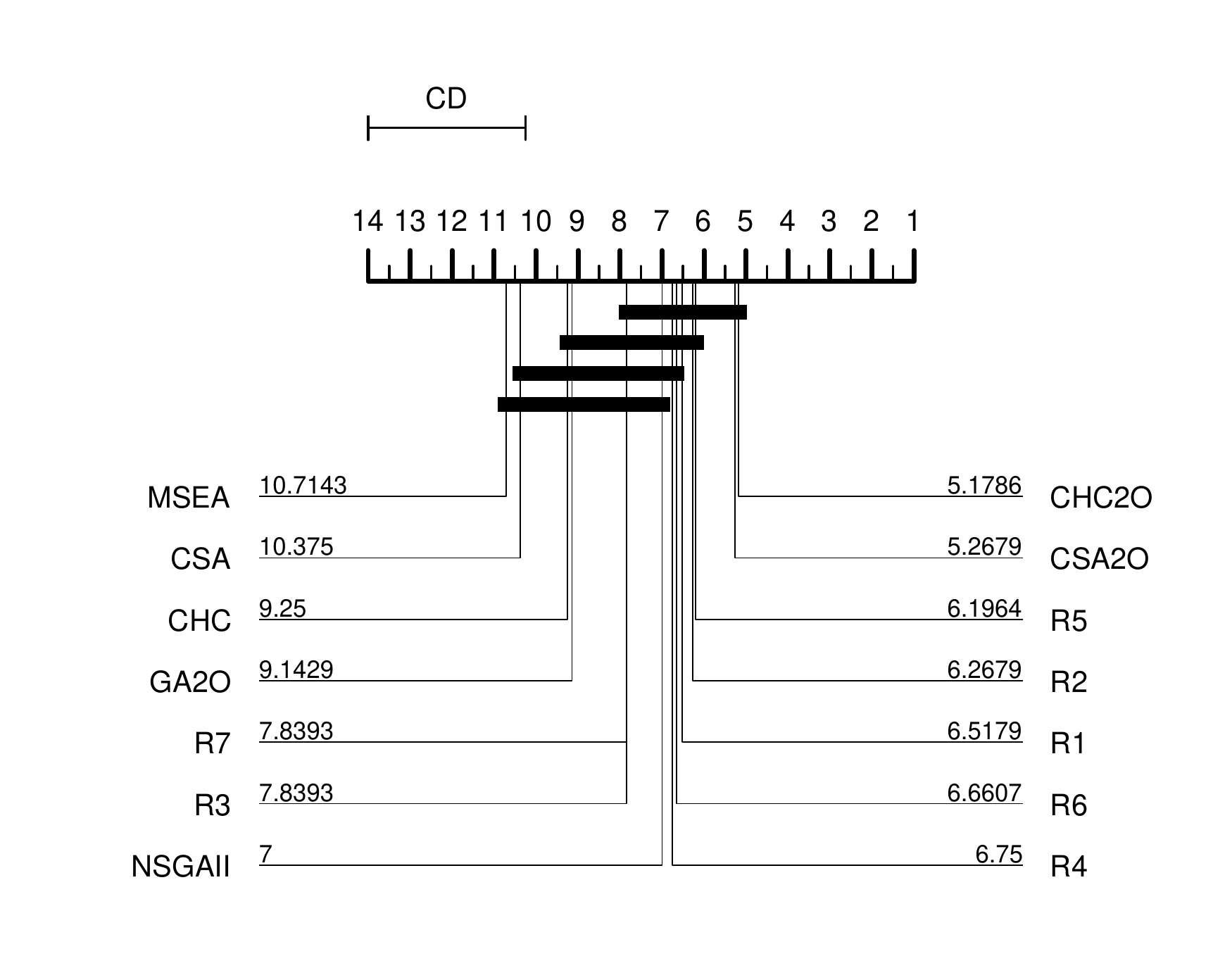}
 \caption{Critical differences of \ALGABRV~vs the compared algorithms on random networks with 36 nodes, $cd=0.5$ and $cp=1$.}
 \label{fig:cd36cp1}
\end{figure}

\begin{figure}[!ht]
 \centering
 \includegraphics[width=0.6\columnwidth]{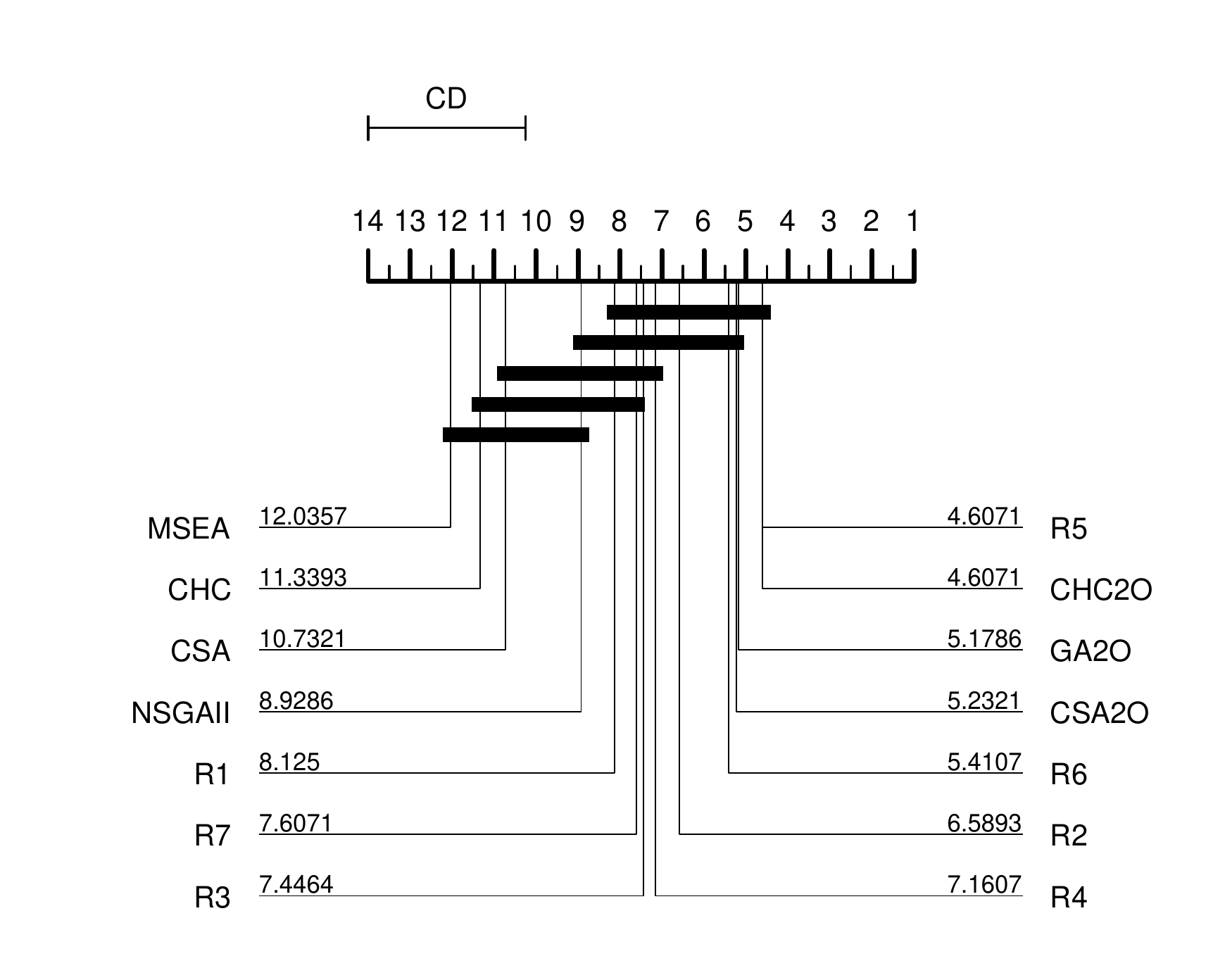}
 \caption{Critical differences of \ALGABRV~vs the compared algorithms on random networks with 36 nodes, $cd=0.5$ and $cp=0.5$.}
 \label{fig:cd36cp05}
\end{figure}

\begin{figure}[!ht]
 \centering
 \includegraphics[width=0.6\columnwidth]{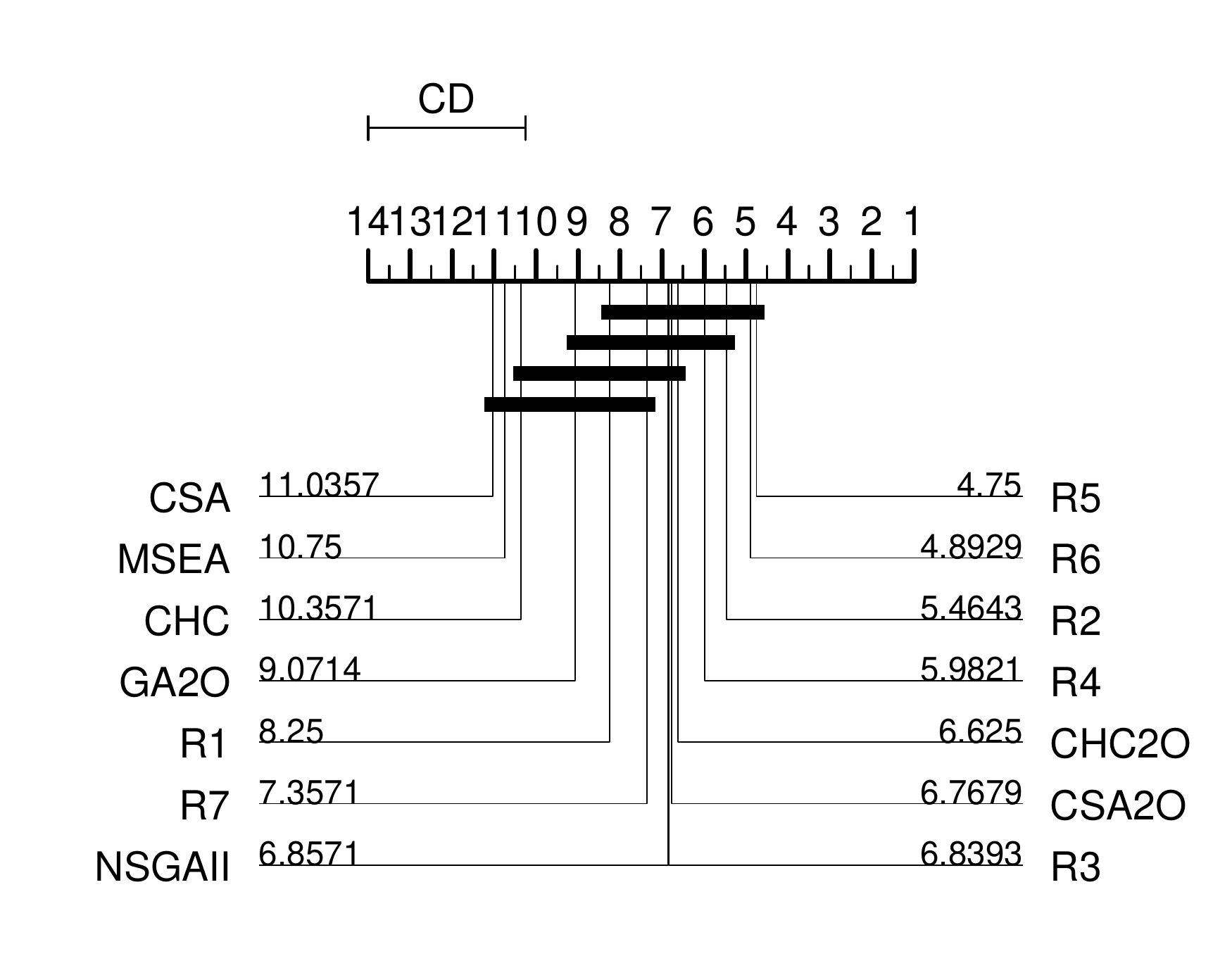}
 \caption{Critical differences of \ALGABRV~vs the compared algorithms on random networks with 36 nodes, $cd=0.5$ and $cp=0.25$.}
 \label{fig:cd36cp025}
\end{figure}

\begin{figure}[!ht]
 \centering
 \includegraphics[width=0.6\columnwidth]{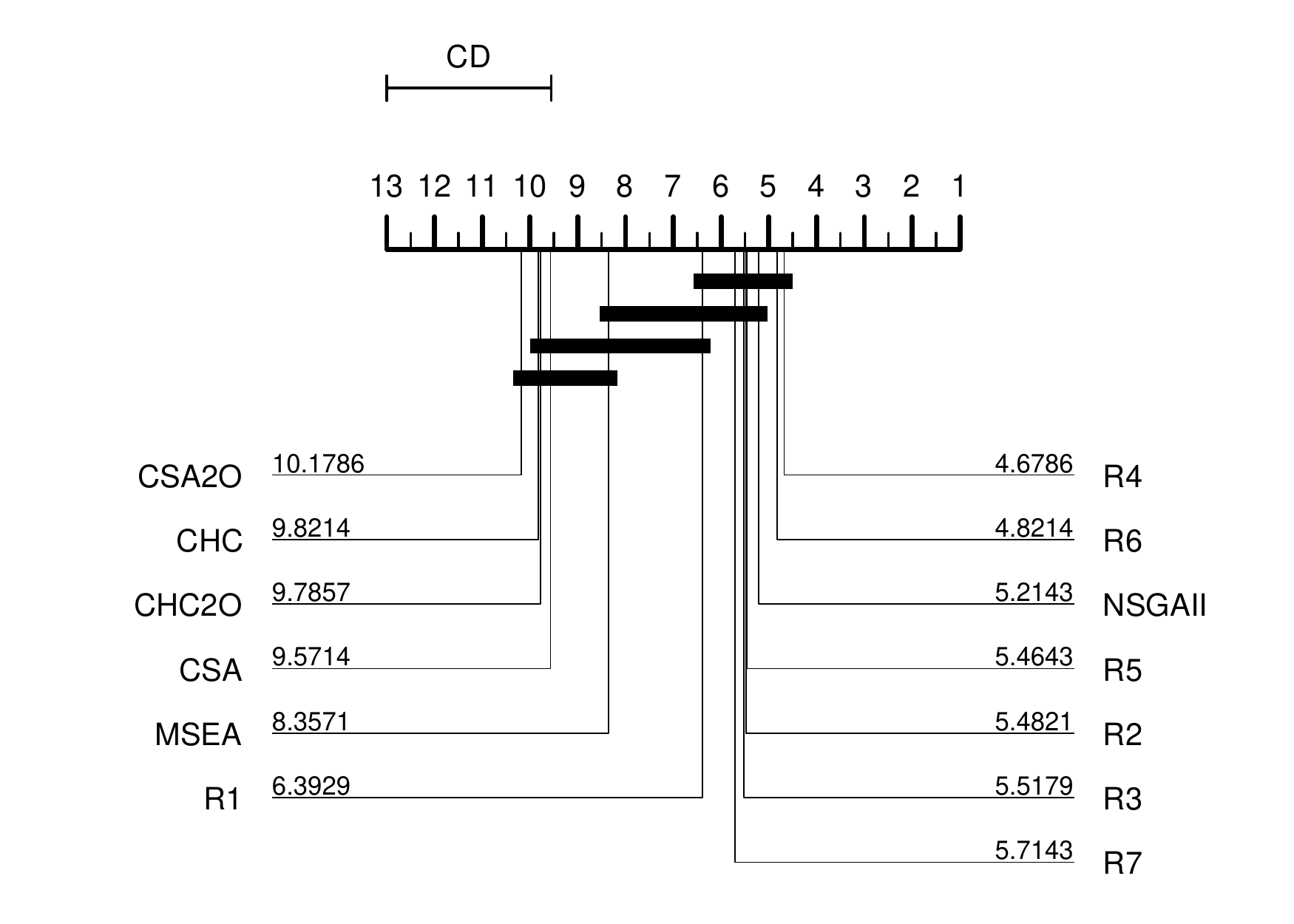}
 \caption{Critical differences of \ALGABRV~vs the compared algorithms on random networks with 36 nodes, $cd=0.5$ and $cp=0.125$.}
 \label{fig:cd36cp0125}
\end{figure}

\begin{figure}[!ht]
 \centering
 \includegraphics[width=0.6\columnwidth]{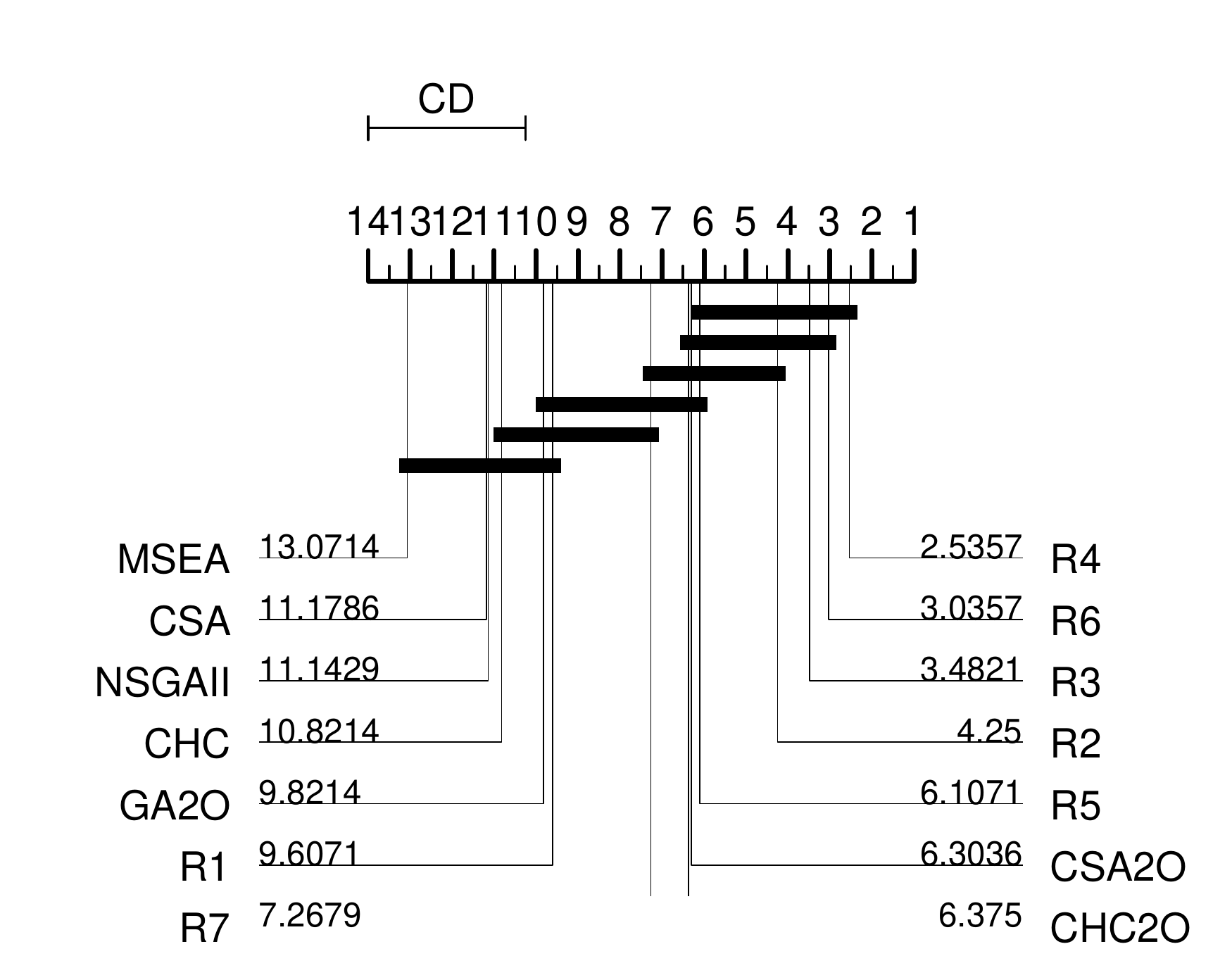}
 \caption{Critical differences of \ALGABRV~vs the compared algorithms on grid networks with 81 nodes.}
 \label{fig:cdGrid81}
\end{figure}

\begin{figure}[!ht]
 \centering
 \includegraphics[width=0.6\columnwidth]{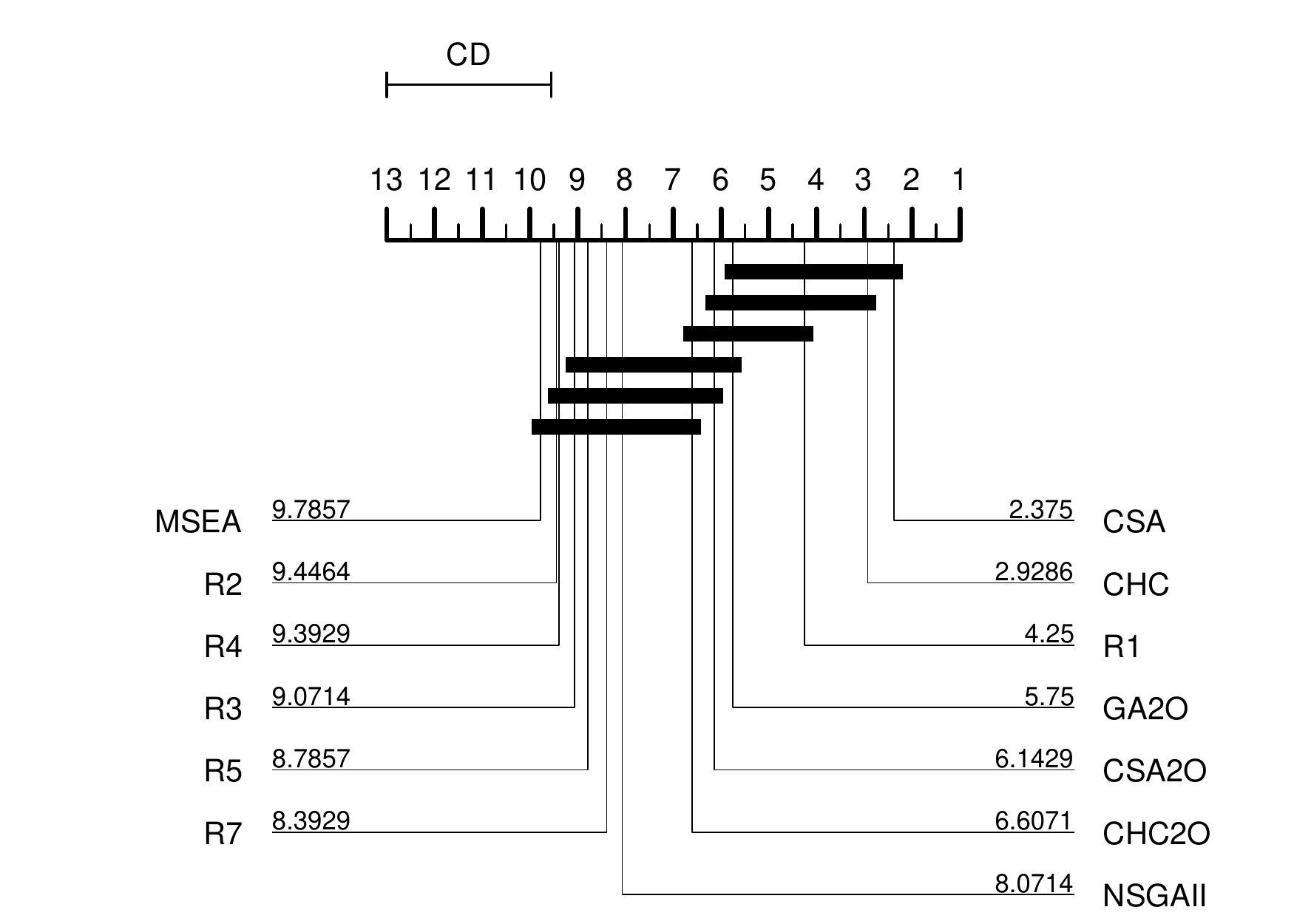}
 \caption{Critical differences of \ALGABRV~vs the compared algorithms on random networks with 81 nodes, $cd=0.3$ and $cp=1$.}
 \label{fig:cd81cp1}
\end{figure}

\begin{figure}[!ht]
 \centering
 \includegraphics[width=0.6\columnwidth]{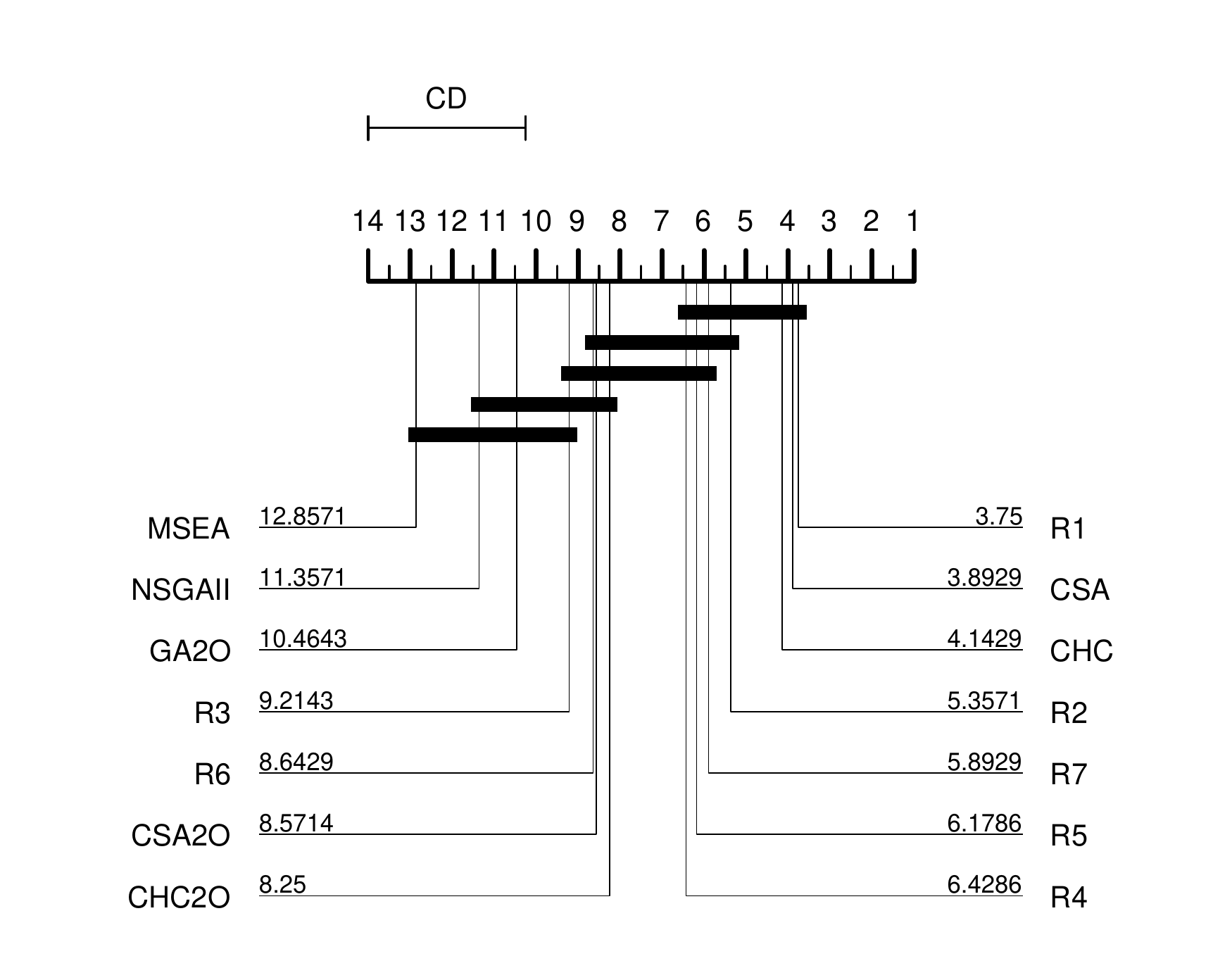}
 \caption{Critical differences of \ALGABRV~vs the compared algorithms on random networks with 81 nodes, $cd=0.3$ and $cp=0.5$.}
 \label{fig:cd81cp05}
\end{figure}

\begin{figure}[!ht]
 \centering
 \includegraphics[width=0.6\columnwidth]{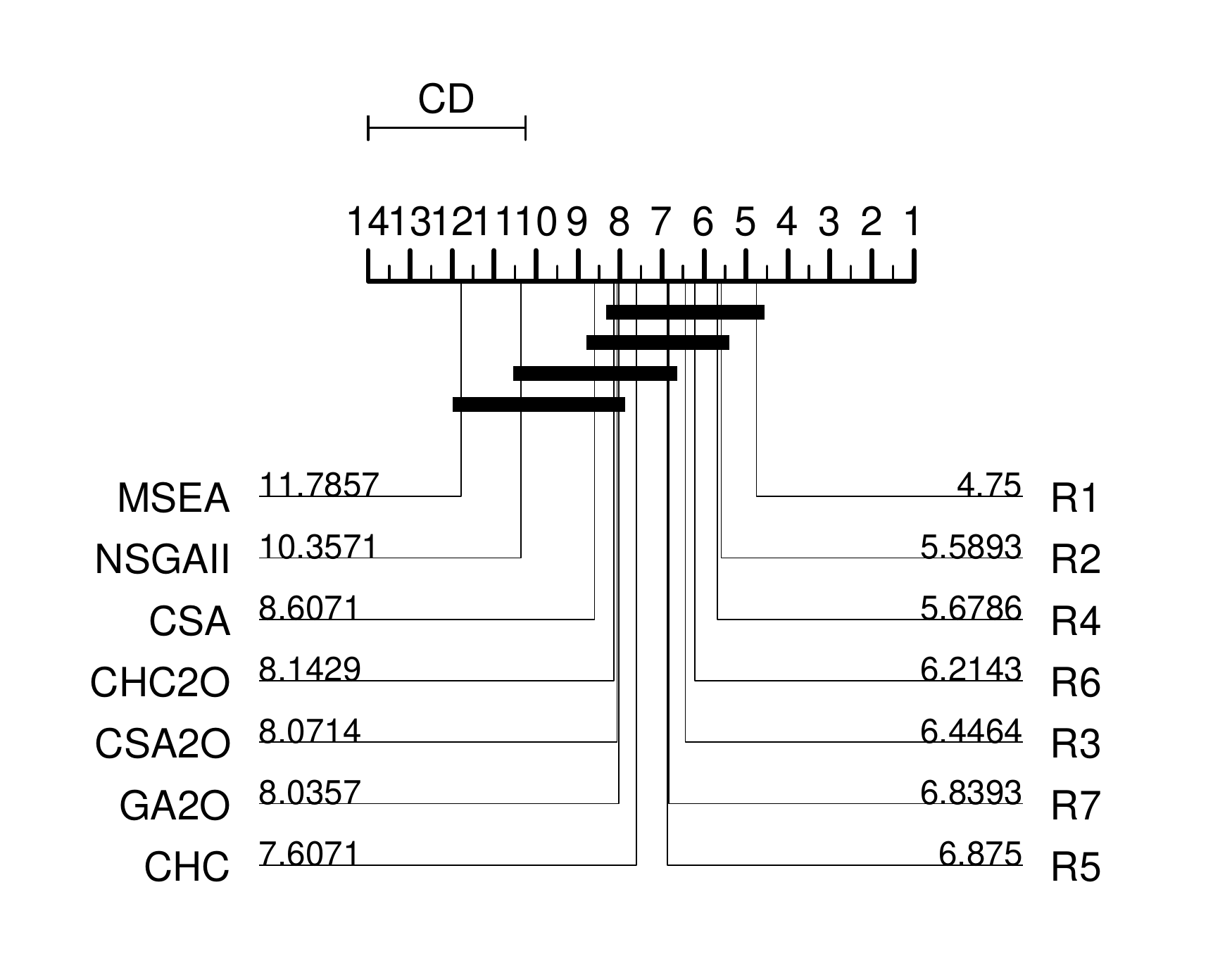}
 \caption{Critical differences of \ALGABRV~vs the compared algorithms on random networks with 81 nodes, $cd=0.3$ and $cp=0.25$.}
 \label{fig:cd81cp025}
\end{figure}

\begin{figure}[!ht]
 \centering
 \includegraphics[width=0.6\columnwidth]{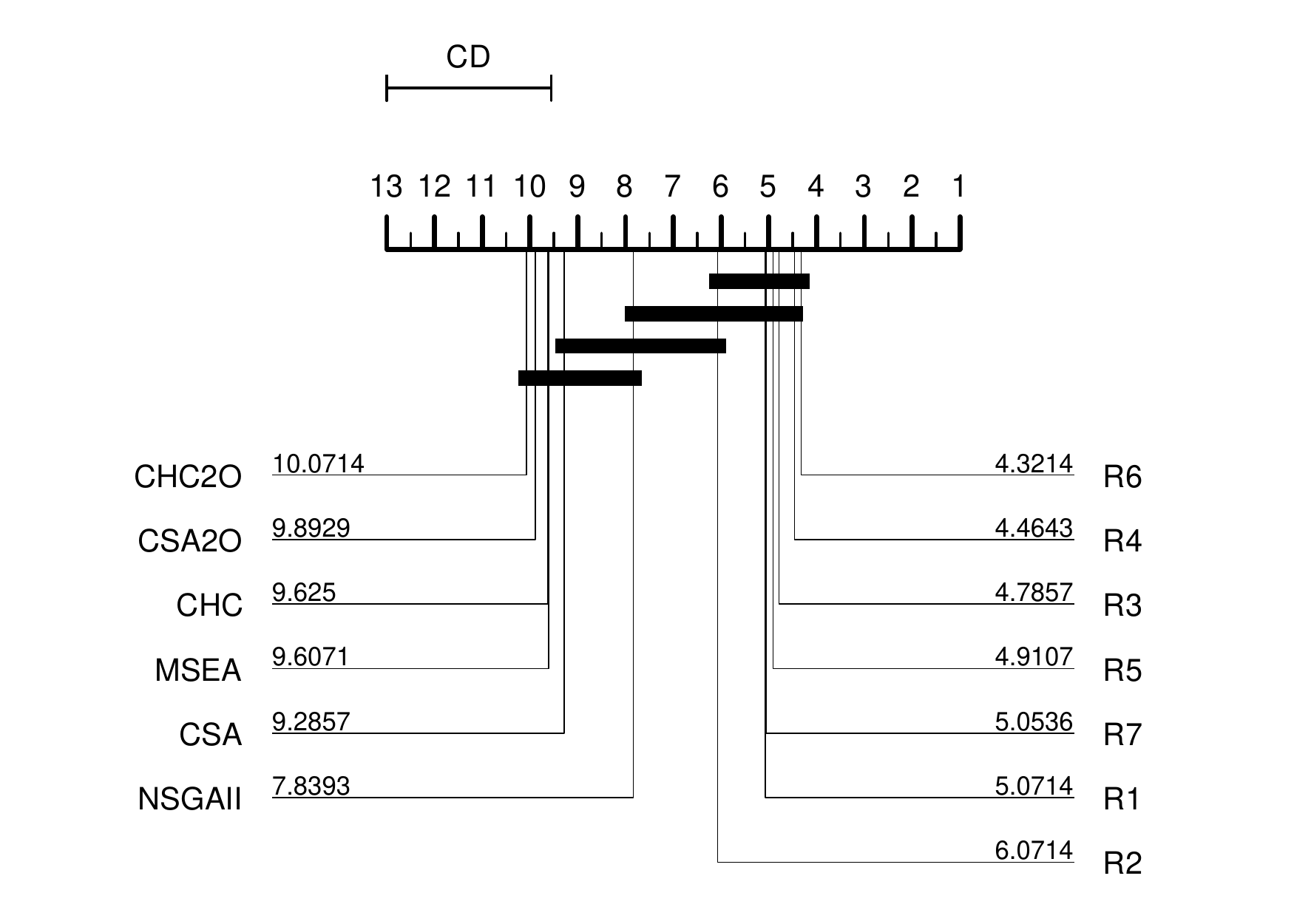}
 \caption{Critical differences of \ALGABRV~vs the compared algorithms on random networks with 81 nodes, $cd=0.3$ and $cp=0.125$.}
 \label{fig:cd81cp0125}
\end{figure}

Furthermore, we provide the pairwise comparisons (Wilcoxon rank-sum test, $N=28$, $\alpha = 0.05$ \cite{wilcoxon1992individual}) and a post-hoc analysis (Nemenyi test, $N=28$, $\alpha=0.05$ \cite{demvsar2006statistical}) of the compared algorithms. More specifically:
\begin{itemize}[leftmargin=*]
 \item Tables \ref{tab:grid9}-\ref{tab:rand81cp0125} show the $p$-values of the pairwise comparisons computed based on the number of slots used by each algorithm. We indicate the results that do not show statistical significance ($p$-value greater than $0.05$) with ``='' . Otherwise, we leave the cell empty. 
 \item Figures \ref{fig:cdGrid9}-\ref{fig:cd81cp0125} show the statistical ranking of the algorithms in the form of critical difference (CD) plots. Lower ranks indicate better performance in terms of number of slots. In the plots, algorithms connected by a thick line show statistically equivalent results. We should note that the CD plots are based on the average of the results obtained from multiple runs. Therefore, there may be some differences with the results reported in the main text, which are based on median values. Nevertheless, the CD plots support our main conclusions.
\end{itemize}
In the Tables and Figures, we exclude from the comparisons the algorithms that could not found a solution on the corresponding problem instance (see the ``-'' symbols in Table~\ref{tab:resultsComparisonSTD}).

The discussion provided in the main text is based on this statistical analysis. Mainly, the analysis shows that the various variants of \ALGABRV~demonstrate significantly better performances on the large networks, whereas the algorithms based on a centralized approach demonstrate significantly better performances on small networks.

%%%%%%%%%%%%%%%%%%%%%%%%%%%%%%%%%%%
\begin{table}[!ht]
\caption{Pairwise $p$-values of the compared algorithms on grid networks with 9 nodes.}\label{tab:grid9}
\resizebox{\columnwidth}{!}{
\begin{tabular}{|l|l|l|l|l|l|l|l|l|l|l|l|l|l|l|}
\hline
 & \textbf{R1} & \textbf{R2} & \textbf{R3} & \textbf{R4} & \textbf{R5} & \textbf{R6} & \textbf{R7} & \textbf{CHC} & \textbf{CSA} & \textbf{CHC2O} & \textbf{CSA2O} & \textbf{GA2O} & \textbf{NSGA-II} & \textbf{MSEA} \\ \hline
\textbf{R1} & & & & & & & & = & = & & & & & \\ \hline
\textbf{R2} & & & = & = & = & = & & & & & & & & \\ \hline
\textbf{R3} & & = & & & = & = & = & & & & & & & \\ \hline
\textbf{R4} & & = & & & = & & & & & & & & & \\ \hline
\textbf{R5} & & = & = & = & & = & = & & & & & & & \\ \hline
\textbf{R6} & & = & = & & = & & = & & & & & & & \\ \hline
\textbf{R7} & & & = & & = & = & & & & & & & & \\ \hline
\textbf{CHC} & = & & & & & & & & = & & & & & \\ \hline
\textbf{CSA} & = & & & & & & & = & & & & & & \\ \hline
\textbf{CHC2O} & & & & & & & & & & & = & & & \\ \hline
\textbf{CSA2O} & & & & & & & & & & = & & & & \\ \hline
\textbf{GA2O} & & & & & & & & & & & & & & \\ \hline
\textbf{NSGA-II} & & & & & & & & & & & & & & \\ \hline
\textbf{MSEA} & & & & & & & & & & & & & & \\ \hline
\end{tabular}
}
\end{table}

%%%%%%%%%%%%%%%%%%%%%%%%%%%%%%%%%%%
\begin{table}[!ht]
\caption{Pairwise $p$-values of the compared algorithms on random networks with 9 nodes, $cd=0.8$ and $cp=1$.}
\resizebox{\columnwidth}{!}{
\begin{tabular}{|l|l|l|l|l|l|l|l|l|l|l|l|l|l|l|}
\hline
 & \textbf{R1} & \textbf{R2} & \textbf{R3} & \textbf{R4} & \textbf{R5} & \textbf{R6} & \textbf{R7} & \textbf{CHC} & \textbf{CSA} & \textbf{CHC2O} & \textbf{CSA2O} & \textbf{GA2O} & \textbf{NSGA-II} & \textbf{MSEA} \\ \hline
\textbf{R1} & & & & & & & & & = & & & & & \\ \hline
\textbf{R2} & & & = & = & & & & & & & & & & \\ \hline
\textbf{R3} & & = & & = & & = & = & & & & & & & \\ \hline
\textbf{R4} & & = & = & & & & & & & & & & & \\ \hline
\textbf{R5} & & & & & & = & = & & & & & & & = \\ \hline
\textbf{R6} & & & = & & = & & = & & & & & & & \\ \hline
\textbf{R7} & & & = & & = & = & & & & & & & & \\ \hline
\textbf{CHC} & & & & & & & & & = & & & & & \\ \hline
\textbf{CSA} & = & & & & & & & = & & & & & & \\ \hline
\textbf{CHC2O} & & & & & & & & & & & = & = & = & \\ \hline
\textbf{CSA2O} & & & & & & & & & & = & & = & = & \\ \hline
\textbf{GA2O} & & & & & & & & & & = & = & & = & \\ \hline
\textbf{NSGA-II} & & & & & & & & & & = & = & = & & \\ \hline
\textbf{MSEA} & & & & & = & & & & & & & & & \\ \hline
\end{tabular}
}
\end{table}

%%%%%%%%%%%%%%%%%%%%%%%%%%%%%%%%%
\begin{table}[!ht]
\caption{Pairwise $p$-values of the compared algorithms on random networks with 9 nodes, $cd=0.8$ and $cp=0.5$.}
\resizebox{\columnwidth}{!}{
\begin{tabular}{|l|l|l|l|l|l|l|l|l|l|l|l|l|l|l|}
\hline
 & \textbf{R1} & \textbf{R2} & \textbf{R3} & \textbf{R4} & \textbf{R5} & \textbf{R6} & \textbf{R7} & \textbf{CHC} & \textbf{CSA} & \textbf{CHC2O} & \textbf{CSA2O} & \textbf{GA2O} & \textbf{NSGA-II} & \textbf{MSEA} \\ \hline
\textbf{R1} & & & & & & & & = & = & & & & & \\ \hline
\textbf{R2} & & & = & = & = & = & & & & & & & & \\ \hline
\textbf{R3} & & = & & = & = & = & = & & & & & & & \\ \hline
\textbf{R4} & & = & = & & = & = & & & & & & & & \\ \hline
\textbf{R5} & & = & = & = & & = & = & & & & & & & \\ \hline
\textbf{R6} & & = & = & = & = & & = & & & & & & & \\ \hline
\textbf{R7} & & & = & & = & = & & & & & & & & \\ \hline
\textbf{CHC} & = & & & & & & & & = & & & & & \\ \hline
\textbf{CSA} & = & & & & & & & = & & & & & & \\ \hline
\textbf{CHC2O} & & & & & & & & & & & = & = & = & \\ \hline
\textbf{CSA2O} & & & & & & & & & & = & & = & = & \\ \hline
\textbf{GA2O} & & & & & & & & & & = & = & & = & \\ \hline
\textbf{NSGA-II} & & & & & & & & & & = & = & = & & \\ \hline
\textbf{MSEA} & & & & & & & & & & & & & & \\ \hline
\end{tabular}
}
\end{table}

%%%%%%%%%%%%%%%%%%%%%%%%%%%%%%%%%
\begin{table}[!ht]
\caption{Pairwise $p$-values of the compared algorithms on random networks with 9 nodes, $cd=0.8$ and $cp=0.25$.}
\resizebox{\columnwidth}{!}{
\begin{tabular}{|l|l|l|l|l|l|l|l|l|l|l|l|l|}
\hline
 & \textbf{R1} & \textbf{R2} & \textbf{R3} & \textbf{R4} & \textbf{R5} & \textbf{R6} & \textbf{R7} & \textbf{CHC} & \textbf{CSA} & \textbf{CHC2O} & \textbf{CSA2O} & \textbf{GA2O} \\ \hline
\textbf{R1} & & & & & & & = & & & & & \\ \hline
\textbf{R2} & & & = & = & = & = & = & & & & & \\ \hline
\textbf{R3} & & = & & = & = & = & = & & & & & \\ \hline
\textbf{R4} & & = & = & & = & = & = & & & & & \\ \hline
\textbf{R5} & & = & = & = & & = & = & & & & & \\ \hline
\textbf{R6} & & = & = & = & = & & = & & & & & \\ \hline
\textbf{R7} & = & = & = & = & = & = & & = & = & & & \\ \hline
\textbf{CHC} & & & & & & & = & & = & & & \\ \hline
\textbf{CSA} & & & & & & & = & = & & & & \\ \hline
\textbf{CHC2O} & & & & & & & & & & & = & = \\ \hline
\textbf{CSA2O} & & & & & & & & & & = & & = \\ \hline
\textbf{GA2O} & & & & & & & & & & = & = & \\ \hline
\end{tabular}
}
\end{table}

%%%%%%%%%%%%%%%%%%%%%%%%%%%%%%%%%
\begin{table}[!ht]
\caption{Pairwise $p$-values of the compared algorithms on random networks with 9 nodes, $cd=0.8$ and $cp=0.125$.}
\resizebox{\columnwidth}{!}{
\begin{tabular}{|l|l|l|l|l|l|l|l|l|l|l|l|l|}
\hline
 & \textbf{R1} & \textbf{R2} & \textbf{R3} & \textbf{R4} & \textbf{R5} & \textbf{R6} & \textbf{R7} & \textbf{CHC} & \textbf{CSA} & \textbf{CHC2O} & \textbf{CSA2O} & \textbf{GA2O} \\ \hline
\textbf{R1} & & & & & & & = & & & & & \\ \hline
\textbf{R2} & & & = & = & & = & = & & & & & \\ \hline
\textbf{R3} & & = & & = & & = & = & & & & & \\ \hline
\textbf{R4} & & = & = & & & = & = & & & & & \\ \hline
\textbf{R5} & & & & & & & = & & & & & \\ \hline
\textbf{R6} & & = & = & = & & & = & & & & & \\ \hline
\textbf{R7} & = & = & = & = & = & = & & & & & & \\ \hline
\textbf{CHC} & & & & & & & & & = & & & \\ \hline
\textbf{CSA} & & & & & & & & = & & & & \\ \hline
\textbf{CHC2O} & & & & & & & & & & & & = \\ \hline
\textbf{CSA2O} & & & & & & & & & & & & \\ \hline
\textbf{GA2O} & & & & & & & & & & = & & \\ \hline
\end{tabular}
}
\end{table}

%%%%%%%%%%%%%%%%%%%%%%%%%%%%%%%%%
\begin{table}[!ht]
\caption{Pairwise $p$-values of the compared algorithms on grid networks with 36 nodes.}
\resizebox{\columnwidth}{!}{
\begin{tabular}{|l|l|l|l|l|l|l|l|l|l|l|l|l|l|l|}
\hline
 & \textbf{R1} & \textbf{R2} & \textbf{R3} & \textbf{R4} & \textbf{R5} & \textbf{R6} & \textbf{R7} & \textbf{CHC} & \textbf{CSA} & \textbf{CHC2O} & \textbf{CSA2O} & \textbf{GA2O} & \textbf{NSGA-II} & \textbf{MSEA} \\ \hline
\textbf{R1} & & & & = & & = & & & & & & & & \\ \hline
\textbf{R2} & & & = & = & = & = & & & & & & & = & \\ \hline
\textbf{R3} & & = & & = & = & = & & & & & & & = & \\ \hline
\textbf{R4} & = & = & = & & = & = & = & = & = & = & = & = & = & = \\ \hline
\textbf{R5} & & = & = & = & & = & & & & & & & & \\ \hline
\textbf{R6} & = & = & = & = & = & & = & = & = & & & = & = & = \\ \hline
\textbf{R7} & & & & = & & = & & & & & & & & \\ \hline
\textbf{CHC} & & & & = & & = & & & = & & & & & \\ \hline
\textbf{CSA} & & & & = & & = & & = & & & & & & \\ \hline
\textbf{CHC2O} & & & & = & & & & & & & = & & & \\ \hline
\textbf{CSA2O} & & & & = & & & & & & = & & & & \\ \hline
\textbf{GA2O} & & & & = & & = & & & & & & & & \\ \hline
\textbf{NSGA-II} & & = & = & = & & = & & & & & & & & \\ \hline
\textbf{MSEA} & & & & = & & = & & & & & & & & \\ \hline
\end{tabular}
}
\end{table}

%%%%%%%%%%%%%%%%%%%%%%%%%%%%%%%%%
\begin{table}[!ht]
\caption{Pairwise $p$-values of the compared algorithms on random networks with 36 nodes, $cd=0.5$ and $cp=1$.}
\resizebox{\columnwidth}{!}{
\begin{tabular}{|l|l|l|l|l|l|l|l|l|l|l|l|l|l|l|}
\hline
 & \textbf{R1} & \textbf{R2} & \textbf{R3} & \textbf{R4} & \textbf{R5} & \textbf{R6} & \textbf{R7} & \textbf{CHC} & \textbf{CSA} & \textbf{CHC2O} & \textbf{CSA2O} & \textbf{GA2O} & \textbf{NSGA-II} & \textbf{MSEA} \\ \hline
\textbf{R1} & & = & = & = & = & = & = & & & = & = & & = & \\ \hline
\textbf{R2} & = & & = & = & = & = & = & = & & = & = & = & = & = \\ \hline
\textbf{R3} & = & = & & = & = & = & = & = & = & = & = & = & = & = \\ \hline
\textbf{R4} & = & = & = & & = & = & = & = & & = & = & = & = & = \\ \hline
\textbf{R5} & = & = & = & = & & = & = & & & = & = & & = & \\ \hline
\textbf{R6} & = & = & = & = & = & & = & = & & = & = & = & = & = \\ \hline
\textbf{R7} & = & = & = & = & = & = & & = & & & & = & = & \\ \hline
\textbf{CHC} & & = & = & = & & = & = & & = & & & = & & = \\ \hline
\textbf{CSA} & & & = & & & & & = & & & & & & = \\ \hline
\textbf{CHC2O} & = & = & = & = & = & = & & & & & = & & & \\ \hline
\textbf{CSA2O} & = & = & = & = & = & = & & & & = & & & & \\ \hline
\textbf{GA2O} & & = & = & = & & = & = & = & & & & & & \\ \hline
\textbf{NSGA-II} & = & = & = & = & = & = & = & & & & & & & \\ \hline
\textbf{MSEA} & & = & = & = & & = & & = & = & & & & & \\ \hline
\end{tabular}
}
\end{table}

%%%%%%%%%%%%%%%%%%%%%%%%%%%%%%%%%
\begin{table}[!ht]
\caption{Pairwise $p$-values of the compared algorithms on random networks with 36 nodes, $cd=0.5$ and $cp=0.5$.}
\resizebox{\columnwidth}{!}{
\begin{tabular}{|l|l|l|l|l|l|l|l|l|l|l|l|l|l|l|}
\hline
 & \textbf{R1} & \textbf{R2} & \textbf{R3} & \textbf{R4} & \textbf{R5} & \textbf{R6} & \textbf{R7} & \textbf{CHC} & \textbf{CSA} & \textbf{CHC2O} & \textbf{CSA2O} & \textbf{GA2O} & \textbf{NSGA-II} & \textbf{MSEA} \\ \hline
\textbf{R1} & & = & = & = & & & = & & & & & & = & \\ \hline
\textbf{R2} & = & & = & = & & = & = & & & = & = & = & = & \\ \hline
\textbf{R3} & = & = & & = & & = & = & & & = & = & = & = & \\ \hline
\textbf{R4} & = & = & = & & & = & = & & & = & = & = & = & \\ \hline
\textbf{R5} & & & & & & = & & & & = & = & = & & \\ \hline
\textbf{R6} & & = & = & = & = & & = & & & = & = & = & & \\ \hline
\textbf{R7} & = & = & = & = & & = & & & = & & & & = & \\ \hline
\textbf{CHC} & & & & & & & & & = & & & & & = \\ \hline
\textbf{CSA} & & & & & & & = & = & & & & & & = \\ \hline
\textbf{CHC2O} & & = & = & = & = & = & & & & & = & = & & \\ \hline
\textbf{CSA2O} & & = & = & = & = & = & & & & = & & = & & \\ \hline
\textbf{GA2O} & & = & = & = & = & = & & & & = & = & & & \\ \hline
\textbf{NSGA-II} & = & = & = & = & & & = & & & & & & & \\ \hline
\textbf{MSEA} & & & & & & & & = & = & & & & & \\ \hline
\end{tabular}
}
\end{table}

%%%%%%%%%%%%%%%%%%%%%%%%%%%%%%%%%
\begin{table}[!ht]
\caption{Pairwise $p$-values of the compared algorithms on random networks with 36 nodes, $cd=0.5$ and $cp=0.25$.}
\resizebox{\columnwidth}{!}{
\begin{tabular}{|l|l|l|l|l|l|l|l|l|l|l|l|l|l|l|}
\hline
 & \textbf{R1} & \textbf{R2} & \textbf{R3} & \textbf{R4} & \textbf{R5} & \textbf{R6} & \textbf{R7} & \textbf{CHC} & \textbf{CSA} & \textbf{CHC2O} & \textbf{CSA2O} & \textbf{GA2O} & \textbf{NSGA-II} & \textbf{MSEA} \\ \hline
\textbf{R1} & & & = & = & & & = & & & & & = & & \\ \hline
\textbf{R2} & & & = & = & = & = & = & & & = & = & & & \\ \hline
\textbf{R3} & = & = & & = & = & = & = & = & = & = & = & = & = & = \\ \hline
\textbf{R4} & = & = & = & & = & = & = & & & = & = & & = & \\ \hline
\textbf{R5} & & = & = & = & & = & & & & & & & & \\ \hline
\textbf{R6} & & = & = & = & = & & = & & & & & & & \\ \hline
\textbf{R7} & = & = & = & = & & = & & & & = & = & & = & \\ \hline
\textbf{CHC} & & & = & & & & & & = & & & & & = \\ \hline
\textbf{CSA} & & & = & & & & & = & & & & & & = \\ \hline
\textbf{CHC2O} & & = & = & = & & & = & & & & = & & & \\ \hline
\textbf{CSA2O} & & = & = & = & & & = & & & = & & & = & \\ \hline
\textbf{GA2O} & = & & = & & & & & & & & & & & \\ \hline
\textbf{NSGA-II} & & & = & = & & & = & & & & = & & & \\ \hline
\textbf{MSEA} & & & = & & & & & = & = & & & & & \\ \hline
\end{tabular}
}
\end{table}

%%%%%%%%%%%%%%%%%%%%%%%%%%%%%%%%%
\begin{table}[!ht]
\caption{Pairwise $p$-values of the compared algorithms on random networks with 36 nodes, $cd=0.5$ and $cp=0.125$.}
\resizebox{\columnwidth}{!}{
\begin{tabular}{|l|l|l|l|l|l|l|l|l|l|l|l|l|l|}
\hline
 & \textbf{R1} & \textbf{R2} & \textbf{R3} & \textbf{R4} & \textbf{R5} & \textbf{R6} & \textbf{R7} & \textbf{CHC} & \textbf{CSA} & \textbf{CHC2O} & \textbf{CSA2O} & \textbf{NSGA-II} & \textbf{MSEA} \\ \hline
\textbf{R1} & & & & & = & = & & & & = & = & & \\ \hline
\textbf{R2} & & & = & = & = & = & = & & & & & & \\ \hline
\textbf{R3} & & = & & = & = & = & = & & & & & & \\ \hline
\textbf{R4} & & = & = & & = & = & = & & & & & & \\ \hline
\textbf{R5} & = & = & = & = & & = & = & = & = & = & = & = & = \\ \hline
\textbf{R6} & = & = & = & = & = & & = & = & = & = & = & = & = \\ \hline
\textbf{R7} & & = & = & = & = & = & & & & & & & \\ \hline
\textbf{CHC} & & & & & = & = & & & = & & & & \\ \hline
\textbf{CSA} & & & & & = & = & & = & & & & & \\ \hline
\textbf{CHC2O} & = & & & & = & = & & & & & = & & \\ \hline
\textbf{CSA2O} & = & & & & = & = & & & & = & & & \\ \hline
\textbf{NSGA-II} & & & & & = & = & & & & & & & \\ \hline
\textbf{MSEA} & & & & & = & = & & & & & & & \\ \hline
\end{tabular}
}
\end{table}

%%%%%%%%%%%%%%%%%%%%%%%%%%%%%%%%%
\begin{table}[!ht]
\caption{Pairwise $p$-values of the compared algorithms on grid networks with 81 nodes.}
\resizebox{\columnwidth}{!}{
\begin{tabular}{|l|l|l|l|l|l|l|l|l|l|l|l|l|l|l|}
\hline
 & \textbf{R1} & \textbf{R2} & \textbf{R3} & \textbf{R4} & \textbf{R5} & \textbf{R6} & \textbf{R7} & \textbf{CHC} & \textbf{CSA} & \textbf{CHC2O} & \textbf{CSA2O} & \textbf{GA2O} & \textbf{NSGA-II} & \textbf{MSEA} \\ \hline
\textbf{R1} & & & & & & & & & & & & = & & \\ \hline
\textbf{R2} & & & = & & & & & & & & & & & \\ \hline
\textbf{R3} & & = & & = & & & & & & & & & & \\ \hline
\textbf{R4} & & & = & & & = & & & & & & & & \\ \hline
\textbf{R5} & & & & & & & = & & & = & = & & & \\ \hline
\textbf{R6} & & & & = & & & & & & & & & & \\ \hline
\textbf{R7} & & & & & = & & & & & = & = & & & \\ \hline
\textbf{CHC} & & & & & & & & & = & & & = & = & \\ \hline
\textbf{CSA} & & & & & & & & = & & & & & = & \\ \hline
\textbf{CHC2O} & & & & & = & & = & & & & = & & & \\ \hline
\textbf{CSA2O} & & & & & = & & = & & & = & & & & \\ \hline
\textbf{GA2O} & = & & & & & & & = & & & & & & \\ \hline
\textbf{NSGA-II} & & & & & & & & = & = & & & & & \\ \hline
\textbf{MSEA} & & & & & & & & & & & & & & \\ \hline
\end{tabular}
}
\end{table}

%%%%%%%%%%%%%%%%%%%%%%%%%%%%%%%%%
\begin{table}[!ht]
\caption{Pairwise $p$-values of the compared algorithms on random networks with 81 nodes, $cd=0.3$ and $cp=1$.}
\resizebox{\columnwidth}{!}{
\begin{tabular}{|l|l|l|l|l|l|l|l|l|l|l|l|l|l|}
\hline
 & \textbf{R1} & \textbf{R2} & \textbf{R3} & \textbf{R4} & \textbf{R5} & \textbf{R7} & \textbf{CHC} & \textbf{CSA} & \textbf{CHC2O} & \textbf{CSA2O} & \textbf{GA2O} & \textbf{NSGA-II} & \textbf{MSEA} \\ \hline
\textbf{R1} & & & & & & & = & = & = & = & = & = & = \\ \hline
\textbf{R2} & & & = & = & = & = & & & & & & & \\ \hline
\textbf{R3} & & = & & = & = & = & & & & & & & \\ \hline
\textbf{R4} & & = & = & & = & = & & & & & & & \\ \hline
\textbf{R5} & & = & = & = & & = & & & & & & & \\ \hline
\textbf{R7} & & = & = & = & = & & & & & & & & \\ \hline
\textbf{CHC} & = & & & & & & & = & & & & & \\ \hline
\textbf{CSA} & = & & & & & & = & & & & & & \\ \hline
\textbf{CHC2O} & = & & & & & & & & & = & = & & = \\ \hline
\textbf{CSA2O} & = & & & & & & & & = & & = & & = \\ \hline
\textbf{GA2O} & = & & & & & & & & = & = & & & = \\ \hline
\textbf{NSGA-II} & = & & & & & & & & & & & & \\ \hline
\textbf{MSEA} & = & & & & & & & & = & = & = & & \\ \hline
\end{tabular}
}
\end{table}

%%%%%%%%%%%%%%%%%%%%%%%%%%%%%%%%%
\begin{table}[!ht]
\caption{Pairwise $p$-values of the compared algorithms on random networks with 81 nodes, $cd=0.3$ and $cp=0.5$.}
\resizebox{\columnwidth}{!}{
\begin{tabular}{|l|l|l|l|l|l|l|l|l|l|l|l|l|l|l|}
\hline
 & \textbf{R1} & \textbf{R2} & \textbf{R3} & \textbf{R4} & \textbf{R5} & \textbf{R6} & \textbf{R7} & \textbf{CHC} & \textbf{CSA} & \textbf{CHC2O} & \textbf{CSA2O} & \textbf{GA2O} & \textbf{NSGA-II} & \textbf{MSEA} \\ \hline
\textbf{R1} & & = & & = & & & & = & = & & & & & \\ \hline
\textbf{R2} & = & & & = & = & & = & = & = & & & & & \\ \hline
\textbf{R3} & & & & & & = & & & & = & = & = & = & = \\ \hline
\textbf{R4} & = & = & & & = & = & = & = & = & & = & & & \\ \hline
\textbf{R5} & & = & & = & & & = & & & & & & & \\ \hline
\textbf{R6} & & & = & = & & & & & & = & = & = & & \\ \hline
\textbf{R7} & & = & & = & = & & & & & & & & & \\ \hline
\textbf{CHC} & = & = & & = & & & & & = & & & & & \\ \hline
\textbf{CSA} & = & = & & = & & & & = & & & & & & \\ \hline
\textbf{CHC2O} & & & = & & & = & & & & & = & & & \\ \hline
\textbf{CSA2O} & & & = & = & & = & & & & = & & & & \\ \hline
\textbf{GA2O} & & & = & & & = & & & & & & & & \\ \hline
\textbf{NSGA-II} & & & = & & & & & & & & & & & \\ \hline
\textbf{MSEA} & & & = & & & & & & & & & & & \\ \hline
\end{tabular}
}
\end{table}

%%%%%%%%%%%%%%%%%%%%%%%%%%%%%%%%%
\begin{table}[!ht]
\caption{Pairwise $p$-values of the compared algorithms on random networks with 81 nodes, $cd=0.3$ and $cp=0.25$.}
\resizebox{\columnwidth}{!}{
\begin{tabular}{|l|l|l|l|l|l|l|l|l|l|l|l|l|l|l|}
\hline
 & \textbf{R1} & \textbf{R2} & \textbf{R3} & \textbf{R4} & \textbf{R5} & \textbf{R6} & \textbf{R7} & \textbf{CHC} & \textbf{CSA} & \textbf{CHC2O} & \textbf{CSA2O} & \textbf{GA2O} & \textbf{NSGA-II} & \textbf{MSEA} \\ \hline
\textbf{R1} & & = & = & = & = & = & = & & & & & & & \\ \hline
\textbf{R2} & = & & = & = & = & = & = & = & & & & & & \\ \hline
\textbf{R3} & = & = & & = & = & = & = & = & & = & = & = & & \\ \hline
\textbf{R4} & = & = & = & & = & = & = & = & & & & & & \\ \hline
\textbf{R5} & = & = & = & = & & = & = & = & = & = & = & = & = & = \\ \hline
\textbf{R6} & = & = & = & = & = & & = & = & & & & & & \\ \hline
\textbf{R7} & = & = & = & = & = & = & & = & = & & & & & \\ \hline
\textbf{CHC} & & = & = & = & = & = & = & & = & = & = & = & = & \\ \hline
\textbf{CSA} & & & & & = & & = & = & & & & & & \\ \hline
\textbf{CHC2O} & & & = & & = & & & = & & & = & = & & \\ \hline
\textbf{CSA2O} & & & = & & = & & & = & & = & & = & & \\ \hline
\textbf{GA2O} & & & = & & = & & & = & & = & = & & & \\ \hline
\textbf{NSGA-II} & & & & & = & & & = & & & & & & \\ \hline
\textbf{MSEA} & & & & & = & & & & & & & & & \\ \hline
\end{tabular}
}
\end{table}

%%%%%%%%%%%%%%%%%%%%%%%%%%%%%%%%%
\begin{table}[!ht]
\caption{Pairwise $p$-values of the compared algorithms on random networks with 81 nodes, $cd=0.3$ and $cp=0.125$.}\label{tab:rand81cp0125}
\resizebox{\columnwidth}{!}{
\begin{tabular}{|l|l|l|l|l|l|l|l|l|l|l|l|l|l|}
\hline
 & \textbf{R1} & \textbf{R2} & \textbf{R3} & \textbf{R4} & \textbf{R5} & \textbf{R6} & \textbf{R7} & \textbf{CHC} & \textbf{CSA} & \textbf{CHC2O} & \textbf{CSA2O} & \textbf{NSGA-II} & \textbf{MSEA} \\ \hline
\textbf{R1} & & = & & = & & & & & & & & & \\ \hline
\textbf{R2} & = & & = & = & = & = & = & = & = & = & = & = & = \\ \hline
\textbf{R3} & & = & & = & = & = & = & & & & & & \\ \hline
\textbf{R4} & = & = & = & & = & = & = & & & & & & \\ \hline
\textbf{R5} & & = & = & = & & = & = & & & & & & \\ \hline
\textbf{R6} & & = & = & = & = & & = & & & & & & \\ \hline
\textbf{R7} & & = & = & = & = & = & & & & & & & \\ \hline
\textbf{CHC} & & = & & & & & & & = & & & & = \\ \hline
\textbf{CSA} & & = & & & & & & = & & & & & = \\ \hline
\textbf{CHC2O} & & = & & & & & & & & & & & \\ \hline
\textbf{CSA2O} & & = & & & & & & & & & & & \\ \hline
\textbf{NSGA-II} & & = & & & & & & & & & & & \\ \hline
\textbf{MSEA} & & = & & & & & & = & = & & & & \\ \hline
\end{tabular}
}
\end{table}

%%%%%%%%%%%%%%%%%%%%%%%%%%%%%%%%%%%%%%%%%%%%%%%%%%%%%%%%%%%%%%%%%%

\clearpage

%%%%%%%% 9 Grid %%%%%%%%%%%%%%%%

%%%%% 9cp1 %%%%%%%%%%%%%%%%%%%%%%

%%%%% 9cp05 %%%%%%%%%%%%%%%%%%%%%%

%%%%% 9cp025 %%%%%%%%%%%%%%%%%%%%%%

%%%%% 9cp0125 %%%%%%%%%%%%%%%%%%%%%%

%%%%% 36Grid %%%%%%%%%%%%%%%%%%%%%%

%%%%% 36cp1 %%%%%%%%%%%%%%%%%%%%%%

%%%%% 36cp05 %%%%%%%%%%%%%%%%%%%%%%

%%%%% 36cp025 %%%%%%%%%%%%%%%%%%%%%%

%%%%% 36cp0125 %%%%%%%%%%%%%%%%%%%%%%

%%%%% 81Grid %%%%%%%%%%%%%%%%%%%%%%

%%%%% 81cp1 %%%%%%%%%%%%%%%%%%%%%%

%%%%% 81cp05 %%%%%%%%%%%%%%%%%%%%%%

%%%%% 81cp025 %%%%%%%%%%%%%%%%%%%%%%

%%%%% 81cp0125 %%%%%%%%%%%%%%%%%%%%%%

\clearpage

\section{Runtime behavior of the evolutionary process}

\begin{figure*}[!hb]
\begin{subfigures}
\subfloat[Delivery rate (Add 36cp1)]{\includegraphics[width=0.33\columnwidth]{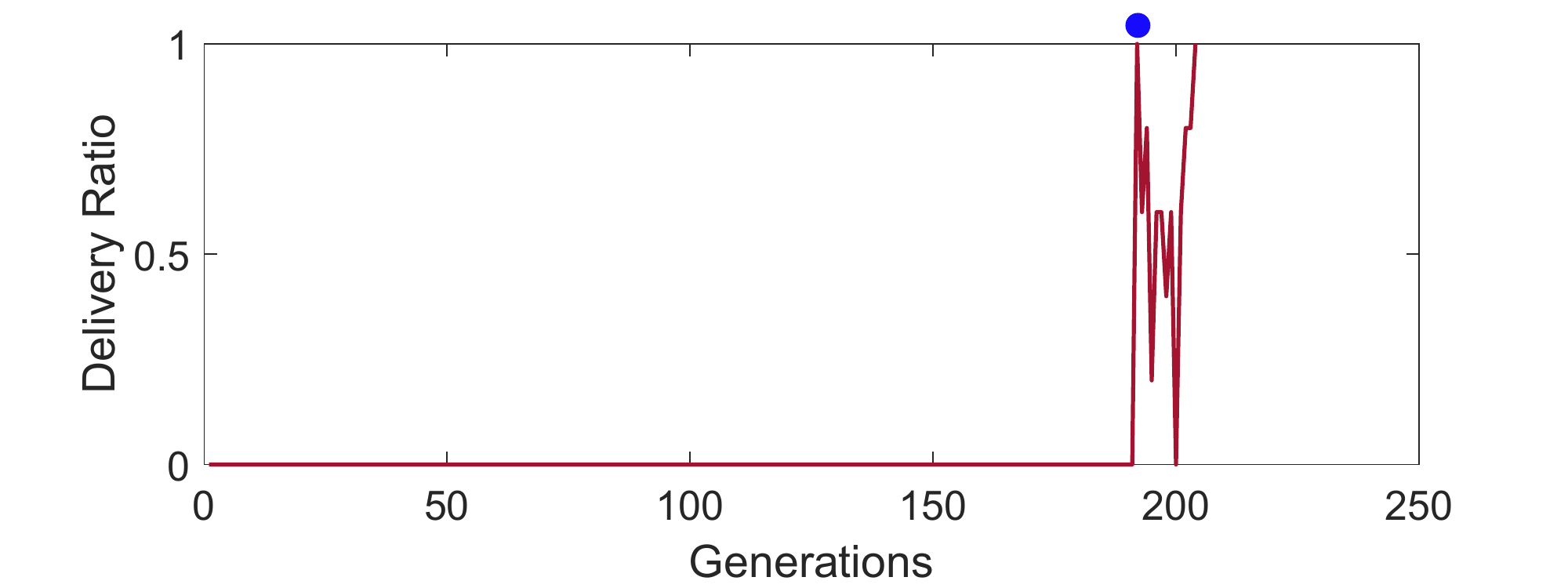}\label{fig:deliveryRatioAdd1}}
\subfloat[Node-1 fitness (Add 36cp1)]{\includegraphics[width=0.33\columnwidth]{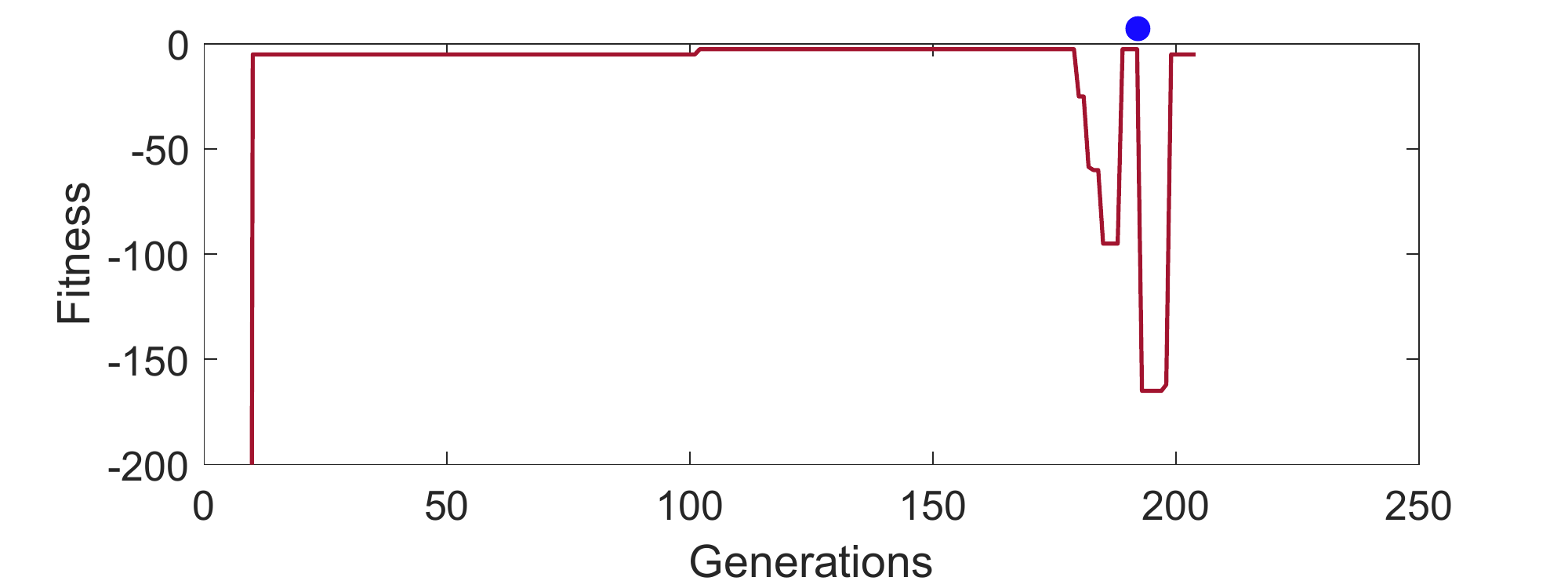}\label{fig:a1Add1}}
\subfloat[Node-2 fitness (Add 36cp1)]{\includegraphics[width=0.33\columnwidth]{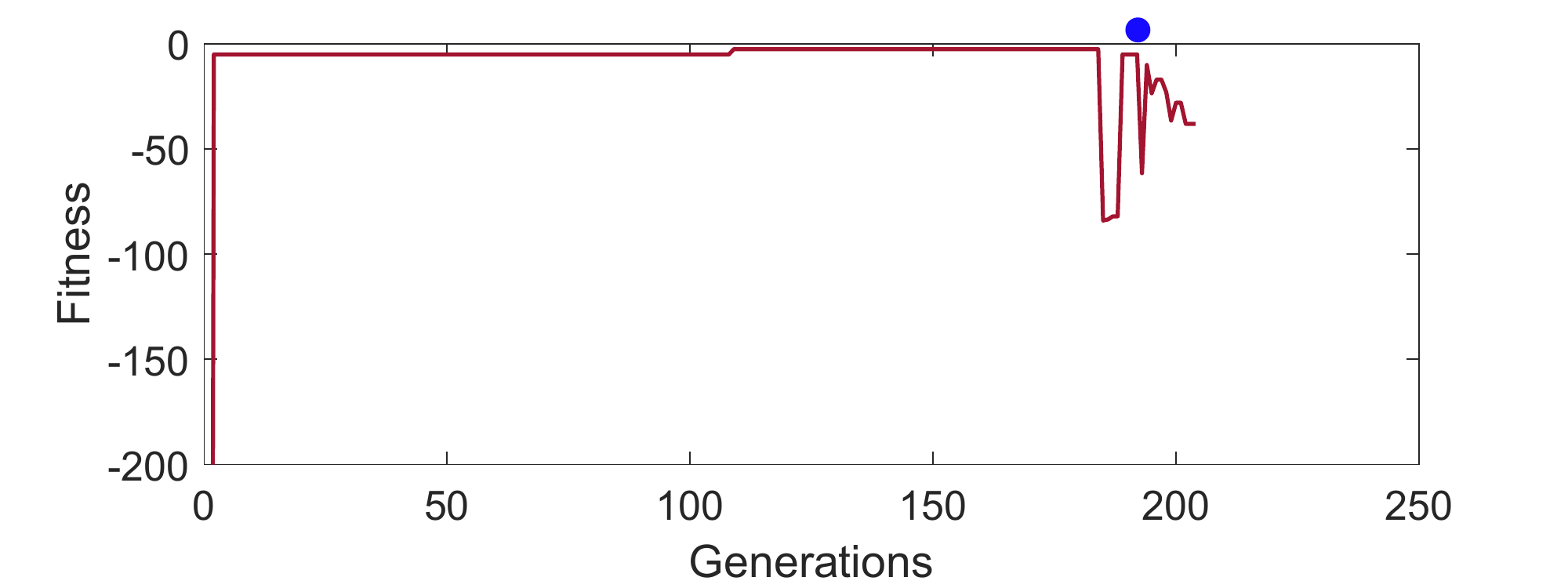}\label{fig:a2Add1}}

\subfloat[Delivery rate (Relocate 36cp05)]{\includegraphics[width=0.33\columnwidth]{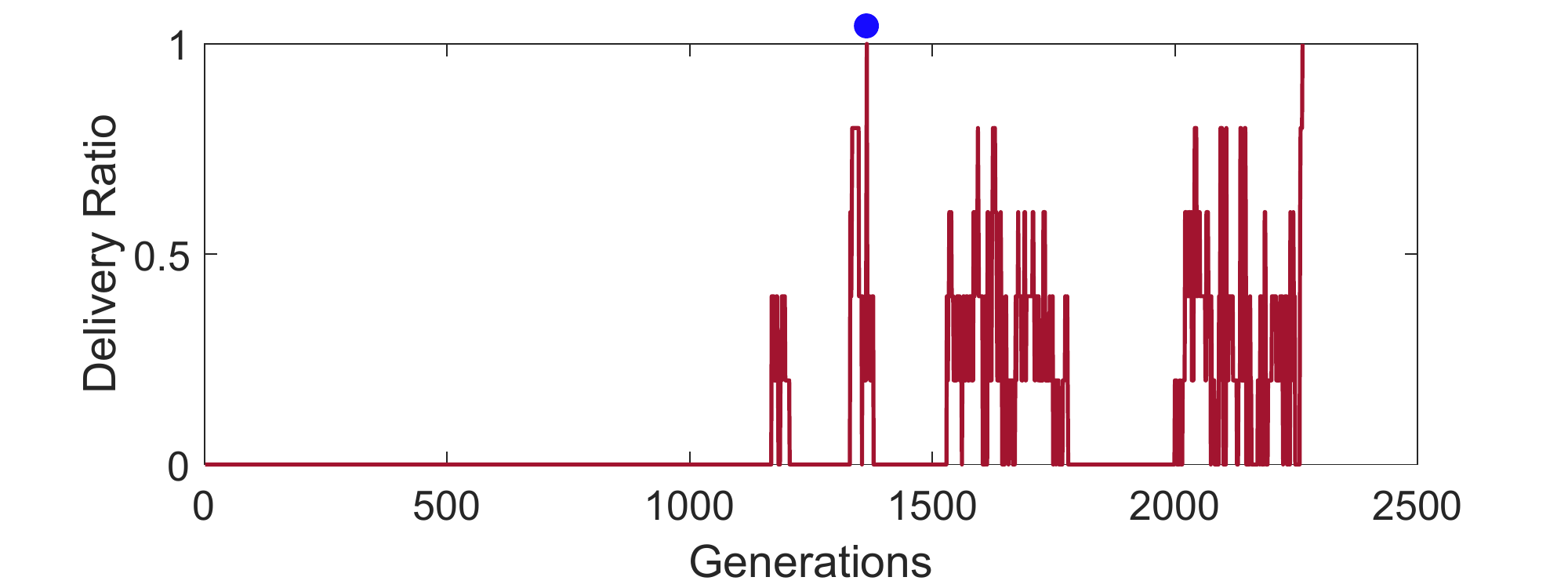}\label{fig:deliveryRatio05}}
\subfloat[Node-1 fitness (Relocate 36cp05)]{\includegraphics[width=0.33\columnwidth]{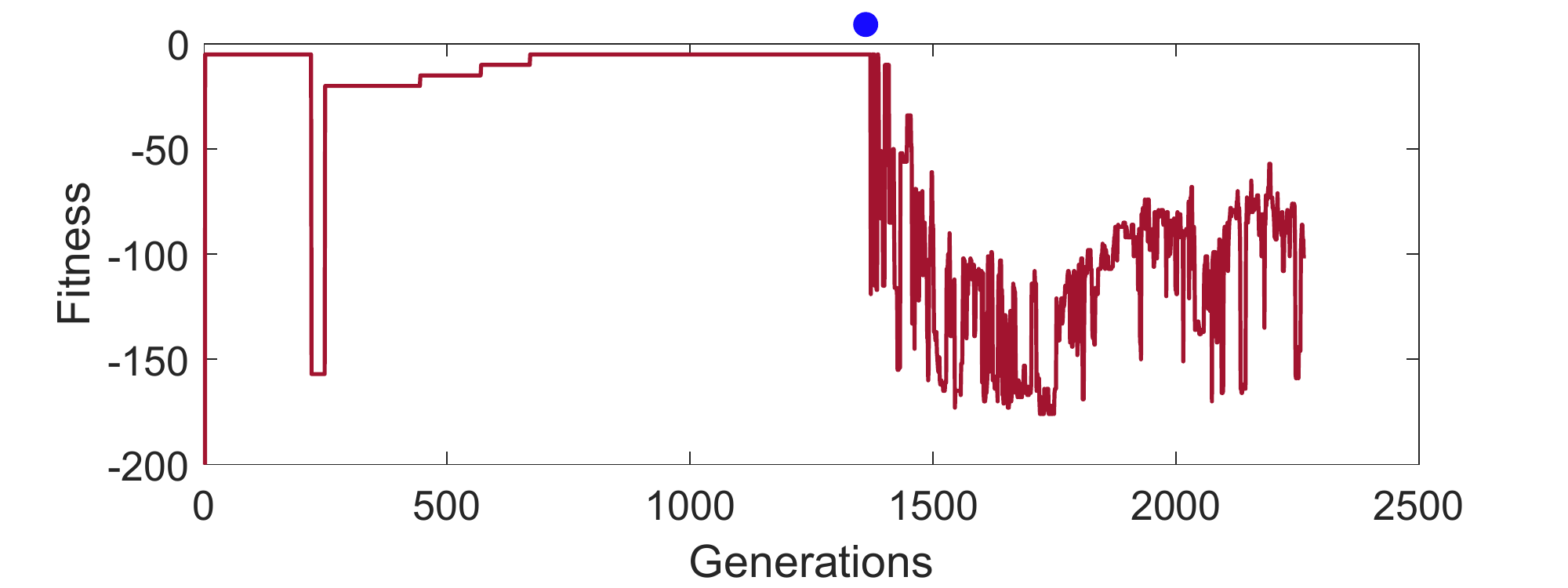}\label{fig:a1Relocate05}}
\subfloat[Node-2 fitness (Relocate 36cp05)]{\includegraphics[width=0.33\columnwidth]{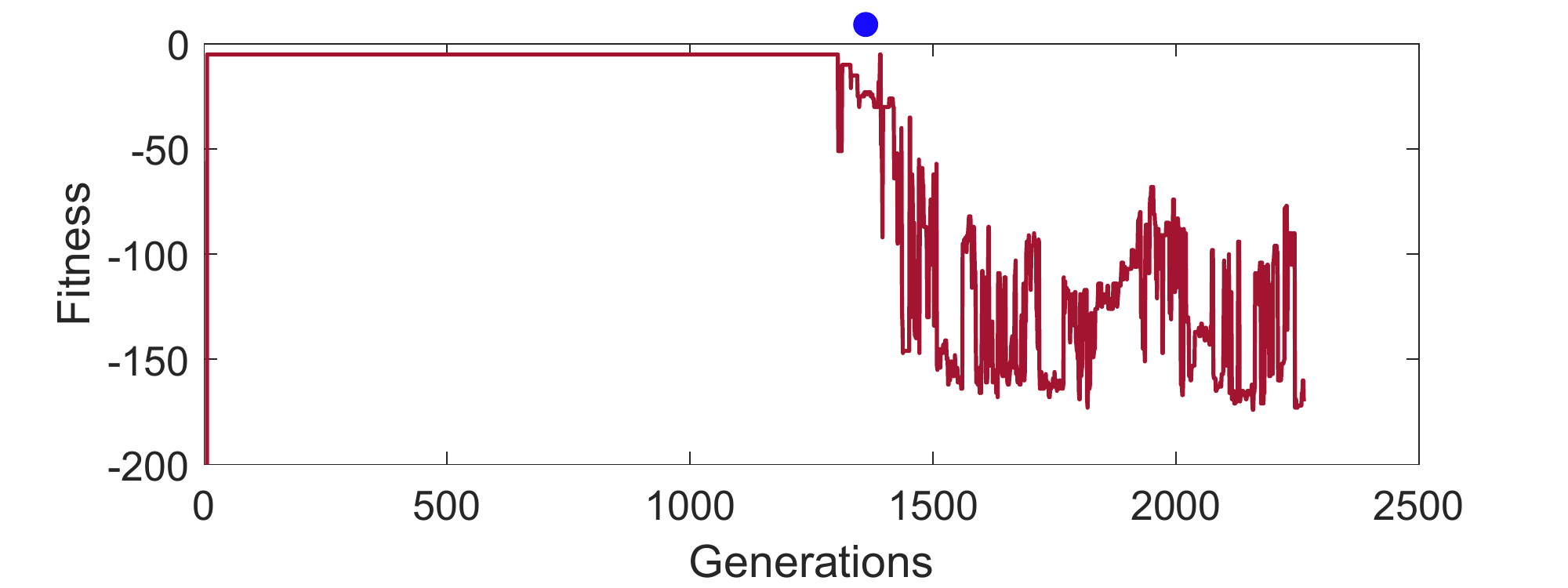}\label{fig:a2Relocate05}}

\subfloat[Delivery rate (Remove 36cp025)]{\includegraphics[width=0.33\columnwidth]{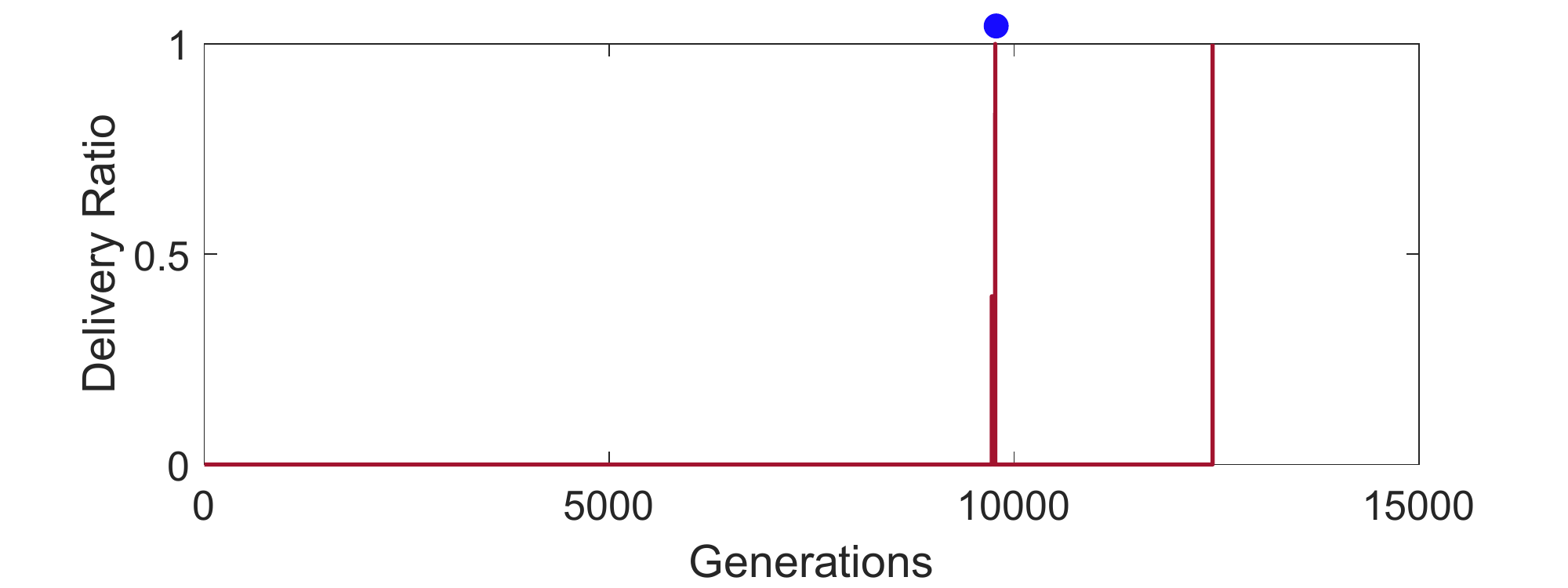}\label{fig:deliveryRatioRemove025}}
\subfloat[Node-1 fitness (Remove 36cp025)]{\includegraphics[width=0.33\columnwidth]{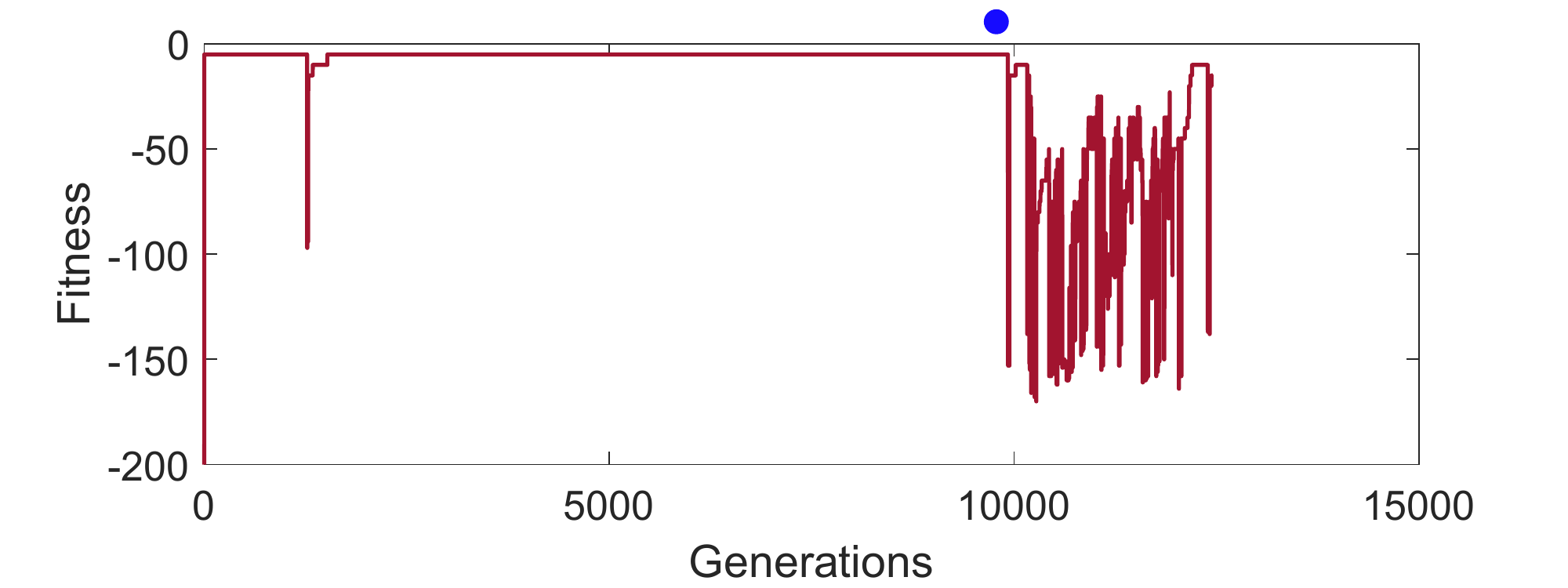}\label{fig:a1Remove025}}
\subfloat[Node-2 fitness (Remove 36cp025)]{\includegraphics[width=0.33\columnwidth]{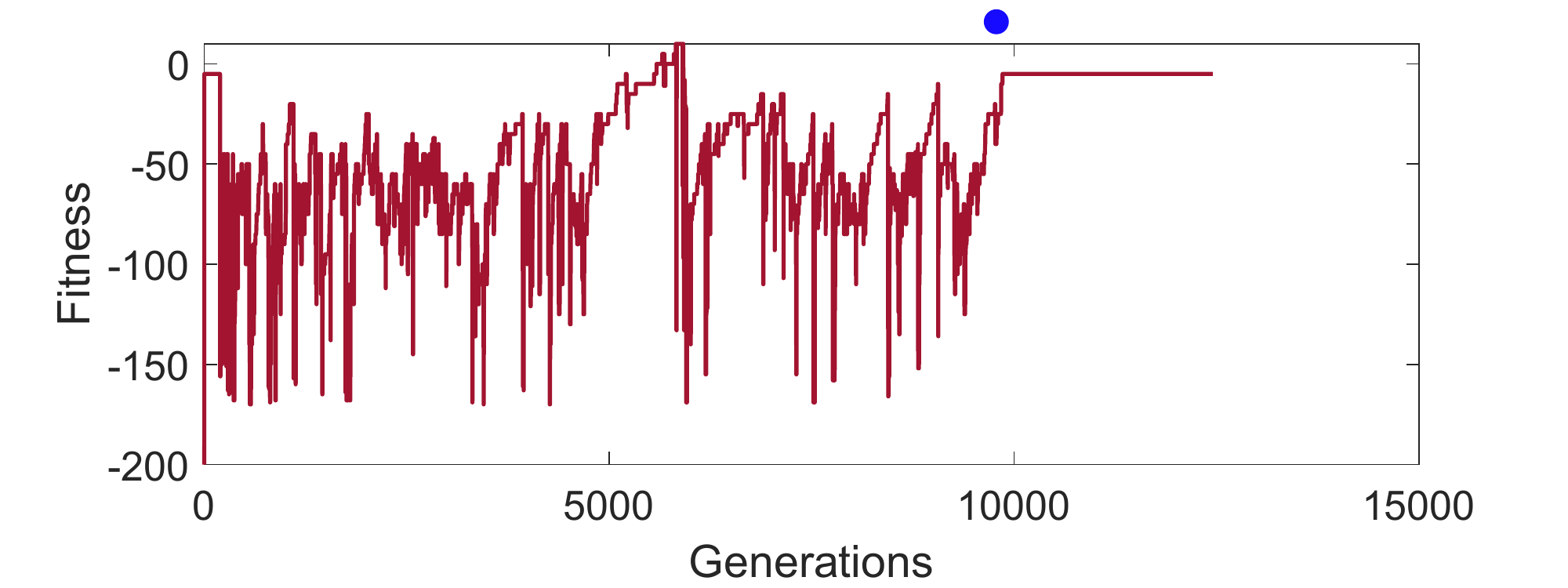}\label{fig:a2Remove025}}

\subfloat[Delivery rate (Reinit. 36cp0125)]{\includegraphics[width=0.33\columnwidth]{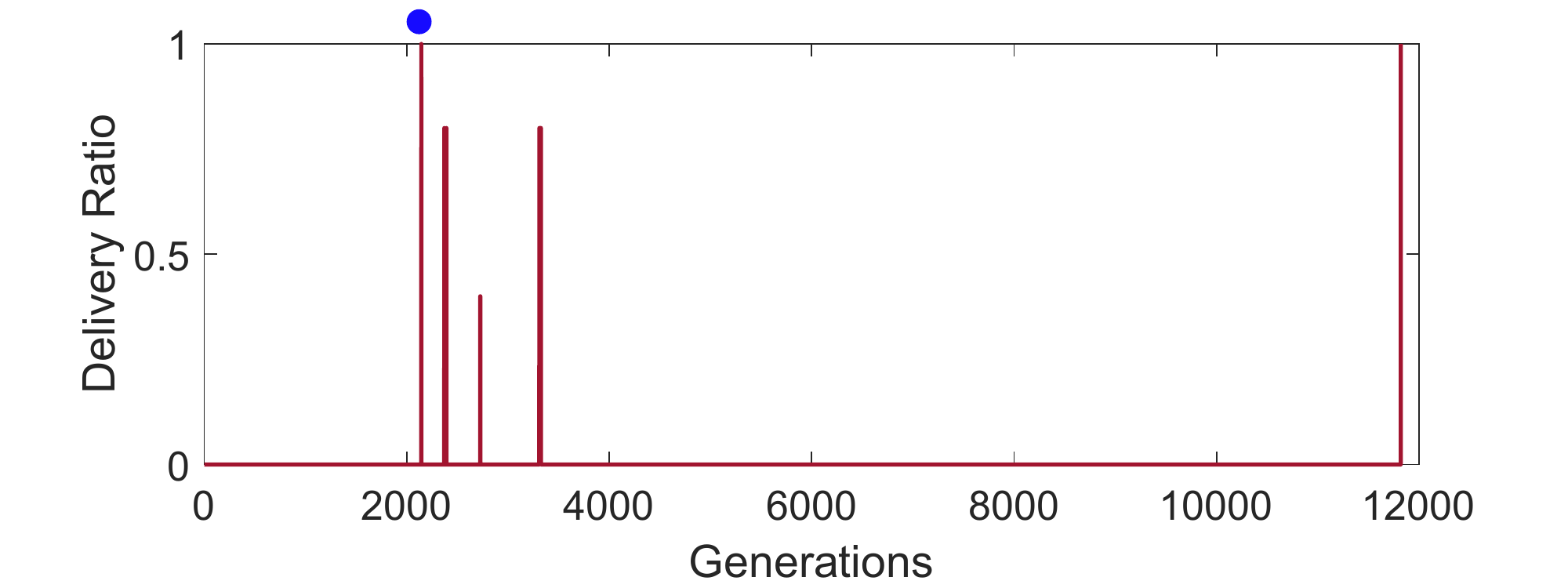}\label{fig:deliveryRatioInit0125}}
\subfloat[Node-1 fitness (Reinit. 36cp0125)]{\includegraphics[width=0.33\columnwidth]{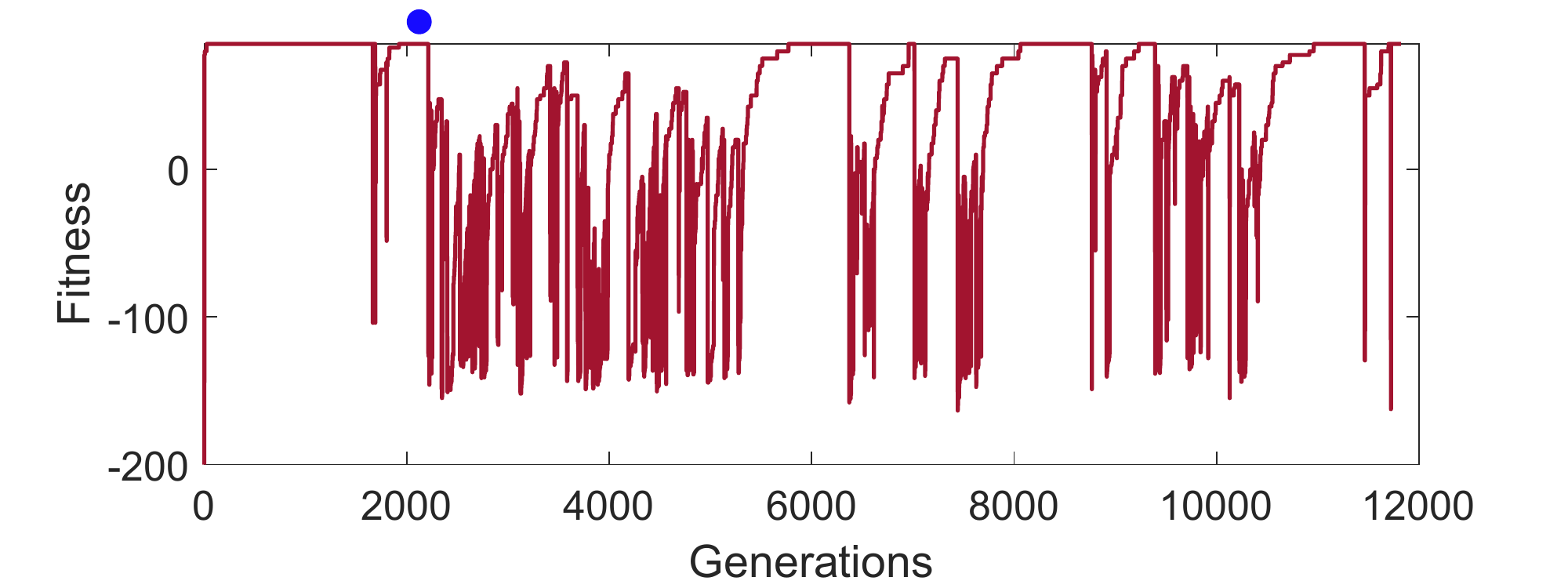}\label{fig:a1Init0125}}
\subfloat[Node-2 fitness (Reinit. 36cp0125)]{\includegraphics[width=0.33\columnwidth]{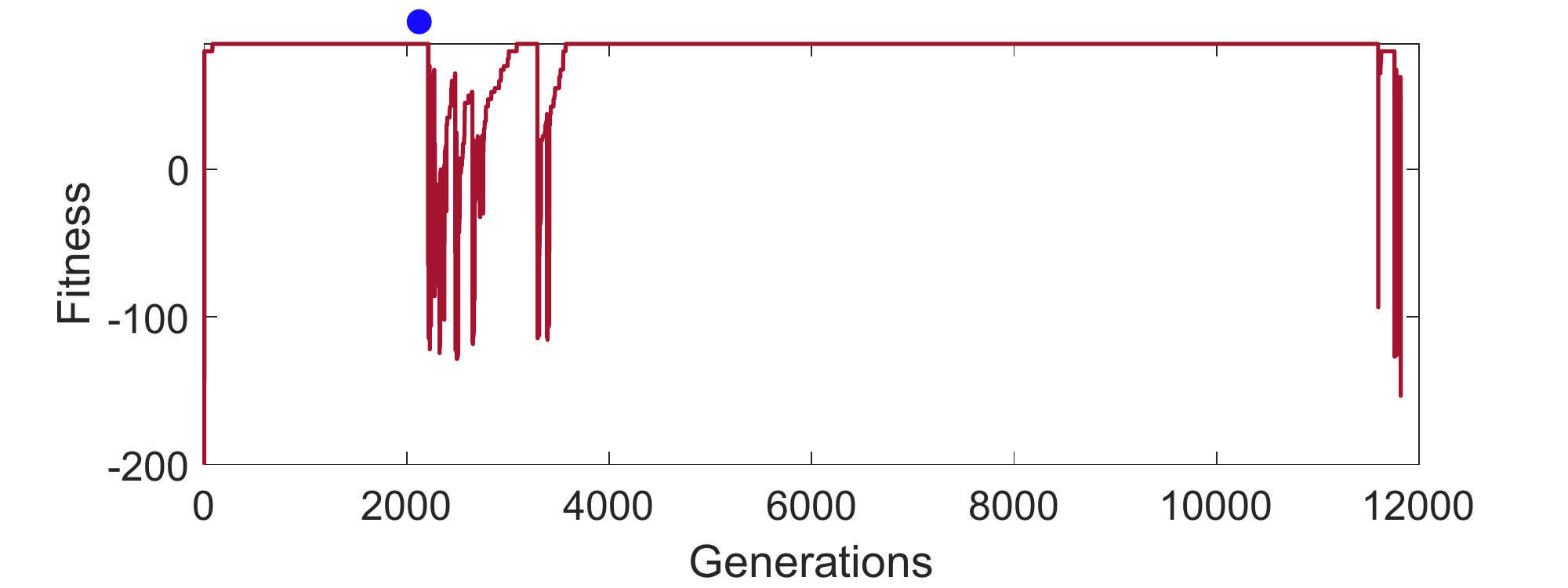}\label{fig:a2Init0125}}

\end{subfigures}
\caption{Delivery rate of the network (first column) and fitness trend of two randomly selected nodes (second and third columns) during example evolutionary processes performed for the robustness experiments. The rows provide example processes for 36cp1, 36cp05, 36cp025 and 36cp0125 problem instances with Add, Relocate, Remove and Reinitialization perturbations respectively. The blue dots indicate the end of the evolutionary process before applying the perturbation (starting from the initial network until $100\%$ delivery rate is achieved).} \label{fig:runtimeAnalysis}
\end{figure*}

Figure~\ref{fig:runtimeAnalysis} shows the delivery rate of the networks and the fitness trends of two randomly selected nodes during the evolutionary processes performed for the robustness experiments. For example, in the experiment with the addition perturbation (shown in the first row), 100\% delivery rate is achieved around the $200$th generation with the initial network. Then, the perturbation decreases the delivery rate. However, \ALGABRV~is able to recover from the perturbation and achieve 100\% delivery rate again in about 30 generations.

\end{document}